\documentclass{mfm3_preprint}

\usepackage{amsmath,amssymb,amsthm}
\usepackage{todonotes}
\usepackage{float}
\usepackage{caption}
\usepackage{tikz}
\usepackage[ruled,noend]{algorithm2e}
\usepackage[table]{xcolor}
\usepackage{fontawesome5}
\graphicspath{{figures/}}

\renewcommand{\paragraph}[1]{\vspace{.1em}\noindent\textbf{#1}}

\def\tr{\operatorname{tr}}
\def\sg{\operatorname{sg}}

\newcommand{\bb}[1]{\mathbb{#1}}

\newcommand{\ca}[1]{\mathcal{#1}}
\newcommand{\pt}[1]{\partial_{\theta}}
\newcommand{\tp}[1]{^{\top}}

\usepackage{xcolor}
\definecolor{trcol}{HTML}{1769E0}
\definecolor{bdcol}{HTML}{E8590C}
\definecolor{gdcol}{HTML}{0E8F7E}

\newcommand{\transport}[1]{{\color{trcol}#1}}
\newcommand{\boundary}[1]{{\color{bdcol}#1}}
\newcommand{\guide}[1]{{\color{gdcol}#1}}

\definecolor{rowhl}{HTML}{E3F1EA}

\theoremstyle{plain}
\newtheorem{theorem}{Theorem}
\newtheorem{proposition}[theorem]{Proposition}

\newtheorem{remark}[theorem]{Remark}
\newtheorem{lemma}[theorem]{Lemma}
\usepackage[most]{tcolorbox}
\usepackage{wrapfig}

\definecolor{palered}{RGB}{195, 145, 145}
\definecolor{paleblue}{RGB}{150, 200, 235}

\usepackage{xcolor}

\definecolor{boxframe}{HTML}{B8AA95}   % warm taupe border
\definecolor{boxback}{HTML}{FBF7EE}    % soft cream fill
\definecolor{boxtext}{HTML}{2F2A24}    % warm dark text

\usetikzlibrary{arrows.meta}   % for the arrow tip

\definecolor{posfill}{HTML}{E2EDE4}  \definecolor{posdraw}{HTML}{438A53}
\definecolor{negfill}{HTML}{F4E2E9}  \definecolor{negdraw}{HTML}{BB5078}
\definecolor{cfill}{HTML}{EFEDFD}    \definecolor{cdraw}{HTML}{9087D4}
\definecolor{ctext}{HTML}{5849B3}
\definecolor{xdraw}{HTML}{D6D6D3}    \definecolor{xtext}{HTML}{3F3E3B}
\definecolor{arrowgray}{HTML}{8F8E88} 
\newtcolorbox{figbox}[1][]{
  enhanced,
  colback=boxback,
  colframe=boxframe,
  boxrule=0.8pt,
  arc=3mm,
  left=3mm, right=4mm, top=2mm, bottom=3mm,
  #1
}

\newtcolorbox{propbox}{
    enhanced,
    colback=paleblue!35!white,
    colframe=paleblue,
    boxrule=0pt, leftrule=5pt, arc=0pt,
    left=8pt, right=8pt, top=8pt, bottom=8pt,
    fonttitle=\bfseries\sffamily
}

\newtcolorbox{theorembox}{
    enhanced,
    colback=palered!35!white,
    colframe=palered,
    boxrule=0pt, leftrule=5pt, arc=0pt,
    left=8pt, right=8pt, top=2pt, bottom=2pt,
    fonttitle=\bfseries\sffamily
}

\definecolor{MethodFill}{HTML}{FBF6E9}
\definecolor{MethodLine}{HTML}{A5BFA0}
\definecolor{ResultFill}{HTML}{FBF6E9}
\definecolor{ResultLine}{HTML}{D7C699}

\definecolor{AbstractBorder}{HTML}{B8AA95}
\tcbset{
  resultbox/.style={
    enhanced,breakable,
    colback=ResultFill,colframe=ResultLine,
    boxrule=0pt,leftrule=5pt,
    arc=0pt,outer arc=0pt,
    left=6pt,right=4pt,top=6pt,bottom=6pt,
    before skip=10pt,after skip=10pt
  }
}
\newtcolorbox{methodbox}[1]{
  enhanced,breakable,
  colback=MethodFill,colframe=MethodLine,
  boxrule=0pt,leftrule=5pt,
  arc=0pt,outer arc=0pt,
  left=8pt,right=8pt,top=8pt,bottom=8pt,
  before skip=10pt,after skip=10pt,
  before upper={\noindent{\bfseries#1.} \ }
}

\newtcolorbox{contribbox}[1][]{
  enhanced,
  colback=MethodFill,
  colframe=AbstractBorder,
  boxrule=0.8pt,
  arc=3mm,
  left=3mm, right=4mm, top=2mm, bottom=3mm,
  #1
}

\theoremstyle{plain}
\tcolorboxenvironment{theorem}{resultbox}
\tcolorboxenvironment{proposition}{resultbox}
\tcolorboxenvironment{lemma}{resultbox}
\tcolorboxenvironment{corollary}{resultbox}
\theoremstyle{remark}
\newcommand{\R}{\mathbb R}
\newcommand{\E}{\mathbb E}
\newcommand{\id}{\operatorname{id}}

\title{\fontsize{20}{24}\selectfont Scalable Beckmann Maps Improve One-step Generation}
\author[1,$\star$]{Ulrik Unneberg}
\author[1,$\star$]{Shiyi Wang}
\author[1]{Lee Cheuk-Kit}
\author[2]{Felix Berg}
\author[1]{Boris Kozinsky}
\author[1]{Yilun Du}
\author[1]{Michael S. Albergo}
\author[3,4]{Eric Vanden-Eijnden}
\affiliation[1]{Harvard University}
\affiliation[2]{Imperial College London}
\affiliation[3]{New York University}
\affiliation[4]{Capital Fund Management}

\abstract{Beckmann Transport Models (BTM) learn one-step generative maps end-to-end from data, without a teacher, a distillation stage, or pretrained features. We introduce Scalable Beckmann Maps (SBM), which bring to BTM the two ingredients behind the quality of modern multistep generators: classifier-free guidance (CFG) and a training recipe that scales, while keeping sampling at a single network evaluation. CFG combines conditional and unconditional velocity or score predictions, which a Beckmann map does not have. Our guidance rests on one observation: the BTM current is a positive multiple of an autonomous denoising field, to which CFG applies as it does to a velocity. With exact denoisers this gives guided BTM. In practice we substitute the map's own conditional and null predictions for the denoisers, which requires no additional network and introduces a bias that we characterize. We show that the guided update is consistent with a guided conservation equation, and that guidance can move the boundaries between basins, sending more noise to the modes a condition favors. Trained from scratch in the latent of a representation autoencoder, SBM reaches a one-step FID of 1.6 on ImageNet 256×256 outperforming the Improved-MeanFlow baseline with 3.2x less FLOPs without a teacher or a feature loss, and a GenEval score of 0.86 in text-to-image generation with one network evaluation. As a world model, where we train without guidance, the recipe alone outperforms DriftWorld on all five metrics of two robot-manipulation benchmarks, with 1.9–2.8× less training compute and no pretrained auxiliary feature encoder.}

\abstractfooter{%
  \faGlobe\ \href{https://tullebulle.github.io/projects/sbm/}{Project Page}
}

\begin{document}
\maketitle
\begingroup
\renewcommand{\thefootnote}{}
\footnotetext{Preprint. \textsuperscript{$\star$}Equal Contribution. Correspondence: ulrikunneberg@g.harvard.edu; fwang@math.harvard.edu. Code and checkpoints are available at the project page, \url{https://tullebulle.github.io/projects/sbm/}.}
\endgroup

\begingroup
\setlength{\intextsep}{8pt}
\begin{figure}[H]
 \centering
 \captionsetup{skip=6pt}
 \includegraphics[width=\linewidth]{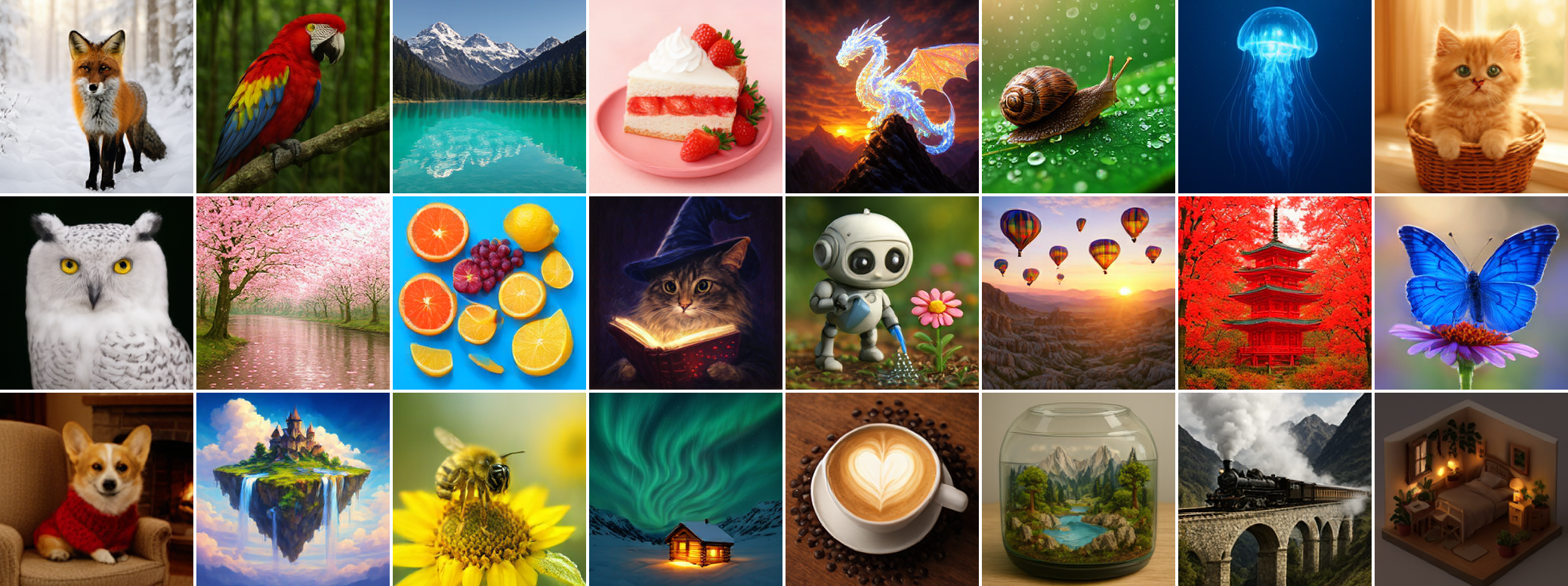}
 \caption{\textbf{Text-to-image samples of Scalable Beckmann Maps (SBM).} Selected samples of SBM-XL, a one-step map trained from scratch on the MiniT2I data: one network evaluation per image, $256\times256$ pixels.}
 \label{fig:teaser}
\end{figure}
\endgroup

\section{Introduction}
\label{sec:intro}
One-step generators replace the many network evaluations of diffusion and
flow-matching samplers
\citep{ho2020denoising,song2021score,lipman2023flow,albergo2023building}
with a single network call. Most existing one-step samplers are trained
either in two stages, where a multistep sampler is first trained and then
distilled into one step
\citep{salimans2022progressive,yin2024dmd,yin2024dmd2,salimans2024moments},
or by self-distillation \citep{song2023consistency,boffi2025build}. The
multistep flow and the one-step sampler may also be learned jointly
\citep{geng2025meanflow,geng2025improvedmeanflow}. Beckmann Transport
Models (BTM) \citep{lee2026btm,tang2026discrete} take a more direct route.
The endpoint map of a time-independent flow is constant along the
flow's probability current. Enforcing this property gives a loss that
trains the map directly from data, without a teacher, a second
prediction head, or an auxiliary pretrained feature network. We call that
endpoint map, learned directly, a Beckmann map.

BTM has so far lacked the combination of two ingredients behind modern
generators: a mature training recipe and classifier-free guidance
(CFG). Diffusion and flow models benefit from years of refinement of
noise schedules, loss weighting, preconditioning, architectures, and
sampling procedures \citep{karras2022,esser2024sd3}. The original BTM
paper reports a one-step FID of 17.6 at 1280 epochs on ImageNet $256\times256$
\citep{lee2026btm}, whereas recent one-step generators reach FIDs below 2
\citep{geng2025improvedmeanflow,deng2026drifting}. CFG \citep{cfg} is
particularly important for strong conditional generation on ImageNet
and text-to-image tasks
\citep{ma2024sit,peebles2023scalablediffusionmodelstransformers,glide},
and \citet{lee2026btm} leave its extension to BTM as an open problem.
In our experiments, unguided Beckmann maps under-separate classes and
over-disperse within them (Section~\ref{subsec:btm_comp}).

Guidance does not carry over directly to native one-step generators. CFG
combines two predictions of the same field, a conditional and an
unconditional velocity or score, and training-time variants build this
combination into a velocity target \citep{geng2025meanflow}. A Beckmann map
has no velocity or score to combine; it has only the map. Drifting models
also guide during training, by mixing unconditional data into the negative
samples of the drifting field \citep{deng2026drifting}, but BTM has no
negative samples to mix. We want to guide a Beckmann map during training,
so that sampling still takes a single network function evaluation (NFE).

We introduce Scalable Beckmann Maps (SBM), which bring guidance to
Beckmann maps through an autonomous denoising interpretation
(Section~\ref{sec:guidance}). The BTM current is a positive multiple of
a field pointing toward a weighted posterior mean of the data.
Applying CFG to this denoiser and writing it in implicit form gives
a correction involving the guided and null denoisers. We substitute
the map's own conditional and null predictions for these quantities,
giving the ordinary BTM direction $x_1-x_0$ plus a correction
proportional to their disagreement (Figure~\ref{fig:toy}a).
With exact denoisers, the conditional mean of this direction is a
positive rescaling of the autonomous CFG field. Substituting map
predictions introduces a bias, which we characterize in
Appendix~\ref{app:autonomous-guidance}.  For a fixed map, the correction preserves the conditional covariance of the
BTM transport target given the input and interpolation time, whereas the
explicit guided target scales it by $\omega^2$
(Proposition~\ref{prop:guidance-variance}, Table \ref{tab:guide-variance}). Both predictions come from
one network through condition dropout, so no additional denoiser is
trained and sampling remains one NFE. The guided current depends on
the map itself, and we seek a self-consistent map that predicts the
endpoints of its own current. This construction can move basin
boundaries and assign more source mass to modes favored by the
condition (Figure~\ref{fig:toy}c).

In a controlled comparison with the same DiT-B/2 architecture, SD-VAE
latent space, data, and one-step sampler, SBM reaches in ten epochs
an FID that native BTM does not reach in a hundred
(Section~\ref{subsec:btm_comp}). This improvement combines guidance
with our second ingredient: a training recipe whose choices we
describe in Algorithm~\ref{algo:training_complicated} and ablate in
Section~\ref{sec:ablations}. The guided targets provide the largest
improvement. As for diffusion and flows, guidance is not always essential. For world modeling, we train without
guidance (or auxiliary feature loss); the training recipe alone
outperforms the one-step DriftWorld baseline \citep{driftworld}.

Our contributions are:
\begin{contribbox}
\begin{enumerate}
  \item \textbf{Guidance and a training recipe for one-step maps.}
  We introduce training-time guidance for Beckmann maps together
  with a recipe for training at scale. The method uses the map's
  own conditional and null predictions, requires no additional
  denoiser, and retains one NFE at sampling time.

    \item \textbf{Theoretical interpretation and guarantees.}
    We derive the guidance rule through an autonomous denoising
    interpretation of BTM and characterize the bias introduced by
    substituting map predictions for denoisers. For the idealized
    population objective with clean boundary supervision, we establish
    consistency with a guided conservation equation and show, through
    a two-mode construction, how guidance can move basin boundaries
    and increase the probability of modes favored by the condition.

  \item \textbf{Performant one-step generation across tasks.}
  Our SBM-XL generator trained from scratch achieves a one-step FID of 1.6 on
  ImageNet $256\times256$
  without a teacher or auxiliary feature loss, outperforming the Improved-MeanFlow baseline with 3.2x less FLOPs. In the RAE latent, SBM-XL reaches in 3.5 epochs the FID that BTM-XL/2 reaches in 1,280 epochs.
  For Text-to-Image generation, SBM also reaches a GenEval score of 0.86, comparable to MiniT2I flow-matching at 100 steps. On BridgeData V2 \citep{bridge2} and RT-1
  \citep{rt1}, our world models use neither guidance nor an
  auxiliary feature loss and outperform every reported baseline on all five metrics.
\end{enumerate}
\end{contribbox}

\begin{figure}[ht]
    \centering
    \includegraphics[width=1\linewidth]{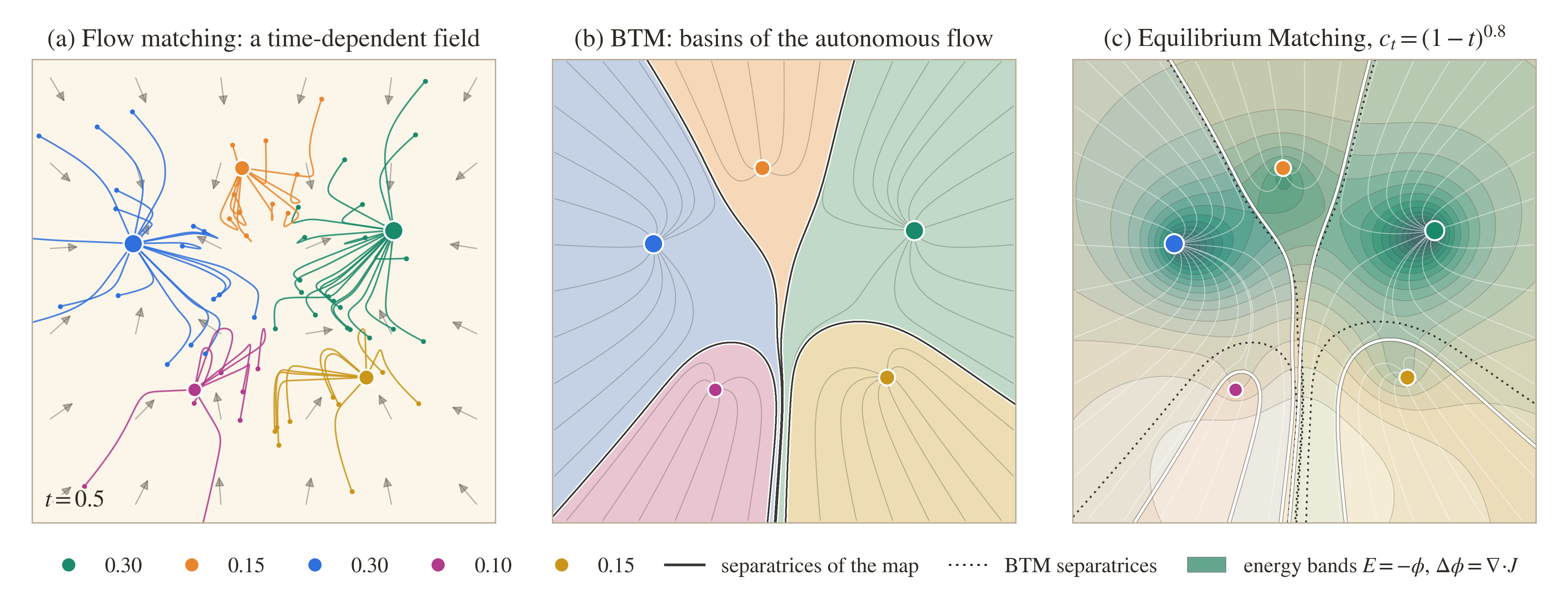}
    \caption{Flow matching, BTM and Equilibrium Matching on the same two-dimensional problem: noise is transported to five data points, whose weights are given in the legend. (a) Flow matching learns a time-dependent velocity; arrows show the field at $t=0.5$. Because the velocity changes over time, trajectories can bend sharply and even reverse direction, and this curvature makes the ODE hard to integrate in few steps. (b) BTM assumes an autonomous flow. Its flow lines are fixed and never cross, the endpoint map is constant along each of them, and the separatrices of the map (solid) divide the plane into one basin per data point, each receiving source mass equal to that point’s weight. (c) Equilibrium Matching \citep{wang2025equilibrium} also learns a time-invariant field, whose magnitude is set by a schedule $c_t$, here $c_t = (1-t)^{0.8}$; the shaded bands are a potential reconstructed from the current, $\Delta\phi=\nabla\cdot J$. A nonconstant schedule $c_t$ can bias the endpoint distribution by shifting the basin boundaries: the separatrices of its map (solid) depart from those of BTM (dotted), and the basins of the two lower modes shrink, so these modes receive too little mass. At $c_t = 1$ the autonomous field is identical to that of BTM.}
    \label{fig:placeholder}
\end{figure}

\subsection{Related work}
\paragraph{Flow matching and flow maps.}
Flow matching~\citep{lipman2023flow,albergo2023building,liu2023rectifiedflow}
and diffusion models~\citep{ho2020denoising,song2021score} learn
time-dependent drifts that transport a base distribution to data. They
have scaled to strong generation on ImageNet~\citep{ma2024sit} and
text-to-image tasks~\citep{esser2024sd3}. Flow-based world models have
also scaled to billions of parameters and hundreds of millions of
training video clips~\citep{cosmospredict25}. Sampling from these
models typically requires integrating the drift over many network
evaluations. To accelerate inference, consistency models and flow map
matching~\citep{song2023consistency,boffi2025flowmap,boffi2025build}
instead learn the solution operator of a flow. This usually involves
either distilling a trained multistep model in a separate stage or
self-distilling a map alongside an auxiliary velocity. Empirically,
prior works~\citep{geng2025meanflow,geng2025improvedmeanflow,frans2025shortcut}
have shown that joint training can produce strong one-step ImageNet
models. Other works ~\citep{minit2i2026, shaul2026pdd, yang2026arcflow, sabour2025alignyourflow}
have shown that flow maps also scale to text-to-image generation. SBM 
instead presents a scalable one-stage training protocol without
self-distillation, simplifying the training recipe for strong one-step
generation.

\paragraph{Autonomous flows and their terminal maps.}
Autonomous flow matching~\citep{lee2026btm} has recently emerged as a
promising alternative to time-dependent flow matching by learning a
time-independent drift. Prior empirical studies have demonstrated its
potential for ImageNet and video
generation~\citep{sun2025noise,wang2025equilibrium,lillemark2026equilibriumforcingadaptivevideo}.
BTM~\citep{lee2026btm} provided a theoretical justification for these
empirical successes by modeling generation as an autonomous flow
terminated upon reaching the data manifold. This perspective leads to
a direct-map objective that learns the terminal map from data without
self-distillation, with promising results in one-step
generation. Building on this foundation, SBM demonstrates that guidance
and an improved training protocol enable direct terminal-map learning
to scale and achieve competitive performance with state-of-the-art
methods on class-conditional ImageNet generation, text-to-image
generation, and world modeling.

\paragraph{Feature-based distribution matching.}
Another class of distillation-free approach learns a one-step map by matching its pushforward distribution to the data in a feature space. Naively minimizing a distribution divergence can fail to converge; Drifting
\citep{deng2026drifting} and W-Flow~\citep{wflow} instead compare the
distributions in feature spaces supplied by pretrained encoders, often
in an ensemble, and yield strong one-step results on class-conditional ImageNet generation~\citep{deng2026drifting,wflow,zhang2026teacher}, text-to-image generation ~\citep{priming}, and world
modeling~\citep{driftworld}
However, the reliance on external encoders ties the generator's training
signal to features learned at a fixed pretraining scale. SBM
instead trains its map from scratch by enforcing conservation along an
autonomous flow, without a teacher or an auxiliary feature-matching or
alignment loss. Our strongest ImageNet and text-to-image models
nevertheless use DINOv2-based latents, training SBM directly in pixel
space remains open.

\section{Beckmann Transport Models}
\label{sec:background}
% =====================================================================

Let the interpolant be $I_s=\alpha_sI_0+\beta_sI_1$, $s\in[0,1]$, with noise
$I_0\sim\mu_0=\mathcal N(0,I_d)$ and data $I_1\sim\mu_1$, with
$\alpha_0=\beta_1=1$ and $\alpha_1=\beta_0=0$. Let $\rho_s$ be the density of $I_s$ and
$b_s(x)=\bb E[\dot I_s\mid I_s=x]$ be the flow-matching velocity. BTM
\citep{lee2026btm} replaces $b_s$ by a time-independent current which averages over
time. The current and occupation measure is defined by
\begin{equation}
\label{eqn:current}
    J(x)=\int_0^1\rho_s(x)\,b_s(x)\,ds,\qquad \nu(x)=\int_0^1\rho_s(x)\,ds.
\end{equation}
The autonomous velocity $b(x)$ is defined by the ratio $b=J(x)/\nu(x)$, which can be written $b(x)$ as the minimizer of
\begin{equation}
    \label{eqn:auto loss}
    b(x)=\arg\min_{\hat b}\bb E_{s,I_0,I_1}\|\hat b(I_s)-\dot I_s\|^2,
\end{equation} from which $b(x)$ can be trained. The occupation measure $\nu(x)$ is the density of $I_s$ for
$s\sim U(0,1)$, and $\nu(x)\,dx$ is the expected time an interpolant spends
in $dx$. Integrating the continuity equation
$\partial_s\rho_s+\nabla\cdot(\rho_sb_s)=0$ over $s$ gives
\begin{equation}
\label{eqn:divergence}
    \nabla\cdot J=\mu_0-\mu_1,
\end{equation}
the flux constraint of Beckmann's transportation problem
\citep{beckmann1952continuous,santambrogio2015ot}. When $\mu_1$ is
supported on a manifold $M$ of codimension at least two, under
the regularity and straight-interpolant assumptions of BTM, the ODE $\dot X=b(X)$ pushes the distribution $\mu_0$
forward to $\mu_1$ as supported on manifold $M$ \citep{lee2026btm} along the trajectories of the ODE. The
map $T$ sending a point along a trajectory to the endpoint is
constant along the flow lines of $b(x)$. The flow lines of $b(x)$ are also flow lines of $J(x)$.
\begin{equation}
\label{eqn:conservation}
    J(x)\cdot\nabla T(x)=0\quad\text{for }x\notin M,\qquad T(x)=x\quad\text{for }x\in M,
\end{equation}
where $J\cdot\nabla T(x)$ is the derivative of $T$ in the direction $J(x)$. BTM learns the conservation of $T(x)$ in the $J(x)$ direction with a stop-gradient \transport{transport loss} and a
\boundary{boundary loss}:
\begin{equation}
    \label{eqn:btm loss}
    \mathcal{L}(T)=\transport{\underbrace{\bb E_{s,I_0,I_1}\big\|T(I_s)-\sg\big(T(I_s)+\dot I_s\cdot\nabla T(I_s)\big)\big\|^2}_{\ca L_{\tr}} }+\lambda \boundary{\underbrace{\bb E_{I_1}\big\|T(I_1)-I_1\big\|^2}_{\ca L_{\partial}}}.
\end{equation}
The population stopped update of~\eqref{eqn:btm loss} vanishes at the ground truth Beckmann Transport Map \citep[Theorem~3]{lee2026btm}.

The name comes from the flux constraint~\eqref{eqn:divergence}: Beckmann's
transportation problem asks for a flux $J$ with $\nabla\cdot J=\mu_0-\mu_1$,
and the time-averaged current of an interpolant is one such flux.
\citet{lee2026btm} call the framework built on it Beckmann Transport Models.
It covers both the autonomous velocity $b=J/\nu$, trained
through~\eqref{eqn:auto loss} and sampled in many steps, and the endpoint map
$T$ of its flow, trained directly through~\eqref{eqn:btm loss} and sampled in
one. We use BTM for the framework as a whole and \emph{Beckmann map} for the
endpoint map $T$ trained end to end. It is the map that we scale, hence
Scalable Beckmann Maps.

For class-conditional generation, every quantity
above is conditioned on a label $c$, the map $T_\theta(x,c)$ takes $c$ as input,
and $I_1\sim\mu_1^c$. Neither term of \eqref{eqn:btm loss} ever shows the map what a wrong image for a given class looks like.

\begin{remark}
\label{remark:linear}
Currents in equation \eqref{eqn:current} can be added when target distributions compose, but velocities in \eqref{eqn:auto loss} can not. Since $I_0$ is independent of $I_1$, $J(x)=\int_0^1\bb E[\dot I_s\,\delta(x-I_s)]\,ds$
and $\nu$ are expectations under $\mu_0\otimes\mu_1$ and are linear in $\mu_1$:
if $\mu_1=\sum_{c'}\pi_{c'}\mu_1^{c'}$, then $J=\sum_{c'}\pi_{c'}J_{c'}$ and
$\nu=\sum_{c'}\pi_{c'}\nu_{c'}$, where $J_{c'}$ and $\nu_{c'}$ are computed from $\mu_1^{c'}$. The velocity $b=J/\nu$ is a ratio, and is not additive across the individual velocities $b_{c'}(x)$.
\end{remark}

\section{Guidance for one-step Beckmann maps}
\label{sec:guidance}

A Beckmann map predicts the final sample directly. We introduce guidance
during training by comparing its conditional and null predictions, while
retaining a single network evaluation at sampling time. We motivate the guidance rule through an autonomous denoising interpretation of BTM, identify its induced current, connect the training objective to endpoint conservation, and show that guidance can reweight output probabilities.

Let $\mu_1^c$ be the data distribution for condition $c$, let
$\pi_{c}>0$ be the probability of condition $c$, and let
$\mu_1^\emptyset=\sum_{c'}\pi_{c'}\mu_1^{c'}$ be the null distribution. Write $T_c^1$ for the ordinary conditional Beckmann map and
$T_\emptyset$ for the null map. For a fixed condition and guidance strength
$\omega\geq1$, write $T$ for the candidate guided map.

\begin{figure}
    \centering
    \includegraphics[width=1\linewidth]{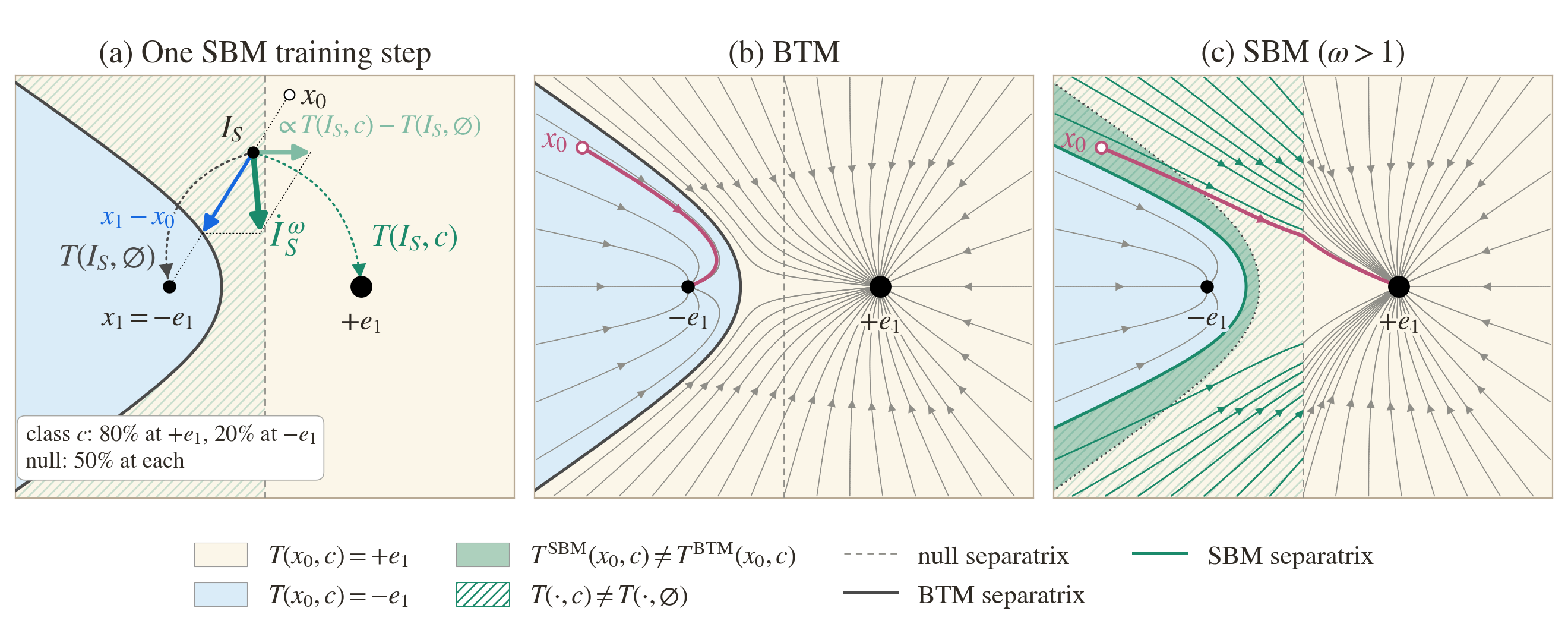}
    \caption{\textbf{How SBM guides a one-step map.}
Two-mode example of Section~\ref{sec:reweighting}: Gaussian noise in $\R^2$, data at $\pm e_1$,
class $c$ with weights $0.8/0.2$ and null with $0.5/0.5$. Background colors show where the
conditional map sends each noise point.
\textbf{(a)} One training step. A class-$c$ pair whose data point is the minority mode,
$x_1=-e_1$, gives the BTM transport direction $x_1-x_0$ (blue). At $I_S$ the conditional and
null predictions disagree (hatched), and the correction
$\frac{1-\omega^{-1}}{1-S}\bigl(T(I_S,c)-T(I_S,\emptyset)\bigr)$ (light green) tilts the
direction toward the mode the condition favors, giving the guided direction
$\dot I_S^\omega$ of~\eqref{eq:direction} (dark green).
\textbf{(b)} BTM: flow lines of the conditional current and the basins of $T(\cdot,c)$.
\textbf{(c)} SBM with $\omega>1$: the current changes only where
$T(\cdot,c)\neq T(\cdot,\emptyset)$ (green flow segments). The separatrix moves left, so noise
in the green region is reassigned from $-e_1$ to $+e_1$.}
    \label{fig:toy}
\end{figure}

\subsection{From endpoint predictions to a guided current}

Draw $x_0\sim\mu_0$, $x_1\sim\mu_1^c$, and $S\sim U(0,1)$ independently,
and use the same linear interpolant as ordinary BTM:
\begin{equation}
 I_s=(1-s)x_0+sx_1,
 \qquad \dot I_s=x_1-x_0.
 \label{eq:interpolant}
\end{equation}
We write $\E_c$ for expectation over these draws.

Let $\rho_s^c$ denote the density of $I_s$, and let $\nu_c$
denote the law of $I_S$, with density
$\nu_c(x)=\int_0^1\rho_s^c(x)\,ds$. Let $h_c(x)
 =\int_0^1\frac{\rho_s^c(x)}{1-s}\,ds$, and
define the autonomous denoiser
\begin{equation}
\begin{aligned}
 \bar z_c(x)
 &=\frac{\E_c[x_1/(1-S)\mid I_S=x]}
         {\E_c[(1-S)^{-1}\mid I_S=x]},
\end{aligned}
\label{eq:autonomous-denoiser}
\end{equation}
where $0<h_c(x)<\infty$, with null quantities defined analogously.
Since $\dot I_s=(x_1-I_s)/(1-s)$, the ordinary conditional current satisfies
\begin{equation}
 J_c(x)
 =\nu_c(x)\E_c[\dot I_S\mid I_S=x]
 =h_c(x)\bigl(\bar z_c(x)-x\bigr).
 \label{eq:denoiser-current}
\end{equation}
Thus $\bar z_c-\id$ defines an autonomous denoising field with the
same oriented trajectories as the ordinary BTM current. Applying the usual CFG combination~\citep{cfg,guidedflows} to these denoisers gives the equivalent explicit and implicit forms
\begin{equation}
\begin{aligned}
 \bar z_c^\omega
 &=\bar z_\emptyset+\omega(\bar z_c-\bar z_\emptyset)\\
 &=\bar z_c+
   \guide{(1-\omega^{-1})
   (\bar z_c^\omega-\bar z_\emptyset)}.
\end{aligned}
\label{eq:velocitycfg}
\end{equation}
We substitute the guided and null endpoint predictions
$(T,T_\emptyset)$ for $(\bar z_c^\omega,\bar z_\emptyset)$
in the implicit correction, giving the guided transport direction:
\begin{methodbox}{Guided transport direction}
\begin{equation}
 \dot I_S^\omega
 =\dot I_S+
 \guide{\frac{1-\omega^{-1}}{1-S}
 \bigl(T(I_S)-T_\emptyset(I_S)\bigr)}.
 \label{eq:direction}
\end{equation}
\end{methodbox}

The factor $1-\omega^{-1}$ comes from implicit CFG, while
$1/(1-S)$ converts endpoint displacements into interpolant
velocities. No additional denoiser network is introduced.
At $\omega=1$, the direction reduces to ordinary conditional BTM.
For $\omega>1$, the correction follows conditional--null endpoint
disagreement at $I_S$, while the interpolation path remains unchanged. At fixed input and interpolation time, the correction in \eqref{eq:direction}
preserves the conditional covariance of the unguided transport
target (Proposition~\ref{prop:guidance-variance}). We could alternatively use the explicit guided direction
$\dot I_S^{\omega,\mathrm{exp}}=\omega \dot I_S-(\omega-1)\frac{T_\emptyset(I_S)-I_S}{1-S}$,
which carries $\omega$ times the noise of the unguided target and has the same
conditional mean as \eqref{eq:direction} when the map and its null prediction
are the exact denoisers (Appendix~\ref{app:autonomous-guidance}). At $\omega=5$ the boundary
term's gradient variance is 85 times that of \eqref{eq:direction} and the map
reads 66 against 21 FID at 20 epochs (Table~\ref{tab:guide-variance}). More details on the guidance are given in Appendix \ref{app:autonomous-guidance}.

Averaging at a fixed input identifies the induced autonomous
velocity and its current:
\begin{equation}
\begin{aligned}
 \E_c[\dot I_S^\omega\mid I_S=x]
 &=\frac{J_c^\omega[T](x)}{\nu_c(x)}\\
 &=\frac{h_c(x)}{\nu_c(x)}
   \left[
   \bar z_c(x)-x+
   (1-\omega^{-1})\bigl(T(x)-T_\emptyset(x)\bigr)
   \right],\\
 J_c^\omega[T](x)
 &=J_c(x)+\Delta J_c^\omega[T](x),\\
 \Delta J_c^\omega[T](x)
 &=(1-\omega^{-1})h_c(x)
   \bigl(T(x)-T_\emptyset(x)\bigr).
\end{aligned}
\label{eq:current}
\end{equation}
The sampled direction is therefore conditionally unbiased for
the map-dependent velocity $J_c^\omega[T]/\nu_c$.
With exact denoisers in place of the maps, this velocity becomes
$(h_c/\nu_c)(\bar z_c^\omega-\id)$, a positive rescaling of
the autonomous CFG field. 

Endpoint maps generally differ from these posterior means. The resulting substitution error is exactly:
\begin{equation}
 \frac{J_c^\omega[T](x)}{h_c(x)}
 -\bigl(\bar z_c^\omega(x)-x\bigr)
 =(1-\omega^{-1})
 \left[
 (T-\bar z_c^\omega)(x)
 -(T_\emptyset-\bar z_\emptyset)(x)
 \right].
 \label{eq:guidance-bias}
\end{equation}

Thus the field error depends on the mismatch between the
conditional--null contrasts; the two individual substitution
errors can cancel. Appendix~\ref{app:autonomous-guidance} derives
the identities of this section.

\subsection{Endpoint consistency}
\label{sec:existence}

The current now depends on the map we seek. We therefore ask for a map
that predicts the endpoint of its own current: starting from $x$ and
following $J_c^\omega[T]$ should end at $T(x)$. We call this property
\emph{self-consistency}. For such a map, all points on a trajectory have the same prediction.

\begin{methodbox}{Guided conservation equation} Where the derivative exists, the guided conservation equation is:
\begin{equation}
 J_c^\omega[T]\cdot\nabla T=0\quad\text{a.e. off }M,
 \qquad T(x)=x\quad\mu_1^c\text{-a.e.},
 \label{eq:conservation}
\end{equation}
where $M$ is the data manifold and $J_c^\omega[T]\cdot\nabla T$ is the derivative of $T$ along the guided current, as in~\eqref{eqn:conservation}.
\end{methodbox}

Conversely, if conservation and the chain rule hold along a
trajectory reaching $M$, an identity terminal trace identifies
the conserved prediction with that trajectory's endpoint.
We seek a learning rule consistent with these conditions.

\subsection{Learning the map directly}
\label{sec:main}

We apply the BTM stopped update~\citep{lee2026btm} to the guided
direction~\eqref{eq:direction}. Its conditional-expectation
representation~\eqref{eq:current} lets us enforce conservation
along the guided current without evaluating that current explicitly.
For the boundary target, we use the same implicit guidance correction
with $x_1$ as the clean regression target:

\begin{methodbox}{Guided training losses} 
\begin{align}
 \transport{\mathcal L_{\mathrm{tr},c}^\omega(T)}
 &\transport{=\E_c\left\|T(I_S)-
   \sg\!\left(T(I_S)+\guide{\dot I_S^\omega}\cdot\nabla T(I_S)\right)\right\|^2 } ,
 \label{eq:transport-loss}\\
 \boundary{\mathcal L_{\partial,c}^\omega(T)}
 &\boundary{= \E_c\left\|T(x_1)-\sg\!\left(
 x_1+\guide{(1-\omega^{-1})\bigl(T(x_1)-T_\emptyset(x_1) \bigr)}
 \right)\right\|^2}.
 \label{eq:boundary-loss}
\end{align}
The complete objective is
\begin{equation}
 \mathcal L_c^\omega
 =\transport{\mathcal L_{\mathrm{tr},c}^\omega }
  +\lambda \boundary{\mathcal L_{\partial,c}^\omega},
 \qquad\lambda>0.
 \label{eq:loss}
\end{equation}
\end{methodbox}
The \guide{green} highlights modifications from the unguided BTM loss. The stop-gradient operator freezes the entire target in each term,
including the guidance correction. At $\omega=1$, these losses reduce
to ordinary conditional BTM.

\begin{theorem}[Consistency of the guided stopped update]
\label{thm:vanishing}
Let $1\leq\omega<\infty$ and $\lambda>0$. Assume that
$\mu_1^c$ is supported on a smooth lower-dimensional manifold $M$,
that $\nu_c\ll dx$, and that $T$ is differentiable a.e. off $M$.
Suppose
\[
 T_\emptyset=\id\quad\mu_1^c\text{-a.e.},
 \qquad
 \E_c\|\dot I_S^\omega\|+\mathcal L_c^\omega(T)<\infty.
\]
Then
\[
 \eqref{eq:conservation}
 \quad\Longleftrightarrow\quad
 \delta_{\mathrm{sg}}\mathcal L_c^\omega(T)[H]=0
 \quad
 \forall H\in C_c^\infty(\R^d;\R^d),
\]
where $\delta_{\mathrm{sg}}$ denotes variation of the live prediction
with all stopped targets held fixed.
\end{theorem}

The proof is given in Appendix~\ref{proof:vanishing}.
The ideal null Beckmann map fixes the data on each class support
because every class has positive weight in the null mixture.
Under this trace, boundary guidance preserves the identity
condition and changes its stopped-update weight to $\lambda/\omega$.
The theorem characterizes the unweighted population stopped update
with the clean boundary loss; the sampled squared transport loss need not vanish.
It does not establish uniqueness of the guided map or convergence
of neural-network training.

In practice, the conditional and null maps are branches of one
network, $T_\theta(x,c,\gamma)$, trained with condition dropout;
$c=\emptyset$ denotes the null branch. The clean boundary loss uses input
$x_1$, while the noisy boundary loss uses
$I_\gamma=(1-\gamma)x_0+\gamma x_1$, retaining $x_1$ as the clean
base target in the guided loss. We call $\gamma$ the
\emph{boundary level} and also provide it as a network input.
In Algorithm~\ref{algo:training_complicated}, $\gamma$ controls
the sampled boundary time $S'\geq\gamma$; the actual boundary input
is $I_{S'}$, including examples with $S'=\gamma$.

To understand this supervision, first consider the idealized noisy
boundary loss alone at a fixed level $\gamma$, with every map
prediction evaluated at $I_\gamma$ and the null branch held fixed.
Its population stopped regression equation is
\begin{equation}
 T_\theta(x,c,\gamma)
 = \omega\,\E_c[x_1\mid I_\gamma=x]
   -(\omega-1)T_\theta(x,\emptyset,\gamma).
 \label{eq:anchor-denoising}
\end{equation}
At $\omega=1$, this is the conditional mean denoiser.
Writing $m_{\gamma,c}(x)=\E_c[x_1\mid I_\gamma=x]$, with the
null mean defined analogously, subtracting the null prediction from
\eqref{eq:anchor-denoising} gives
$T_\theta(x,c,\gamma)-T_\theta(x,\emptyset,\gamma)
=\omega\bigl(m_{\gamma,c}(x)-T_\theta(x,\emptyset,\gamma)\bigr)$.
If the null branch also approximates its mean denoiser,
$T_\theta(x,\emptyset,\gamma)\approx m_{\gamma,\emptyset}(x)$,
then
\begin{equation}
T_\theta(x,c,\gamma)-T_\theta(x,\emptyset,\gamma)
\approx
\omega\bigl(m_{\gamma,c}(x)-m_{\gamma,\emptyset}(x)\bigr).
\label{eq:contrast_id}
\end{equation}

Thus the map contrast approximates the denoiser contrast, scaled
by $\omega$. For the guidance correction, we set $\gamma=s$ in
both predictions, giving
$T_\theta(I_s,c,s)-T_\theta(I_s,\emptyset,s)$.
Multiplying this difference by $(1-\omega^{-1})/(1-s)$ gives
approximately
$(\omega-1)\bigl(m_{s,c}(I_s)-m_{s,\emptyset}(I_s)\bigr)/(1-s)$,
the usual CFG correction to the conditional flow velocity.

For generation, we fix $\gamma=\gamma_{\mathrm{read}}$ and
evaluate $T_\theta(x_0,c,\gamma)$ at a pure noise input $x_0$.
For this fixed boundary level, the transport objective encourages
$D_xT_\theta(x,c,\gamma)b_{c,\gamma}^\omega(x)\approx0$, where
$b_{c,\gamma}^\omega$ denotes the induced velocity for this branch:
the prediction should change little along a flow trajectory.
If that trajectory ends at $z$ and the boundary prediction satisfies
$T_\theta(z,c,\gamma)\approx z$, then
$T_\theta(x_0,c,\gamma)\approx T_\theta(z,c,\gamma)\approx z$.
This explains how transport training supports direct endpoint
prediction from noise.

Providing $\gamma$ as an input therefore encourages
$T_\theta(I_s,c,s)$ to act as a guided denoiser of $I_s$, using
$\gamma=s$, and $T_\theta(x_0,c,\gamma_{\mathrm{read}})$ to predict
the flow endpoint from pure noise, using a fixed read level.
The contrast relation~\eqref{eq:contrast_id} becomes exact when
both branches attain their ideal regression solutions for the
noisy boundary loss alone. The transport loss also influences
these predictions, so under joint training the relation is
generally approximate. Appendix~\ref{app:noisy-anchoring}
derives the regression equations and their balance with transport.

\paragraph{Local existence of guided endpoint maps.} The stationary characterization does not by itself guarantee that
trajectories reach the data manifold. Under the regularity assumptions
of Theorem~\ref{thm:existence} in Appendix~\ref{app:local},
sufficiently weak guidance admits a local map
$T_c^\omega:U\to M$ that predicts the endpoints of its own guided
current and remains close to the ordinary conditional Beckmann map.
Thus the guided equations admit actual endpoint solutions near the
data manifold. This is a local existence result; it does not establish
global arrival or convergence of training to these solutions. 

The stationary characterization and local existence result concern the
ideal objective with a clean boundary loss. For
Algorithm~\ref{algo:training_complicated}, conditional expectation still
identifies the current associated with its sampled transport update,
but frozen predictions, guidance restricted to a time band, and
transport sampling and weighting change that current. The noisy
boundary loss produces an interior balance whose relative weighting
is also affected by gradient balancing. The results therefore explain
the underlying construction but do not directly characterize the
sampled map $T_\theta(x_0,c,\gamma_{\mathrm{read}})$.

\begin{algorithm}[H] 
\caption{Simple SBM training ($\omega = 1$ is BTM)}\label{algo:training_simple}
\KwIn{map $T_\theta(\cdot, c)$, with $T_\theta(\cdot, \emptyset)$ its null branch; guidance strength $\omega \ge 1$; boundary gradient weight $\lambda$}
\For{each training step}{
    Draw $c \sim \pi$, $x_1 \sim \mu_1^c$, $x_0 \sim \mu_0$, $S \sim U(0, 1)$; Randomly replace $c$ by $\emptyset$ for condition dropout; \\ 
    set $I_S = (1 - S)\, x_0 + S\, x_1, $\;
    $\dot I_S^{\omega} = x_1 - x_0 + \dfrac{1 - \omega^{-1}}{1 - S} \bigl( T_\theta(I_S, c) - T_\theta(I_S, \emptyset) \bigr)$ \\
    $\mathcal L_{\mathrm{tr}, c}^\omega = \bigl\| T_\theta(I_S, c) - \sg\bigl( T_\theta(I_S, c) + \dot I_S^{\omega}\cdot\nabla T_\theta(I_S, c) \bigr) \bigr\|^2; $ \\
    $\mathcal L_{\partial, c}^\omega = \bigl\| T_\theta(x_1, c) - \sg\bigl( x_1 + (1 - \omega^{-1}) \bigl( T_\theta(x_1, c) - T_\theta(x_1, \emptyset) \bigr) \bigr) \bigr\|^2; $ \\
    $\theta \leftarrow \theta - \eta\, \nabla_\theta \bigl( \mathcal L_{\mathrm{tr}, c}^\omega + \lambda \mathcal L_{\partial, c}^\omega \bigr);$
}
\Return $T_\theta$
\end{algorithm}

\subsection{Guidance can reweight output probabilities}
\label{sec:reweighting}

The local result concerns where trajectories end near the data. Output
probabilities also depend on how much source mass is assigned to each
endpoint. Guidance can change these assignments. Indeed, wherever the
currents are defined and locally integrable, their divergence satisfies
\begin{equation}
 \nabla\cdot J_c^\omega[T]
 =\mu_0-\mu_1^c+\nabla\cdot\Delta J_c^\omega[T]
 \label{eq:divergence}
\end{equation}
in the distributional sense. The extra term changes the interior mass
balance, so the ordinary BTM transport identity does not determine the
guided output distribution.

\paragraph{Example: strengthening a preference between two modes.}
Take a standard Gaussian source in $\R^2$ and two data points,
$e_1$ and $-e_1$. Suppose
\[
 \mu_1^c=0.8\delta_{e_1}+0.2\delta_{-e_1},
 \qquad
 \mu_1^\emptyset=0.5\delta_{e_1}+0.5\delta_{-e_1}.
\]
The condition favors the positive mode. The ordinary conditional map
assigns $80\%$ of the source probability to that point. The construction
in Appendix~\ref{app:binary} shows that, for each finite $\omega>1$, self-consistent
guided endpoint maps exist with
\[
 \mathbb P_{X\sim\mu_0}\{T_c^\omega(X)=e_1\}>0.8,
 \qquad
 \mathbb P_{X\sim\mu_0}\{T_c^\omega(X)\in\{-e_1,e_1\}\}=1.
\]
These maps fix both data points and have zero numerical loss for the
objective with a clean boundary loss. They agree with the ordinary map near the endpoints but move the
boundary separating the two output regions. The positive region grows,
so it receives more Gaussian source mass. See Figure \ref{fig:toy}.

This example shows that the guided equations permit stronger preference
for a condition-favored mode while retaining data support. The unchanged
ordinary map is also a solution at the same guidance strength, so the
equations alone do not select the reweighting. The example illustrates a
possible change in the generated law; it is not a claim of improved
matching to $\mu_1^c$ or of convergence to that solution during training.

% FIGURE SUGGESTIONS (not additional claims or generated results).
% 1. Training overview: show the original straight interpolant; conditional
%    and null endpoint predictions; add their scaled difference to dot I_s
%    at the same input. Show one conditional network call at inference.
% 2. Local geometry: a circle in R^3 with a shaded tube U (codimension two).
%    Enlarge the endpoint chord: tangent O(r), normal O(r^2).
%    Show trajectories only inside U; rescale current arrows by h_c.
% 3. Two modes: use the same Gaussian source points in two panels, colored
%    by ordinary versus constructed guided endpoints. Show the moved
%    separating boundary and the calculated source probability of each basin.
%    Label the guided panel as ONE self-consistent solution, not a uniquely
%    selected solution or a consequence proved for neural-network training.

\section{Experiments}
\label{sec:experiments}

We evaluate SBM on class-conditional ImageNet $256\times256$
(Section~\ref{sec:imagenet}), text-to-image generation
(Section~\ref{sec:t2i}) and action-conditioned world modeling
(Section~\ref{sec:worldmodel}), and ablate the training recipe on ImageNet
(Section~\ref{sec:ablations}). Unless stated otherwise, every SBM model is
trained from scratch with the full recipe of
Algorithm~\ref{algo:training_complicated} and samples with a single network
evaluation. Training compute is
counted in FLOPs ($2\times$MACs) over the full run, including a method's
feature network where it trains through one (Appendix~\ref{app:in-compute}).

\subsection{ImageNet}
\label{sec:imagenet}

Our first experiments are on class-conditional ImageNet $256\times 256$, with FID-50k against the ADM training-set reference~\citep{dhariwal2021diffusionmodelsbeatgans}. We compare SBM with native BTM under identical conditions (Section~\ref{subsec:btm_comp}), and then with state-of-the-art one-step generators, both those trained without an auxiliary pretrained feature networks and those trained with them (Section~\ref{subsec:baselines}).

\subsubsection{Comparison against native BTM}
\label{subsec:btm_comp}

\begin{figure}[ht]
    \centering
    \includegraphics[width=1\linewidth]{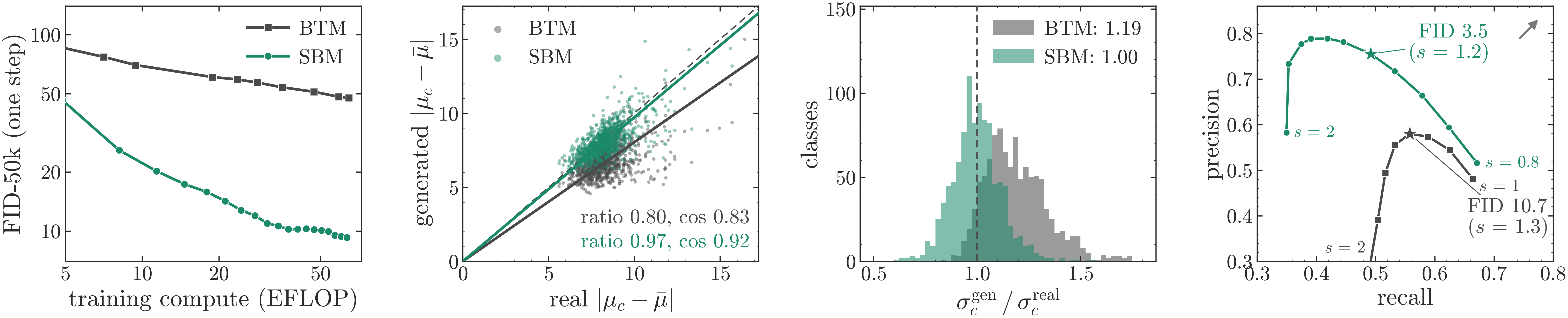}
    \caption{\textbf{Comparison with native BTM in the SD-VAE latent.} \textbf{Left:} one-step FID during pretraining at base size (DiT-B/2), both trained from scratch. The other three panels compare converged DiT-XL/2 maps: the BTM checkpoint and the same checkpoint finetuned with the SBM recipe. \textbf{Center left:} per class, the distance of the generated class mean from the generated grand mean against the same distance for real images (Inception space); SBM separates classes about as much as the data, BTM less. \textbf{Center right:} ratio of generated to real within-class standard deviation; SBM's mean ratio is 1.00, while BTM's classes are wider than the real ones (1.19), suggesting many samples lie off the data manifold of their class. \textbf{Right:} precision against recall as the read-time label scale $\kappa$ varies.}
    \label{fig:btm_comparison}
\end{figure}

By native BTM we mean BTM trained with the recipe of \citet{lee2026btm}. Figure~\ref{fig:btm_comparison} compares it with SBM under identical conditions: the same transformer, the same SD-VAE latent, the same data, the same single network call at sampling time. Only the training recipe differs. The left panel trains both from scratch at base size (DiT-B/2). SBM reads at ten epochs what BTM has not reached after a hundred, and the two end at 9.3 and 51 after the fixed budget of 100 epochs. An SBM update costs 1.6 BTM updates (11.05 against 7.05 forwards per sample, Appendix~\ref{app:in-compute}), so on the compute axis the gap shrinks by that factor and no more.

The two middle panels show where the difference lives. We take two converged DiT-XL/2 maps (the BTM checkpoint and the same checkpoint finetuned with the SBM recipe), draw 50 samples per class, and comparing the distribution of generated samples with that of the validation set in Inception space. The center-left panel plots the distance of the generated class mean from the generated grand mean against the same distance for real images. BTM's class means sit too close together (slope 0.80) and point in wrong directions (cosine 0.83 between the generated and the real class-mean configuration). SBM's lie on the identity (0.97) with cosine 0.92. The center-right panel plots the ratio of generated to real within-class standard deviation. BTM's classes are too wide (mean ratio 1.19), SBM's are matched (1.00). The plain map under-separates the classes and over-disperses within them. The recipe corrects both at once, consistent with its lower FID.

At read time we can sharpen the class signal without a second network call: with $e_\emptyset$ the
null-class embedding the map was trained on (label dropout $0.1$), every class embedding is extrapolated
to $e_c \leftarrow e_\emptyset + \kappa\,(e_c - e_\emptyset)$, so $\kappa = 1$ is the trained map and $\kappa > 1$
plays the role of a classifier-free-guidance scale inside one evaluation. The right panel plots precision against recall for the same two maps as the read-time label scale $\kappa$ moves from 0.8 to 2. Raising $\kappa$ trades recall for precision. SBM's curve lies above BTM's at every recall, so the improvement is a shift of the trade-off curve rather than a different point on it. The best SBM read, FID 3.5 at $\kappa = 1.2$, is three times lower than the best BTM read of the same network.

\begin{figure}[ht]
    \centering
    \includegraphics[width=\linewidth]{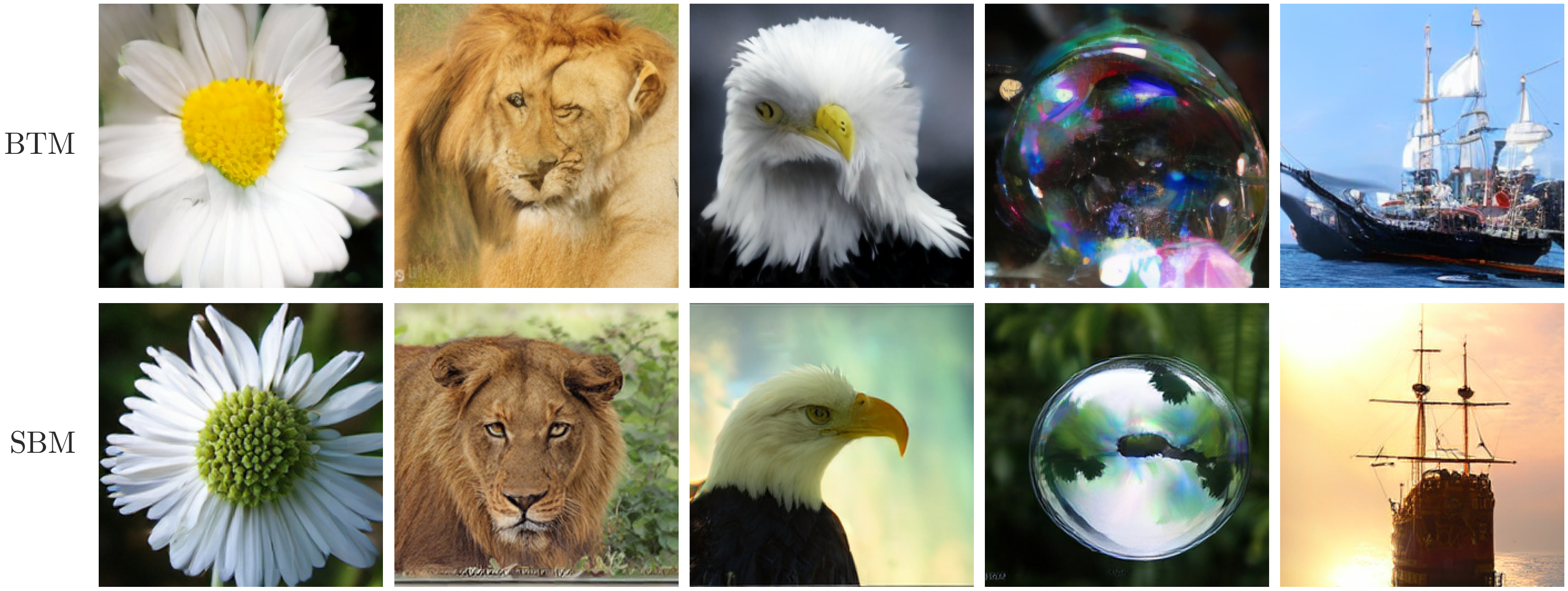}
    \caption{\textbf{Qualitative comparison between BTM and SBM on ImageNet}. For each of the 5 classes daisy, lion, bald eagle, bubble, pirate ship, select the best-looking out of 50 samples for each model.}
    \label{fig:btm_qual}
\end{figure}

Figure~\ref{fig:btm_qual} shows curated samples: for each model and each of five classes (\textit{daisy, lion, bald eagle, bubble, pirate ship}) we draw 50 samples and keep the best-looking one. This is not a rigorous comparison, but it gives a sense of the difference in fidelity.

\subsubsection{Comparison against baselines}
\label{subsec:baselines}

\paragraph{Setup.}
We train class-conditional one-step maps on ImageNet in both the SD-VAE latent space and of a representation autoencoder (RAE)~\citep{zheng2025diffusiontransformersrepresentationautoencoders}, a frozen DINOv2 encoder with a trained decoder, using the RAE paper's DiT$^{\mathrm{DH}}$ transformer at
base and XL size (SBM-B, SBM-XL). Both are trained from scratch at batch 256 with the full recipe. Besides FID-50k we report the FDr6 of~\citet{fdloss}, the mean of six
Fr\'echet distances (Inception, ConvNeXt-v2, DINOv2-L, MAE-L, SigLIP2, CLIP-L), each divided by its value for the ImageNet validation set, so
that the validation set scores 1. Appendix~\ref{app:imagenet} lists the training and evaluation details.

\paragraph{Baselines.}
We compare with three families. One-step diffusion and flow models trained from scratch without an auxiliary pretrained feature loss: iMF~\citep{geng2025improvedmeanflow},
MeanFlow~\citep{geng2025meanflow} and Shortcut models~\citep{frans2025shortcut}. One-step models trained with features of a pretrained encoder:
Drifting~\citep{deng2026drifting} and W-Flow~\citep{wflow} which both uses MAE features, pMF~\citep{pmf} which use a perceptual loss. And multi-step diffusion transformers as a reference. For the pretraining comparison we ran the released code of iMF, MeanFlow, Shortcut and latent Drifting at base size on the same data and scored every checkpoint with the same FID implementation; the final-quality comparisons use the
published numbers. For these curves we ran Drifting with its MAE-256 feature network, which trains markedly faster per FLOP than the MAE-640 behind its published numbers.

\begin{figure}
    \centering
    \includegraphics[width=1\linewidth]{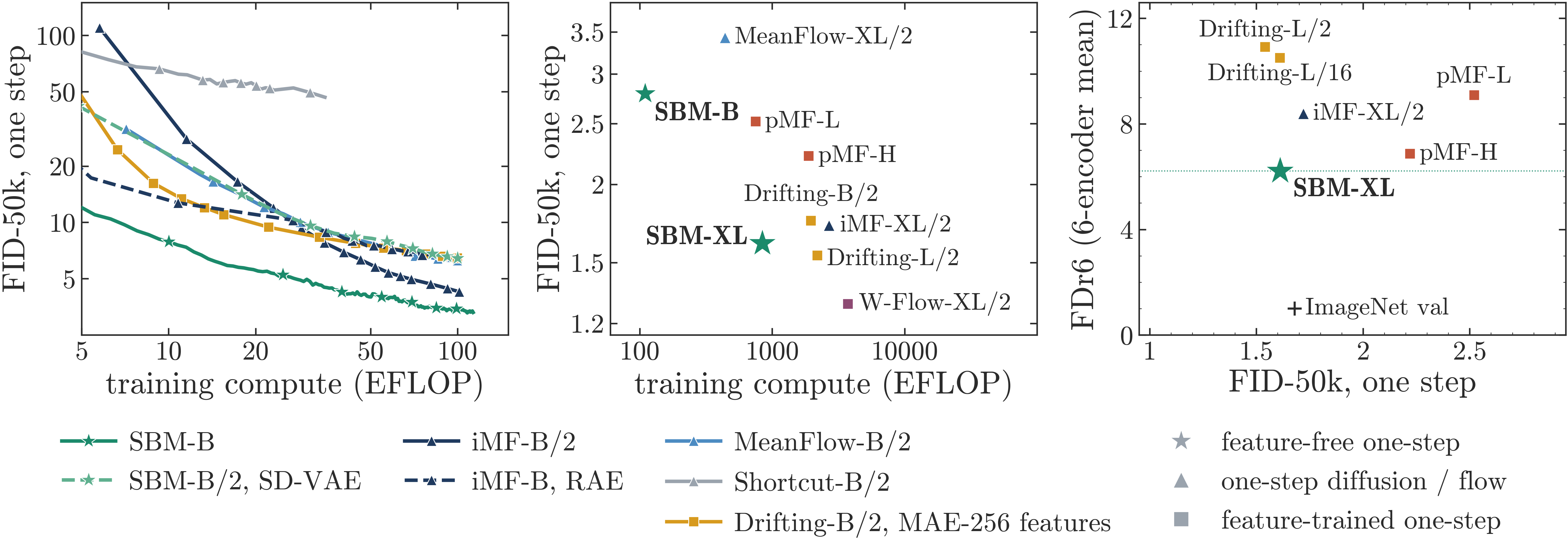}
    \caption{\textbf{ImageNet $256\times256$ against one-step baselines.} \textbf{Left:} one-step FID during pretraining at base size against training compute; SBM-B is in the RAE latent, the baselines run from their released code. The dashed green curve shows SBM-B/2 trained from scratch in SD-VAE latents; the dashed navy curve is iMF's recipe trained on SBM-B's network in the RAE latent (Appendix~\ref{app:in-imf-rae}). \textbf{Center:} final one-step FID against training compute; SBM-B and SBM-XL lie on the compute--FID Pareto front. \textbf{Right:} FDr6, the mean of six normalized Fr\'echet distances, against FID, with the baselines as measured by \citet{fdloss} on the released checkpoints; the dotted line marks SBM-XL.}
    \label{fig:imagenet_baselines}
\end{figure}

\paragraph{Fast training and quality at scale.}
The left panel of Figure~\ref{fig:imagenet_baselines} compares training curves at base size. SBM-B is ahead of every baseline at every epoch and every FLOP
budget. The advantage persists when architecture and latent representation are held fixed: training iMF's recipe on the same network in the same RAE latent gives substantially higher FID. The dashed navy curve in the left panel reaches 6.6 after 150 epochs, twice SBM-B's FID, and trails iMF's own SD-VAE run from 50 epochs on (Appendix~\ref{app:in-imf-rae}). SBM-XL in the RAE latent passes the FID of 17.6 that the original BTM-XL/2 reached after 1{,}280
epochs~\citep{lee2026btm} within 3.5 epochs, and SBM-B within seven. The center panel of Figure~\ref{fig:imagenet_baselines} places the final checkpoints by training compute and Table~\ref{tab:imagenet} gives the numbers. SBM-XL reaches a one-step FID of 1.6 after 840~EFLOP. Among one-step models trained without auxiliary pretrained features this is the
lowest FID we are aware of at any budget: iMF-XL/2 reports 1.72 at 2{,}700~EFLOP by our count, 3.2 times the training compute of SBM-XL, and MeanFlow-XL/2 3.43. Models that train
with pretrained features reach lower FID with more compute, Drifting-L/2 1.54 at 2{,}200~EFLOP and W-Flow-XL/2 1.29 at
3{,}700~EFLOP. Importantly, this is excluding the FLOPs required to pretrain the 2.1B feature networks. Both also need many samples of each class in every update to estimate their feature-space field: Drifting and W-Flow train on 8{,}192 generated and 12{,}288 to 20{,}480 real images per update, and Drifting's own ablation shows FID improving steadily with more samples per class. SBM trains at batch 256.

\paragraph{Feature-space quality.}
FID measures distributional discrepancy in Inception feature space. We also report FDr6, introduced by \citet{fdloss}, to evaluate discrepancy across six feature spaces. The right panel of Figure~\ref{fig:imagenet_baselines} shows FDr6 against FID, with the baselines' values measured by \citet{fdloss} on the released checkpoints (Table~\ref{tab:fdr6}). SBM-XL reads 6.2, below iMF-XL/2 at 8.4, pMF-H at 6.9 and Drifting-L at 10.5. Its Inception component is 0.99, the level of the validation set, and its ConvNeXt component 1.8, while the four semantic encoders still
measure a 5 to 12 times larger distance: the remaining gap to the data is invisible to FID. Our loss uses none of the six evaluation encoders; pMF trains on ConvNeXt-v2. The RAE latent is built on a DINOv2 encoder, which may favor that one component; without it the mean of the other five is 6.4. While RAE provides the strongest performance per training FLOP, the recipe also works in SD-VAE latents. Figure 6 includes SBM-B/2 trained from scratch in SD-VAE space, showing that the pretraining improvements extend to both latent representations with similar speed as MeanFlow and Drifting. At XL size, fine-tuning the released SD-VAE BTM checkpoint with our recipe reaches an FID of 3.5 (Table \ref{tab:imagenet}).

\begin{table}[!htbp]
  \centering
  \caption{\textbf{System-level comparison on ImageNet $256\times256$.} Params: generator + autoencoder decoder. Auxiliary: an
  additional network used in training, either a velocity branch or a pretrained feature encoder. Guidance: CFG on velocity
  predictions (at inference for the multistep models, in the training target for the one-step ones), or guidance built from
  data without a velocity (feature-space repulsion in Drifting, the conditional--null map correction in SBM).
  RAE+DiT$^{\mathrm{DH}}$ uses autoguidance~\citep{karras2024autoguidance}: its second evaluation per step is a small,
  briefly trained guide model, not counted in its parameters. SBM-XL/2
  (SD-VAE) is the BTM checkpoint finetuned with the SBM recipe (Section~\ref{subsec:btm_comp}); the RAE models are trained
  from scratch. SBM numbers are one sampling seed; IS on the same 50k samples as the FID (10 splits), except SBM-B, whose IS is our own
  read of the same checkpoint (FID 3.3 under our protocol). The IS of SiT-XL/2 is the value in its repository, that of MeanFlow-XL/2 the measurement of \citet{geng2025improvedmeanflow};
  Shortcut reports none.}
  \label{tab:imagenet}
  \resizebox{\linewidth}{!}{%
  \begin{tabular}{@{}lll llccc@{}}
  \toprule
  method & space & params & auxiliary & guidance & NFE & FID $\downarrow$ & IS $\uparrow$ \\
  \midrule
  \multicolumn{8}{@{}l}{\textit{Multistep Diffusion / Flows}} \\
  SiT-XL/2 \citep{ma2024sit} & SD-VAE & 675M+49M & -- & -- & 250 & 8.30 & -- \\
  SiT-XL/2 \citep{ma2024sit} & SD-VAE & 675M+49M & -- & Velocity & $2\times250$ & 2.06 & 270.3 \\
  RAE+DiT$^{\mathrm{DH}}$-XL/2 \citep{zheng2025diffusiontransformersrepresentationautoencoders} & RAE & 839M+415M & -- & Velocity & $2\times50$ & 1.13 & 262.6 \\
  \midrule
  \multicolumn{8}{@{}l}{\textit{One-step Diffusion / Flows}} \\
  Shortcut-XL/2 \citep{frans2025shortcut} & SD-VAE & 675M+49M & -- & Velocity & 1 & 10.60 & -- \\
  MeanFlow-XL/2 \citep{geng2025meanflow} & SD-VAE & 676M+49M & -- & Velocity & 1 & 3.43 & 247.5 \\
  iMF-XL/2 \citep{geng2025improvedmeanflow} & SD-VAE & 610M+49M & Velocity (101M) & Velocity & 1 & 1.72 & 282.0 \\
  pMF-H/16 \citep{pmf} & pixel & 956M & LPIPS + ConvNeXt-V2 & Velocity & 1 & 2.22 & 268.8 \\
  \midrule
  \multicolumn{8}{@{}l}{\textit{One-step Pushforward}} \\
  Drifting-L/2 \citep{deng2026drifting} & SD-VAE & 463M+49M & MAE-640 (2.13B) & Data & 1 & 1.54 & 258.9 \\
  % W-Flow XL/2 & SD-VAE & 679M+49M & MAE-640 (2.13B) & Data & 1 & 1.29 & 265.4 \\
  BTM-XL/2 \citep{lee2026btm} & SD-VAE & -- & -- & -- & 1 & 17.58 & -- \\
  \rowcolor{rowhl}
  \textbf{SBM-XL/2 (SD-VAE)} & SD-VAE & 675M+49M & -- & Data & 1 & \textbf{3.5} & 266.0 \\
  \rowcolor{rowhl}
  \textbf{SBM-B (RAE)} & RAE & 129M+415M & -- & Data & 1 & \textbf{2.79} & 200.0 \\
  \rowcolor{rowhl}
  \textbf{SBM-XL (RAE)} & RAE & 669M+415M & -- & Data & 1 & \textbf{1.61} & 256.5 \\
  \bottomrule
  \end{tabular}%
  }
\end{table}

\subsection{Text-to-Image}
\label{sec:t2i}

\paragraph{Setup.} We train SBM-B and SBM-XL from scratch on the training data of MiniT2I~\citep{minit2i2026},
with its schedule and its frozen FLAN-T5-Large text encoder, and evaluate GenEval with one map evaluation and one
decoder call per image. The maps act in the latent of a frozen DINOv2-S encoder and are decoded at 256\,px, whereas
the baselines are pixel-space systems at 512\,px, with training compute as estimated by the MiniT2I authors
and inference compute by our own count (Appendix~\ref{app:t2i-compute}).
Appendix~\ref{app:t2i} gives the architecture and training details.

\begin{figure}[t]
 \centering
 \includegraphics[width=\linewidth]{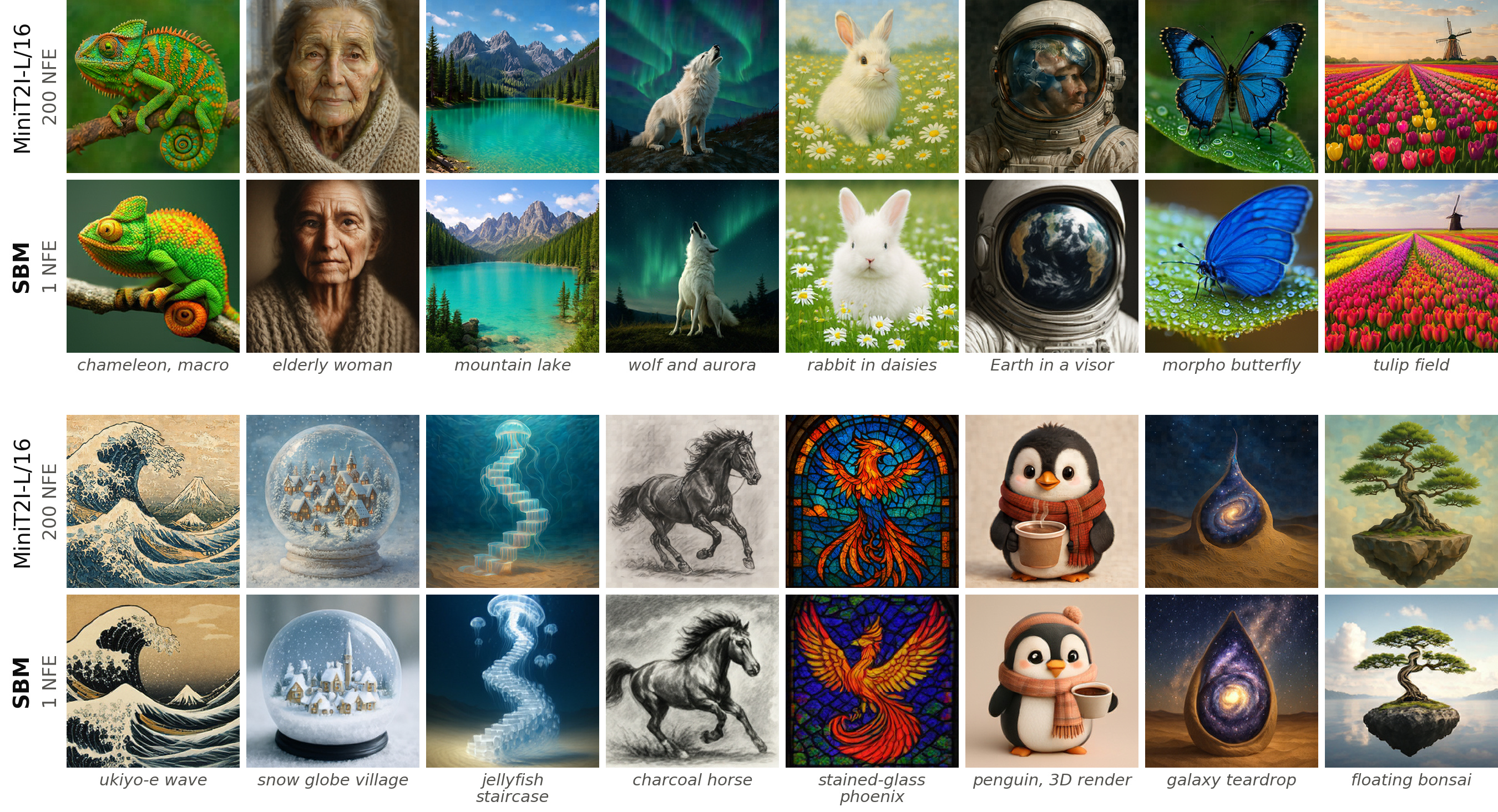}
 \caption{\textbf{SBM-XL against MiniT2I-L/16 on the same prompts}, one draw per prompt. MiniT2I-L/16 is a flow matching model sampled in 100 steps with classifier-free guidance, 200 network evaluations per image; SBM-XL is a one-step map trained from scratch on the same data and reaches similar quality in one.}
 \label{fig:sbm examples}
\end{figure}

\paragraph{Results.} With a single network evaluation, SBM comes within a few GenEval points of multi-step flow
models trained on the same data, at less than a hundredth of their sampling cost (Table~\ref{tab:t2i_systems},
Figure~\ref{fig:t2i}). SBM-XL reaches 0.856 and SBM-B 0.833, at 0.57 and 0.33\,TFLOP per image. At matched training
compute, this is 0.03 to 0.04 below the 100-step MiniT2I-B/16 and L/16 models, which use 200 network
evaluations per image and 140 and 470 times the inference compute of SBM-XL. The four-step MeanFlow distillation of MiniT2I-B/16 scores 0.842, between our two models, but costs 2.8 times the inference compute of SBM-XL and 4.9 times that of SBM-B, and needs a separate distillation stage. Both SBM models also score above the other pixel-space
systems except DeCo. Figure~\ref{fig:sbm examples} shows SBM-XL and MiniT2I-L/16 on the same prompts.

% Use the table float queue so Table 2 cannot overtake Table 1.
\begin{table}[!htbp]
\centering
\captionsetup{justification=raggedright,singlelinecheck=false}% only affects this float
% ---------------- Table ----------------
\begin{minipage}[t]{0.36\textwidth}
  \vspace{0pt}% top-align the panels
  \centering
  \resizebox{\linewidth}{!}{%
  {\footnotesize
   \setlength{\tabcolsep}{3pt}%
   \begin{tabular}{@{}l@{}rrr@{}}
   \toprule
   Model & \begin{tabular}[b]{@{}r@{}}Compute\\(ZFLOP)\,$\downarrow$\end{tabular}
         & NFE\,$\downarrow$ & GenEval\,$\uparrow$ \\
   \midrule
   MiniT2I-B/32     & 0.13 & 200 & 0.826 \\
   MiniT2I-B/16     & 0.36 & 200 & 0.873 \\
   MiniT2I-L/16     & 1.19 & 200 & 0.883 \\
   DeCo-XXL/16      & 0.80 &  50 & 0.86  \\
   PixelGen-XXL/16  & 0.80 &  50 & 0.79  \\
   PixNerd-XXL/16   & 0.73 &  50 & 0.73  \\
   PixelDiT-T2I     & 3.17 &  50 & 0.78  \\
   MiniT2I-B/16-MF  & 0.49\rlap{$^\dagger$} & 4 & 0.842 \\
   \midrule
   \rowcolor{rowhl}
   \textbf{SBM-B}     & 0.35 &   1 & 0.833 \\
   \rowcolor{rowhl}
   \textbf{SBM-XL}    & 1.15 &   1 & 0.856 \\
   \bottomrule
   \end{tabular}}}
  \captionof{table}{GenEval against training compute and network evaluations per image (NFE, incl.\ CFG). Baselines as estimated by \citet{minit2i2026}; $^\dagger$teacher plus our distillation estimate. Ours: one map evaluation and one decoder call; accounting in Appendix~\ref{app:t2i-compute}.}
  \label{tab:t2i_systems}
\end{minipage}\hspace{0.03\textwidth}%
% ---------------- Figure ----------------
\begin{minipage}[t]{0.6\textwidth}
  \vspace{0pt}
  \centering
  \includegraphics[width=\linewidth]{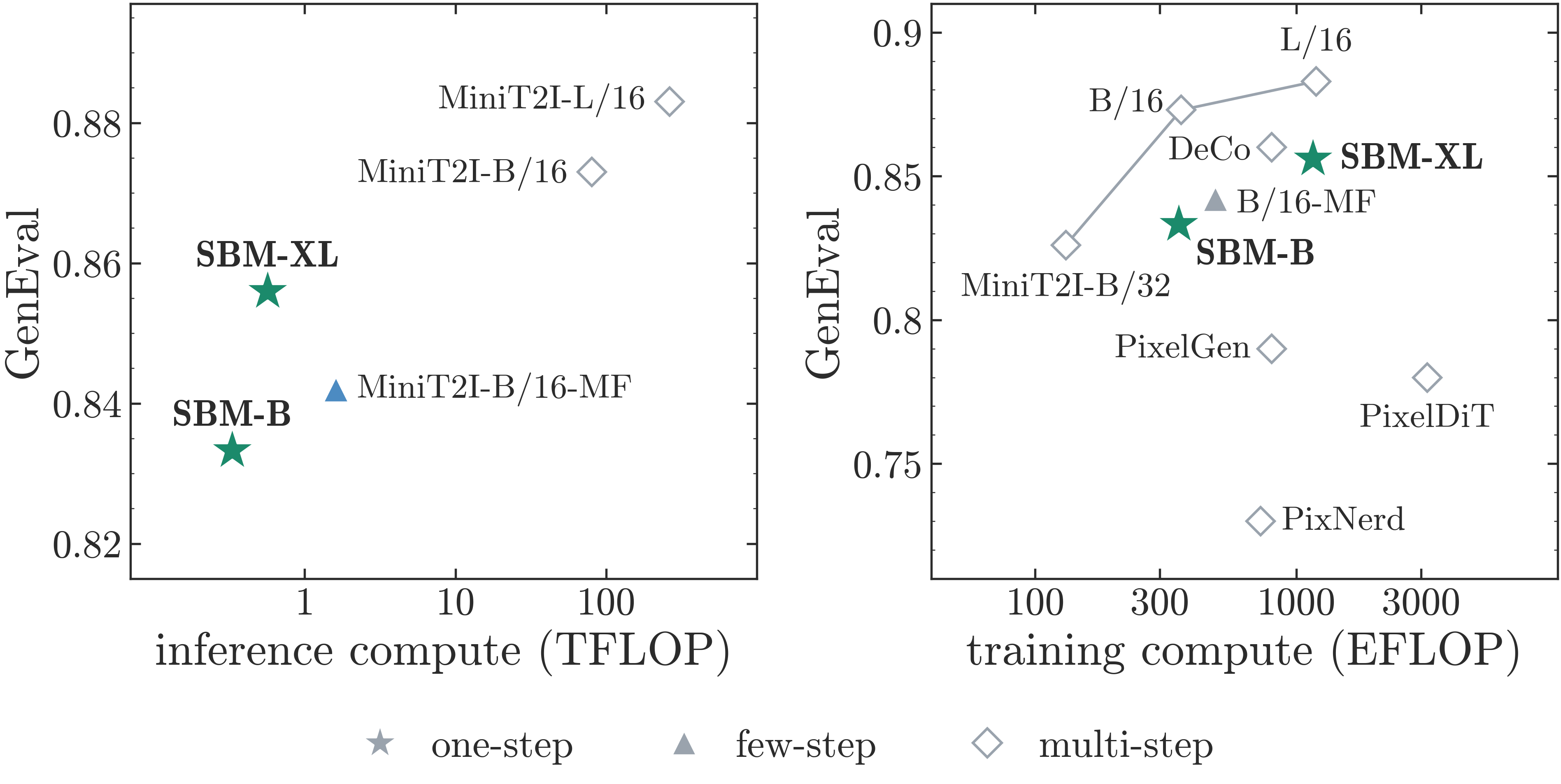}
  \captionof{figure}{GenEval against compute. SBM is the only one-step method, B/16-MF (4-steps) is the only other few-step method, while the rest uses at least 50 NFEs, as reported in Table \ref{tab:t2i_systems}. \textbf{Left:} against inference compute per image.
  \textbf{Right:} against training compute.}
  \label{fig:t2i}
\end{minipage}
\end{table}

\FloatBarrier
\subsection{World Modeling}
\label{sec:worldmodel}

\begin{figure}[!htbp]
\centering
\captionsetup{justification=raggedright,singlelinecheck=false,skip=5pt}
% Proportional widths give both crops the same height, without a gap under RT-1.
\begin{minipage}[t]{0.405\linewidth}
  \vspace{0pt}
  \includegraphics[width=\linewidth,trim=0 925bp 0 0,clip]{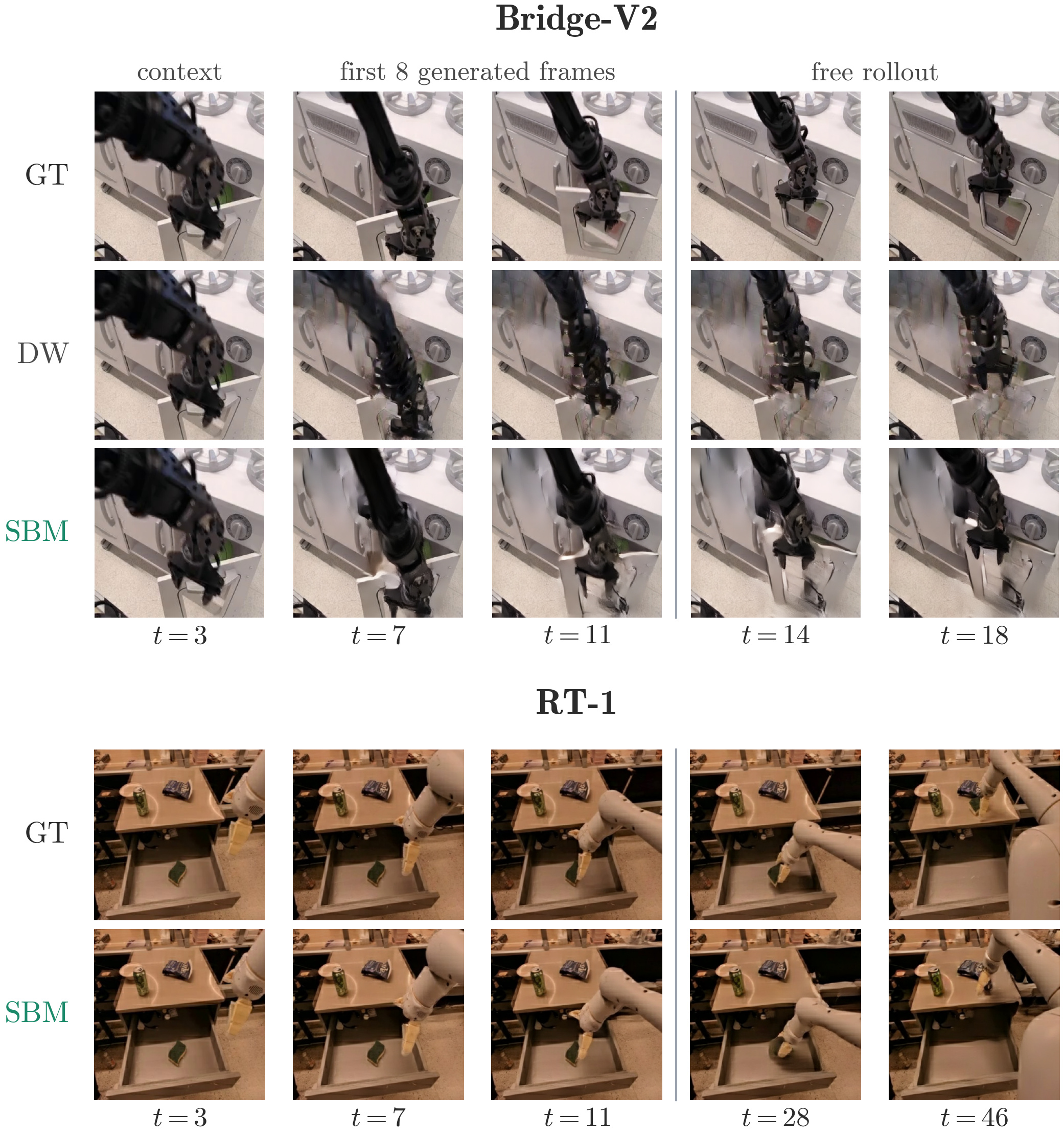}
\end{minipage}\hfill
\begin{minipage}[t]{0.58\linewidth}
  \vspace{0pt}
  \includegraphics[width=\linewidth,trim=0 0 0 1310bp,clip]{wm.jpg}
\end{minipage}
  \caption{Rollouts on Bridge-V2 (left, SBM against DriftWorld's (DW) released Bridge-V2 checkpoint) and RT-1 (right, SBM only).
The first column is the last context frame. Left of the divider are the eight scored frames;
right of it, the rollout continues from the model's own frames without resetting to ground
truth (GT). DW's arm dissolves; SBM keeps the arm and scene intact.}
  \label{fig:wm}
\end{figure}

\begin{table}[!htbp]
  \centering
  \small
  \captionsetup{skip=5pt}
  \setlength{\tabcolsep}{5pt}
\begin{tabularx}{\linewidth}{@{}>{\raggedright\arraybackslash}X*{7}{>{\centering\arraybackslash}X}@{}}
\toprule
Model & EFLOP & SSIM $\uparrow$ & PSNR $\uparrow$ & LPIPS $\downarrow$ & FID $\downarrow$ & FVD $\downarrow$ & NFE \\
\midrule
IRASim  & -- & 0.738 & 18.84 & 0.158 & 10.46 & 91.55 & 59 \\
IWS  & -- & 0.786 & \underline{22.85} & 0.169 & 45.30 & 329.75 & 4 \\
LVDM  & -- & 0.741 & 19.19 & 0.154 & 9.86 & 103.51 & 59 \\
VDM  & -- & 0.725 & 18.42 & 0.167 & 39.88 & 159.16 & 59 \\
DriftWorld  & $\geq$30.4 & \underline{0.835} & 22.69 & \underline{0.091} & \underline{4.86} & \underline{46.05} & 1 \\
\midrule
\rowcolor{rowhl}
\textbf{SBM} &  15.6  & \textbf{0.869} & \textbf{24.80} & \textbf{0.060} & \textbf{4.25} & \textbf{18.1} & 1 \\
\bottomrule
\end{tabularx}
  \caption{Bridge-V2 next-frame prediction under DriftWorld's 8-frame autoregressive protocol. Baseline metrics are reported from DriftWorld. Compute is training compute in EFLOP; DriftWorld's is a lower bound (Appendix~\ref{app:wm-compute}). NFE is the number of network function evaluations per generation call; VAE encoding and decoding are not counted. DriftWorld and SBM generate each frame in a single forward pass. For the other baselines we use the defaults in their released code: IRASim, LVDM and VDM run 50 PNDM steps (59 evaluations, including PNDM's Runge--Kutta warm-up) to produce one 16-frame clip, and IWS uses 1 latent-dynamics and 3 decoder evaluations per frame. Dashes indicate unavailable information. Best reported values are in bold; second-best values are underlined.}
  \label{tab:bridge_driftworld}

\end{table}

\begin{table}[t]
  \centering
  \small
  \setlength{\tabcolsep}{4pt}
\begin{tabularx}{\linewidth}{@{}>{\raggedright\arraybackslash}p{0.19\linewidth}*{8}{>{\centering\arraybackslash}X}@{}}
\toprule
Model & EFLOP & SSIM~$\uparrow$ & PSNR~$\uparrow$ & LPIPS~$\downarrow$ & FID~$\downarrow$ & FVD~$\downarrow$ & NFE & s/frame~$\downarrow$ \\
\midrule
IRASim & -- & 0.757 & 21.05 & 0.151 & \underline{4.60} & 93.0 & 59 &1.104 \\
IWS & -- & 0.807 & 24.29 & 0.167 & 46.81 & 448.2 & 4 & 0.023 \\
LVDM & -- & 0.750 & 21.00 & 0.153 & 5.40 & 94.6 & 59 &0.284 \\
VDM & -- & 0.540 & 13.82 & 0.246 & 31.68 & 406.1 & 59 &3.153 \\
DriftWorld & $\geq$16.1 & 0.841 & 24.20 & 0.097 & 8.99 & 62.8 & 1 & 0.019 \\
\midrule
\rowcolor{rowhl}
\textbf{SBM} (short run) & 5.7 & \underline{0.864} & \underline{25.61} & \underline{0.071} & 7.70 & \underline{28.3} & 1 & 0.021 \\
\rowcolor{rowhl}
\textbf{SBM} (long run) & 43.4 & \textbf{0.880} & \textbf{26.90} & \textbf{0.057} & \textbf{4.02} & \textbf{16.1} & 1 & 0.021 \\
\bottomrule
\end{tabularx}
  \caption{RT-1 world modeling (8-frame autoregressive generation, 4 conditioning frames, $256\times256$, validation split). Baseline rows are taken from DriftWorld, Table~IV. SBM uses one-step generation without an auxiliary feature loss or guidance (Appendix~\ref{app:wm-compute}). The short and long runs use 5.7 and 43.4 EFLOP, respectively. We use DriftWorld's Bridge-V2 U-Net architecture. NFE is defined as in Table~\ref{tab:bridge_driftworld}. Inference time is per generated frame on one GPU, the baselines as reported by DriftWorld, SBM on one H200 at batch 1 in bf16 including the SD3 decoder (14.2 ms for the map, 6.6 ms for the decoder). Best quality values are in bold; second-best distinct values are underlined.}
  \label{tab:rt1}
\end{table}

\paragraph{Setup and baselines.}
We evaluate one-step maps as action-conditioned world models on BridgeData V2~\citep{bridge2} (Bridge-V2 below) and RT-1~\citep{rt1},
under the protocol of DriftWorld~\citep{driftworld}. All reported SBM world models use $\omega=1$ and no auxiliary feature loss. Write $o_k$ for the latent of frame $k$ and $a_k$ for the action taken at step $k$. Given four frames and their actions, a world model samples the next frame from $p(o_{k+1}\mid o_{k-3:k},\,a_{k-3:k})$, and SBM draws this sample with one network evaluation, $\hat o_{k+1}=T_\theta(x_0,\,c_k,\,\gamma_{\mathrm{read}})$ with $x_0\sim\mu_0$ and $c_k=(o_{k-3:k},\,a_{k-3:k})$.
% CHECK: action indices. DriftWorld conditions on the future action chunk a_{t:t+T-1}; if the network only sees a_k, or all five actions of the window, change both occurrences.
Longer videos are generated autoregressively, each generated frame entering the history of the next step: every validation episode is rolled out in chunks of eight generated frames, with the history reset to the ground truth at the start of each chunk. Frames live in the SD3 autoencoder latent~\citep{esser2024sd3} and the
reference is the autoencoder's reconstruction of the ground truth, as in DriftWorld. We report their five metrics: SSIM, PSNR and LPIPS
per video, FID over all generated frames, and FVD on the first sixteen generated frames of every video. Baseline rows are taken from
DriftWorld's Table~IV; training compute is counted as in Section~\ref{sec:experiments}, and DriftWorld's budget is our count of its released training step (Appendix~\ref{app:wm-compute}). Appendix~\ref{app:worldmodel} gives the data, the training and the evaluation details.

\paragraph{Results.}
Tables~\ref{tab:bridge_driftworld} and~\ref{tab:rt1} give the numbers. On Bridge-V2, SBM beats DriftWorld on all five metrics at
about half the training compute, and without the DINOv3 feature encoder that DriftWorld's own ablation finds essential (Appendix~\ref{app:wm-featurefree}). On RT-1, where DriftWorld's row is weaker, the SBM map passes it on all five metrics at $2.8\times$ less compute, and the longer trained map at $43.4$~EFLOP is past every published entry of the table on all five metrics.

\paragraph{Temporal consistency.}
The FVD values are $2.5\times$ to $3.9\times$ below DriftWorld's. Figure~\ref{fig:wm} shows where the difference is. Inside the first chunk of eight frames, the regime the protocol scores and the one DriftWorld's qualitative figures show, both models follow the ground truth. Past it, with the model fed its own frames for the rest of the episode, DriftWorld's arm dissolves within
about ten frames, while the SBM rollout keeps the arm solid and completes the task: the oven door is closed, the sponge is lifted out of the drawer. Appendix~\ref{app:wm-qual} shows more episodes of both kinds.

% Preamble:
% \usepackage{booktabs, graphicx, colortbl, xcolor, caption}
% \definecolor{rowhl}{gray}{0.92}

% World-model rollouts and quantitative tables use separate full-width floats.

\subsection{Ablations}
\label{sec:ablations}

\begin{figure}[H]
\centering
\captionsetup{justification=raggedright,singlelinecheck=false}
  \includegraphics[width=\linewidth]{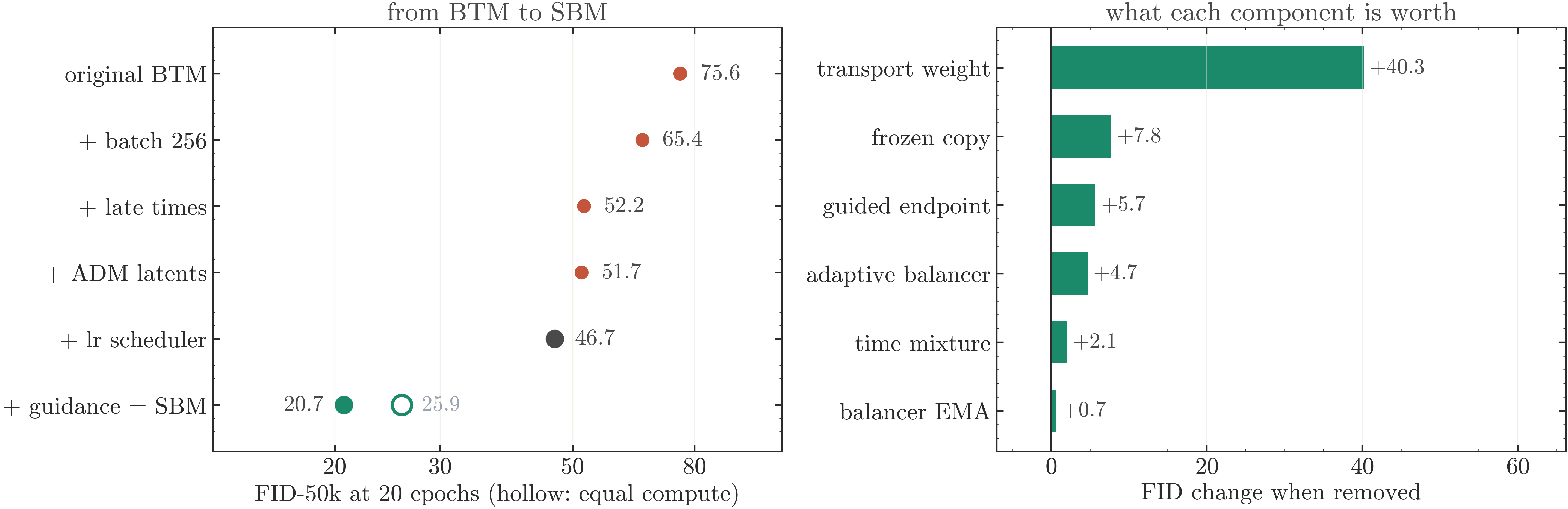}
  \caption{\textbf{From BTM to SBM.} One-step FID-50k after 20 epochs for DiT-B/2 maps trained from scratch in the
SD-VAE latent (one seed). \textbf{Left:} each row adds one change to the row above, from the BTM configuration we
started from to SBM. The hollow marker reads the SBM run at equal compute. \textbf{Right:} change in FID when one component is removed from the full recipe; a positive
value means the component helps. Appendix~\ref{app:ablations} defines every row and component.}
  \label{fig:abl_two}
\end{figure}

% Reserve enough room so the wrapped plot cannot extend past the page.
\Needspace{20\baselineskip}
\begin{wrapfigure}{r}{0.31\textwidth}
  \centering
  \includegraphics[width=\linewidth]{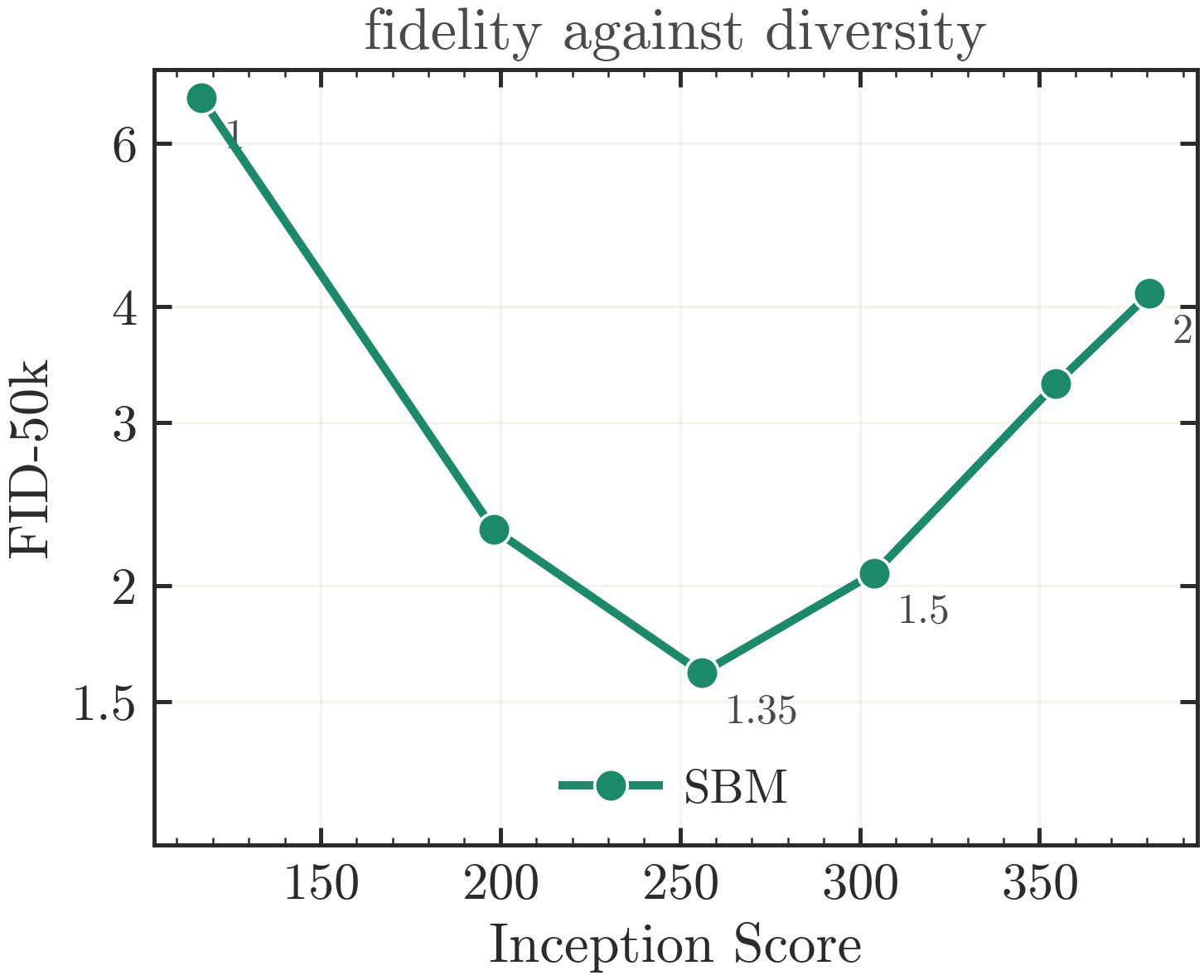}
  \caption{\textbf{Fidelity against diversity.} FID against Inception Score for SBM-XL as the read-time label scale $\kappa$ runs from 1 to 2. FID is lowest, 1.61, at $\kappa = 1.35$.}
  \label{fig:fid-is}
\end{wrapfigure}

Figure~\ref{fig:abl_two} shows what each part of the recipe is worth, at base size and 20 epochs. The BTM configuration we started from reads an FID of 75.6. Four changes that do not
involve guidance bring it to 46.7, the native BTM of Figure~\ref{fig:btm_comparison}: a smaller batch, a late
transport-time distribution, the ADM latent preprocessing and the learning-rate drop of the native
recipe. Guidance is the change that matters most. The guided targets of
Algorithm~\ref{algo:training_complicated}, together with the smoothed gradient balancer, take the map from 46.7 to
20.7 at the same epoch budget, 25.9 on same FLOP budget. The unguided changes that come with the recipe net to zero
(Appendix~\ref{app:ablations}).

Without the per-row normalization of the
transport loss, the map reads worse than native BTM (Figure~\ref{fig:abl_two}, right). Removing any other component costs less than eight points. Figure~\ref{fig:fid-is} traces our best map along the read-time label scale $\kappa$, an inference-time knob that trades fidelity for class-typicality. From $\kappa = 1$ to $2$ the Inception
Score rises, while FID falls to a minimum of 1.61 at $\kappa = 1.35$ before it climbs.
Appendix~\ref{app:abl-tradeoff} traces the ablation maps along the same scale and across the guidance strength $\omega$.

% =====================================================================
\section{Limitations and conclusion}
\label{sec:conclusion}
% =====================================================================

\paragraph{Limitations.}
Our theory characterizes idealized population updates and establishes
local endpoint existence under weak guidance. It does not establish
convergence of the full training recipe or identify the global map
and output distribution selected by optimization. Guidance also adds
network evaluations during training, despite retaining one evaluation
at sampling time. Our strongest image models use
representation autoencoders, so training from scratch refers to the
generator. Text-to-image evaluation focuses on GenEval at
$256\times256$, which does not capture all aspects of visual quality.
World-model metrics use ground-truth history resets every eight
generated frames; longer rollouts are assessed qualitatively, leaving
long-horizon accuracy and downstream control performance untested.

\paragraph{Future work.}
Several directions follow from training the endpoint map directly. We use the
linear interpolant with independent noise--data pairs; other interpolants and
couplings change the autonomous current and may give maps that are easier to
learn. Our analysis leaves open which of the self-consistent maps training
selects, and whether the objective can control that choice. Here the map's own
predictions supply the guidance signal; rewards, physical constraints and
combinations of conditions could steer the map in the same way, without an
auxiliary velocity or score network. The map can also be called more than
once, re-noising its output between calls, a refinement that leaves training
unchanged. For world models, coupling the predictions across frames would
extend the one-step map to long-horizon rollouts.

\paragraph{Conclusion.}
We introduced Scalable Beckmann Maps, combining training-time guidance
with a recipe for training direct generative maps at scale. The
autonomous denoising interpretation of the BTM current provides the
basis for guidance, while the map's own conditional and null predictions
supply the correction without an additional denoiser. Our analysis
characterizes the resulting bias and connects the idealized update to
guided conservation and output reweighting. Together, guidance and
the training recipe substantially improve ImageNet generation and
enable competitive one-step text-to-image generation. For world
modeling, the recipe alone outperforms DriftWorld without guidance or
an auxiliary feature loss. These results demonstrate the practical
potential of learning endpoint maps directly, with sampling remaining
a single network evaluation.

\section*{Acknowledgements}
UU and BK acknowledges support from the NSF through the Harvard University Materials Research Science and Engineering Center Grant No. DMR-2011754 and by the NSF Harnessing the Data Revolution Big Idea under Grant No. 2118201. UU is supported by Aker Scholarship, SW is supported by Kempner Graduate Fellowship. MSA and SW are also supported by the National Science Foundation under Cooperative Agreement PHY-2525568 (The NSF AI Institute for Artificial Intelligence and Fundamental Interactions, http://iaifi.org/). This work has been made possible in part by a gift from the Chan Zuckerberg Initiative Foundation to establish the Kempner Institute for the Study of Natural and Artificial Intelligence.

% =====================================================================
% Statements (check the ICLR 2027 guidelines on whether they count
% toward the 9-page limit; the AI use statement does not)
% =====================================================================

\section*{Ethics statement}
This work studies the optimization of a generative model on ImageNet, a text-to-image dataset, BridgeData V2 and RT-1, and introduces no new data collection and no human subjects. It
inherits the distributional biases of these datasets and the dual-use profile of
image generators and world models, and it neither adds to nor mitigates either.

\section*{Reproducibility statement}
Section~\ref{sec:main} specifies the losses and sampler.
Appendices~\ref{proof:vanishing} and~\ref{app:local} give the population
update and local existence proofs, and Appendix~\ref{app:transport}
states the transport identities with their assumptions.
Appendix~\ref{appendix:training} gives the training algorithm, and
Appendix~\ref{app:experiments} the recipes, checkpoint construction and
evaluation protocol of every experiment. Each reported model is one training
run. ImageNet FIDs use one sampling seed; the final SBM-XL checkpoint reads
1.67 on a second seed against 1.61 (Appendix~\ref{app:in-protocol}). GenEval
uses four sampling seeds: the main text reports the best seed and
Appendix~\ref{app:t2i-protocol} the mean and range. The ablations and the
world-model results use one training run and one evaluation pass each.
Source code and model checkpoints are available at the project page, \url{https://tullebulle.github.io/projects/sbm/}.

\section*{AI use statement}
Generative AI tools assisted with implementation of methods in code, supervising experiments designed by humans, and writing a draft of the paper, and to create tables, and writing implementation details. It helped as sparring partner for designing the toy example accompanying Theorem~\ref{thm:binary}. The idea and methods were developed by humans. The authors are responsible for verifying the final arguments, implementation and reported
results.

\bibliographystyle{plainnat}
\bibliography{references}

\beginappendix

\section{Proofs and theoretical analysis}
\label{app:theory}

We first derive the guidance rule and the stopped-update characterization,
then study noisy boundary supervision, local endpoint existence, and
two-mode reweighting. Unless stated otherwise, $x_0\sim\mu_0$,
$x_1\sim\mu_1^c$, and $S\sim U(0,1)$ are independent,
$I_s=(1-s)x_0+sx_1$, and the null map is fixed when varying the
conditional prediction. Conditional identities hold almost everywhere
under the relevant input law; weighted means require positive finite
normalizers and the stated moments.

\subsection{Autonomous denoising and implicit guidance}
\label{app:autonomous-guidance}

\subsubsection{The current and the weighted posterior mean}
\label{app:weighted-posterior}

\begin{proposition}[Current--denoiser identity]
\label{prop:app-current-denoiser}
Suppose $\E_c\|\dot I_S\|<\infty$. With
$\nu_c(x)=\int_0^1\rho_s^c(x)\,ds$, define
\begin{equation}
 h_c(x)=\int_0^1\frac{\rho_s^c(x)}{1-s}\,ds,
 \qquad
 \bar z_c(x)=
 \frac{\E_c[x_1/(1-S)\mid I_S=x]}
      {\E_c[(1-S)^{-1}\mid I_S=x]}.
 \label{eq:app-weighted-mean}
\end{equation}
Where these quantities are well defined,
\begin{equation}
 J_c(x)=\nu_c(x)\E_c[\dot I_S\mid I_S=x]
       =h_c(x)(\bar z_c(x)-x).
 \label{eq:app-denoiser-current}
\end{equation}
\end{proposition}

\begin{proof}
Conditioning $\E_c[\varphi(I_S)\cdot\dot I_S]$ on $I_S$
identifies the current. Since
$\mathbb P_c(S\in ds\mid I_S=x)=\rho_s^c(x)\,ds/\nu_c(x)$
and $\dot I_S=(x_1-I_S)/(1-S)$,
\[
 \E_c[\dot I_S\mid I_S=x]
 =\E_c\left[\frac{x_1-x}{1-S}\,\middle|\,I_S=x\right]
 =\frac{h_c(x)}{\nu_c(x)}(\bar z_c(x)-x).
\]
Multiplication by $\nu_c$ proves the identity.
\end{proof}

Thus $J_c$, $J_c/\nu_c$, and $\bar z_c-\id$ have the same
oriented trajectories wherever their scalar factors are positive and
finite and their ODEs are locally well posed. Equivalently,
\[
 \bar z_c(x)=\frac{1}{h_c(x)}
 \int_0^1\frac{\rho_s^c(x)m_{s,c}(x)}{1-s}\,ds,
 \qquad m_{s,c}(x)=\E_c[x_1\mid I_s=x].
\]
This is a weighted posterior mean, and the pointwise minimizer of
$\E_c[\|d-x_1\|^2/(1-S)\mid I_S=x]$ when its second moment is
finite. The integrated loss need not be finite because the time weight
has infinite total mass.

\paragraph{Logarithmic clock.}
For $s=1-e^{-\tau}$, $d\tau=ds/(1-s)$ and
$dI_s/d\tau=x_1-I_s$. This explains the weight in $h_c$ and
$\bar z_c$. The time-dependent field in this clock is
$m_{s(\tau),c}-\id$; the transport interpretation of the autonomous
current additionally uses the BTM arguments of \citet{lee2026btm}.

\subsubsection{Implicit guidance, variance, and substitution error}
\label{app:implicit-cfg}

Autonomous denoiser CFG has the equivalent explicit and implicit forms
\begin{equation}
 \bar z_c^\omega
 =\omega\bar z_c-(\omega-1)\bar z_\emptyset
 =\bar z_c+(1-\omega^{-1})(\bar z_c^\omega-\bar z_\emptyset).
 \label{eq:app-implicit-cfg}
\end{equation}
Moving the implicit term to the left gives the explicit form. Substituting
$(T,T_\emptyset)$ for $(\bar z_c^\omega,\bar z_\emptyset)$ in
the correction and conditioning on $I_S$ yields
\begin{equation}
 \begin{aligned}
 J_c^\omega[T]&=J_c+(1-\omega^{-1})h_c(T-T_\emptyset),\\
 \E_c[\dot I_S^\omega\mid I_S=x]&=J_c^\omega[T](x)/\nu_c(x).
 \end{aligned}
 \label{eq:app-guided-current}
\end{equation}
With exact denoisers in the correction, this velocity is
$(h_c/\nu_c)(\bar z_c^\omega-\id)$, a positive rescaling of
the autonomous CFG field. The factor $1-\omega^{-1}$ comes from
implicit CFG and $1/(1-S)$ converts an endpoint displacement into
an interpolant velocity.

\begin{proposition}[Conditional variance of implicit guidance]
\label{prop:guidance-variance}
Fix $T,T_\emptyset$ and assume finite conditional second moments.
At differentiability points of $T$, with
$Y_{\mathrm{tr}}^\omega=T(I_S)+DT(I_S)\dot I_S^\omega$,
\begin{align*}
 \operatorname{Cov}(\dot I_S^\omega\mid I_S,S)
 &=\operatorname{Cov}(\dot I_S\mid I_S,S),\\
 \operatorname{Cov}(Y_{\mathrm{tr}}^\omega\mid I_S,S)
 &=\operatorname{Cov}(Y_{\mathrm{tr}}^1\mid I_S,S),
\end{align*}
where both targets use the same map $T$.
\end{proposition}

\begin{proof}
Conditional on $(I_S,S)$, the correction, $T(I_S)$, and $DT(I_S)$
are deterministic. Adding the correction therefore leaves both
conditional covariances unchanged.
\end{proof}
The explicit direction
$\dot I_S^{\omega,\mathrm{exp}}=\omega\,\dot I_S-(\omega-1)\frac{T_\emptyset(I_S)-I_S}{1-S}$
of Section~\ref{sec:guidance} has the same conditional mean as
\eqref{eq:direction} exactly when $T=\omega\bar z_c-(\omega-1)T_\emptyset$, and,
since $T_\emptyset(I_S)$ is deterministic given $(I_S,S)$,
\[
 \operatorname{Cov}\bigl(\dot I_S^{\omega,\mathrm{exp}}\mid I_S,S\bigr)
 =\omega^2\operatorname{Cov}(\dot I_S\mid I_S,S).
\]
For comparison, independent conditional and null posterior draws at
the same $(x,s)$ give
\[
 \operatorname{Cov}\bigl(\omega x_1^c-(\omega-1)x_1^\emptyset\mid x,s\bigr)
 =\omega^2\operatorname{Cov}(x_1^c\mid x,s)
  +(\omega-1)^2\operatorname{Cov}(x_1^\emptyset\mid x,s).
\]
Implicit guidance avoids this amplification of posterior sampling noise.
The proposition does not assert equality of total minibatch-gradient
variance or explain the empirical gains by variance reduction. Table~\ref{tab:guide-variance} measures the minibatch-gradient variance of both
targets on the B/2 recipe.

Subtracting the exact-denoiser field gives the substitution error
\begin{equation}
 \frac{J_c^\omega[T]}{h_c}-(\bar z_c^\omega-\id)
 =(1-\omega^{-1})\bigl[(T-\bar z_c^\omega)
                         -(T_\emptyset-\bar z_\emptyset)\bigr].
 \label{eq:app-substitution-error}
\end{equation}
For $\omega>1$, the substitution is exact if and only if the two
map--denoiser errors agree. In terms of the instantaneous prediction
of the map-induced field, the same error is
\begin{equation}
 (\omega-1)\Bigl[
 T-\{\bar z_c+(1-\omega^{-1})(T-T_\emptyset)\}
 -(T_\emptyset-\bar z_\emptyset)\Bigr].
 \label{eq:app-own-field-error}
\end{equation}
This is an algebraic identity; neither the current nor the stopped-update
theorem assumes that this error is small. CFG of these normalized
denoising fields need not equal CFG of $J_c/\nu_c$ and
$J_\emptyset/\nu_\emptyset$, whose positive scale factors can differ.
The induced current satisfies
\begin{equation}
 \nabla\cdot J_c^\omega[T]
 =\mu_0-\mu_1^c+
   \nabla\cdot\bigl((1-\omega^{-1})h_c(T-T_\emptyset)\bigr)
 \label{eq:app-guided-divergence}
\end{equation}
when the correction is locally integrable. This identity alone does not
specify a global generated law.

\subsection{Consistency of the stopped update}
\label{proof:vanishing}

We use the unweighted clean losses in
\eqref{eq:transport-loss}--\eqref{eq:loss}. Only the live prediction
varies; all quantities inside $\sg$, including the direction and
$\nabla T$, remain fixed.

\begin{proof}[Proof of Theorem~\ref{thm:vanishing}]
The transport residual is $-\dot I_S^\omega\cdot\nabla T(I_S)$.
Conditioning on $I_S$ gives
\begin{equation}
 \begin{aligned}
 \delta_{\mathrm{sg}}\mathcal L_{\mathrm{tr},c}^\omega(T)[H]
 &=-2\E_c[H(I_S)\cdot(\dot I_S^\omega\cdot\nabla T(I_S))]\\
 &=-2\int H\cdot(J_c^\omega[T]\cdot\nabla T)\,dx.
 \end{aligned}
 \label{eq:app-transport-variation}
\end{equation}
The ideal null trace makes the boundary residual
\[
 T(x_1)-x_1-(1-\omega^{-1})(T-T_\emptyset)(x_1)
 =\omega^{-1}(T(x_1)-x_1).
\]
Consequently,
\begin{equation}
 \delta_{\mathrm{sg}}\mathcal L_c^\omega(T)[H]
 =-2\int H\cdot(J_c^\omega[T]\cdot\nabla T)\,dx
  +\frac{2\lambda}{\omega}\int H\cdot(T-\id)\,d\mu_1^c.
 \label{eq:app-total-variation}
\end{equation}
Both vector measures are finite by the finite-loss assumptions and
Cauchy--Schwarz; for example,
\[
 \int\|J_c^\omega[T]\cdot\nabla T\|\,dx
 \leq\E_c\|\dot I_S^\omega\cdot\nabla T(I_S)\|
 \leq\sqrt{\mathcal L_{\mathrm{tr},c}^\omega(T)}.
\]
The first measure is absolutely continuous with respect to Lebesgue
measure, while the second is supported on the Lebesgue-null set $M$.
If their sum pairs to zero with every smooth compactly supported $H$,
uniqueness of the absolutely continuous and singular parts makes both
vanish. This is exactly \eqref{eq:conservation}. Conversely,
\eqref{eq:conservation} makes the variation zero for every $H$.
\end{proof}

The ideal null trace holds on each class support because
$\mu_1^\emptyset=\sum_{c'}\pi_{c'}\mu_1^{c'}$ with every
$\pi_{c'}>0$. Under this trace the numerical boundary loss scales
as $\omega^{-2}$, but its stopped update scales as $\omega^{-1}$.
The theorem concerns arbitrary output variations; zero update in a
parameterized network need not imply its converse.

\paragraph{Sampled loss.}
Writing $b_c^\omega[T]=\E_c[\dot I_S^\omega\mid I_S=\cdot]$,
conditional orthogonality gives
\begin{equation}
 \begin{aligned}
 \mathcal L_{\mathrm{tr},c}^\omega(T)
 &=\int\|b_c^\omega[T]\cdot\nabla T\|^2\,d\nu_c\\
 &\quad+\E_c\| (\dot I_S^\omega-b_c^\omega[T](I_S))
                       \cdot\nabla T(I_S)\|^2.
 \end{aligned}
 \label{eq:app-residual-variance}
\end{equation}
Conservation cancels the first term, so stopped stationarity does not
require zero sampled loss.

\paragraph{Endpoint interpretation and a counterexample.}
\label{app:endpoint-limitations}
If a trajectory reaches $z\in M$, $T$ satisfies the chain rule along
it, and its terminal trace is $z$, conservation implies
$T(X(0))=T(X(t))\to z$. These trajectory and trace properties do not
follow from the almost-everywhere PDE alone.

For example, take $\mu_0=\mathcal N(0,I_d)$, $d\geq2$,
$\mu_1^c=\mu_1^\emptyset=\delta_0$, $T_\emptyset=0$, and
$T(x)=x/(1-\omega^{-1})$ for $\omega>1$. Then
\[
 \dot I_s^\omega=-x_0+
 \frac{1-\omega^{-1}}{1-s}\frac{I_s}{1-\omega^{-1}}=0,
 \qquad T(0)=0.
\]
Both clean losses vanish, but the zero field never brings off-data inputs
to the data and $T_\#\mu_0\ne\delta_0$. A noisy boundary loss at
$\gamma<1$ excludes this example: its conditional residual is
$x/(\omega-1)$, whereas $T=0$ attains both losses equal to zero.
This observation does not establish uniqueness for general data.

\subsection{Noisy boundary supervision}
\label{app:noisy-anchoring}

\subsubsection{Guided posterior regression}

Fix $0<\gamma<1$, set $m_{\gamma,c}(x)=\E_c[x_1\mid I_\gamma=x]$,
and abbreviate $T=T_\theta(\cdot,c,\gamma)$ and
$T_\emptyset=T_\theta(\cdot,\emptyset,\gamma)$. First consider the
idealized loss in which every prediction uses $I_\gamma$ and the same
boundary level:
\begin{equation}
 \mathcal L_{\partial,c}^{\omega,\gamma}(T)
 =\E_c\left\|T(I_\gamma)-\sg\left[
 x_1+(1-\omega^{-1})(T-T_\emptyset)(I_\gamma)\right]\right\|^2.
 \label{eq:app-noisy-anchor-loss}
\end{equation}

\begin{proposition}[Guided posterior regression]
\label{prop:app-anchor-fixed-point}
Assume finite second moments and hold the null branch fixed. The stopped
variation of \eqref{eq:app-noisy-anchor-loss} vanishes for every smooth
compactly supported output variation if and only if
\begin{equation}
 T=\omega m_{\gamma,c}-(\omega-1)T_\emptyset
 \quad\text{a.e. under the law of }I_\gamma.
 \label{eq:app-anchor-fixed-point}
\end{equation}
If $T_\emptyset=m_{\gamma,\emptyset}$, then
\begin{equation}
 T-T_\emptyset=\omega(m_{\gamma,c}-m_{\gamma,\emptyset}).
 \label{eq:app-anchor-contrast}
\end{equation}
\end{proposition}

\begin{proof}
Conditioning the residual on $I_\gamma=x$ gives
$T/\omega+(1-\omega^{-1})T_\emptyset-m_{\gamma,c}$. Thus
\begin{equation}
 \delta_{\mathrm{sg}}\mathcal L_{\partial,c}^{\omega,\gamma}(T)[H]
 =2\int H\cdot\left[
 T/\omega+(1-\omega^{-1})T_\emptyset-m_{\gamma,c}
 \right]\rho_\gamma^c\,dx.
 \label{eq:app-anchor-variation}
\end{equation}
Vanishing for every $H$ is equivalent to the bracket being zero,
which proves both claims.
\end{proof}

At $\gamma=s$, the contrast in \eqref{eq:app-anchor-contrast} gives
\begin{equation}
 \E_c[\dot I_s^\omega\mid I_s=x]
 =\frac{\omega m_{s,c}(x)-(\omega-1)m_{s,\emptyset}(x)-x}{1-s},
 \label{eq:app-anchor-cfg-velocity}
\end{equation}
the usual CFG combination of conditional and null flow velocities.
These fixed-time posteriors generally differ from the weighted autonomous
mean in Appendix~\ref{app:weighted-posterior}.

\subsubsection{Balance with transport and endpoint prediction}

For the unweighted joint objective
$\mathcal L_{\mathrm{tr},c}^\omega+
\lambda\mathcal L_{\partial,c}^{\omega,\gamma}$,
\eqref{eq:app-transport-variation} and \eqref{eq:app-anchor-variation}
give the stationary balance
\begin{equation}
 J_c^\omega[T]\cdot\nabla T
 =\lambda\rho_\gamma^c
 \left[T/\omega+(1-\omega^{-1})T_\emptyset-m_{\gamma,c}\right].
 \label{eq:app-joint-anchor-balance}
\end{equation}
Both updates act off $M$ and can cancel. Wherever $\rho_\gamma^c>0$,
\begin{equation}
 \begin{aligned}
 (T-T_\emptyset)-\omega(m_{\gamma,c}-m_{\gamma,\emptyset})
 ={}&\frac{\omega}{\lambda\rho_\gamma^c}
        J_c^\omega[T]\cdot\nabla T\\
 &-\omega(T_\emptyset-m_{\gamma,\emptyset}).
 \end{aligned}
 \label{eq:app-anchor-contrast-error}
\end{equation}
Thus the contrast approximation is controlled by the transport residual
relative to $\lambda\rho_\gamma^c$ and the null denoising error.
Joint stationarity alone does not make these errors small.

To describe the effect on endpoint prediction, set
\[
 m_{\gamma,c}^\omega=\omega m_{\gamma,c}-(\omega-1)T_\emptyset,
 \qquad b_{c,\gamma}^\omega=J_c^\omega[T]/\nu_c,
 \qquad k=\lambda\rho_\gamma^c/(\omega\nu_c).
\]
Hold this induced current fixed and let $\dot X=b_{c,\gamma}^\omega(X)$,
$X(0)=x$. Assume that $X$ reaches $M$ at time $u_*<\infty$,
that $T(X(u))$ is absolutely continuous and satisfies the chain rule on
preterminal intervals, and that its terminal trace is $X(u_*)$.
Write $\widetilde T(x)=X(u_*)$ and assume
$\Lambda(x)=\int_0^{u_*}k(X(u))\,du<\infty$, with the following
integrals finite. Equation~\eqref{eq:app-joint-anchor-balance} gives
$\frac{d}{du}T(X(u))=k(X(u))(T(X(u))-m_{\gamma,c}^\omega(X(u)))$.
Multiplying by $e^{-\int_0^u k(X(v))\,dv}$ and integrating yields
\begin{equation}
 \begin{aligned}
 T(x)={}&e^{-\Lambda(x)}\widetilde T(x)\\
 &+\int_0^{u_*}k(X(u))e^{-\int_0^u k(X(v))\,dv}
                 m_{\gamma,c}^\omega(X(u))\,du.
 \end{aligned}
 \label{eq:app-anchor-characteristic}
\end{equation}
The integral weights sum to $1-e^{-\Lambda(x)}$, so
\begin{equation}
 |T(x)-\widetilde T(x)|
 \leq(1-e^{-\Lambda(x)})
 \sup_{0\leq u<u_*}|m_{\gamma,c}^\omega(X(u))-\widetilde T(x)|.
 \label{eq:app-anchor-bias}
\end{equation}
This bounds the influence of noisy boundary supervision for the same
fixed current; $\widetilde T$ is not asserted to be self-consistent for
its own current. Without the assumed identity terminal trace,
\eqref{eq:app-anchor-characteristic} instead contains the terminal
prediction. A separate clean boundary loss enforces identity on the
data at function-space stationarity; continuity then supplies the trace.
The relation to the full training recipe is given in
Appendix~\ref{app:weighted-implementation}.

\subsection{Local existence of guided endpoint maps}
\label{app:local}

This section supplies an existence result in addition to the
stationary characterization. It concerns the original,
untruncated current and clean boundary trace.

\begin{theorem}[Local guided endpoint maps]
\label{thm:existence}
Let \(M\subset\mathbb R^d\) be a compact \(C^3\) embedded manifold
without boundary, of codimension \(q\geq2\). Assume a strictly
positive Schwartz source density and strictly positive \(C^2\)
conditional and null densities on \(M\).
Let \(T_\emptyset\) be the ordinary null Beckmann map.
There exist a tubular neighborhood \(U\) of \(M\), a number
\(\omega_*>1\), and a constant \(C\) such that for
\(1\leq\omega\leq\omega_*\) there is a Lipschitz map
\(T_c^\omega:U\to M\) that fixes \(M\) and is the endpoint map
of its own guided current. It satisfies
\begin{equation}
 |T_c^\omega(x)-T_c^1(x)|
 \leq C(1-\omega^{-1})\operatorname{dist}(x,M).
 \label{eq:app-local-comparison}
\end{equation}
It is unique in the controlled local class used in the proof,
and its localized clean stopped update vanishes.
\end{theorem}

\subsubsection{Terminal estimates for the ordinary fields}

Write \(k=\dim M\), \(q=d-k\), and
\(\ell_2(r)=1+|\log r|\), \(\ell_q(r)=1\) for \(q>2\).
With \(a=1-s\), the change of variables from \(x_0\) to \(I_s\)
gives
\begin{equation}
\begin{aligned}
 K(x,z)&=\int_0^1 a^{-d-1}
                  \rho_0\left(z+\frac{x-z}{a}\right)\,da,\\
 h_{c'}(x)&=\int_M K(x,z)\,\mu_1^{c'}(dz),\\
 V_{c'}(x)&=\bar z_{c'}(x)-x
       =\frac{\int_M(z-x)K(x,z)\,\mu_1^{c'}(dz)}{h_{c'}(x)},
       \qquad c'\in\{c,\emptyset\}.
\end{aligned}
 \label{eq:app-kernel}
\end{equation}
Use tubular coordinates \(x=p+r\xi\), with \(p\in M\)
and \(\xi\) a unit normal. Norms in \((p,\xi)\) below
use a fixed finite atlas of the normal bundle.

\begin{lemma}[Terminal coefficient bounds]
\label{lem:app-terminal}
For sufficiently small \(r>0\), uniformly for \(c'\in\{c,\emptyset\}\),
\begin{equation}
\begin{aligned}
 h_{c'}&\geq c_*r^{-q},&
 \|h_{c'}(\cdot,r)\|_{C^1}&\leq Cr^{-q},\\
 \|P_{T_pM}V_{c'}\|_{C^1}&\leq Cr,&
 \|P_{N_pM}V_{c'}+r\xi\|_{C^1}&\leq Cr^2\ell_q(r).
\end{aligned}
 \label{eq:app-terminal-bounds}
\end{equation}
The derivatives are in \((p,\xi)\) at fixed \(r\), and include
derivatives of the projections.
\end{lemma}

\begin{proof}
Schwartz decay gives \(K(x,z)\leq C|x-z|^{-d}\) for \(x\)
in a fixed compact tube. In a moving chart \(p=\varphi(u)\),
write \(z=\varphi(u+\eta)\). Then
\(|x-z|^2\asymp r^2+|\eta|^2\).
The tangential displacement and its moving-coordinate derivatives
are \(O(|\eta|)\); the normal displacement and its derivatives
are \(O(|\eta|^2)\), by the \(C^3\) graph expansion.
Differentiating the source argument at fixed \(\eta\) costs
at most \(C(1+(r+|\eta|)/a)\), which a stronger Schwartz
bound absorbs. Target-density, volume, and cutoff derivatives
are bounded. The far part has bounded derivatives.

For \(k>0\), the zeroth, first, and second displacement moments
are therefore bounded by constants times
\[
 \int_0^\delta
       \frac{u^{k-1+j}}{(r^2+u^2)^{d/2}}\,du,
 \qquad j=0,1,2.
\]
Their orders are \(r^{-q}\), \(r^{1-q}\), and
\(r^{2-q}\ell_q(r)\), respectively, with the same bounds
after one moving-coordinate derivative. For \(k=0\), compactness
gives finitely many atoms; the local atom has zero displacement
and the other terms are bounded.
For the lower bound on \(h_{c'}\), restrict \(a\) to \([r,2r]\)
and \(z\) to a manifold ball of radius proportional to \(r\).
The source argument stays in a compact set, so strict positivity
gives \(h_{c'}\geq c_*r^{-q}\).
Dividing the moment bounds by \(h_{c'}\) and applying the quotient
rule proves \eqref{eq:app-terminal-bounds}.
\end{proof}

Let \(p(x)\) be the nearest-point projection and \(r(x)\) the
distance to \(M\). Along any field \(V\),
\begin{equation}
 \dot p=Dp(x)V,\qquad \dot r=\xi\cdot V,\qquad
 \dot\xi=\frac{V-Dp(x)V-\xi\dot r}{r}.
 \label{eq:app-tube-ode}
\end{equation}
The bounds above imply, for the ordinary normalized fields,
\[
 \dot r=-r+O(r^2\ell_q(r)),\qquad
 \dot p=O(r),\qquad \dot\xi=O(r\ell_q(r)),
\]
with corresponding coordinate derivative bounds.
Shrinking the tube gives \(\dot r\leq-r/2\).
Using \(r\) as the clock, write \(w=(p,\xi)\) and
\(dw/dr=F_{c'}(w,r)\). Its tangent component and its Lipschitz
constant are bounded, and its full Lipschitz bound is
\(C\ell_q(r)\), integrable at zero. Integration and Gronwall's
inequality give endpoint maps \(T_{c'}\) with
\begin{equation}
 \sup\frac{|T_{c'}(p+r\xi)-p|}{r}<\infty,\qquad
 \sup_r\operatorname{Lip}_{p,\xi}
        \frac{T_{c'}(p+r\xi)-p}{r}<\infty.
 \label{eq:app-ordinary-endpoint-bounds}
\end{equation}

\subsubsection{The endpoint operator is a contraction}

Normalize the guided current by \(h_c\):
\[
 V^\omega[T]=V_c+(1-\omega^{-1})(T-T_\emptyset).
\]
Choose \(A,L\) larger than the ordinary-map bounds in
\eqref{eq:app-ordinary-endpoint-bounds}. On a closed small tube,
consider continuous \(M\)-valued maps fixing \(M\) such that
\[
 a_T(p,\xi,r)=\frac{T(p+r\xi)-p}{r},\qquad
 \|a_T\|_\infty\leq A,\qquad
 \sup_r\operatorname{Lip}_{p,\xi}a_T(\cdot,r)\leq L.
\]
This class is complete under
\(\|T-\widetilde T\|_*=\sup_{r>0}|T-\widetilde T|/r\):
the scaled displacements converge uniformly, their bounds persist,
and closedness of \(M\) preserves the range. The \(O(r)\)
bound supplies the continuous identity trace. No radial derivative
of a candidate map is assumed.

For \(u,v\in M\) within distance \(Ar\) of \(p\), the local graph
of \(M\) gives
\[
 |P_{N_pM}(u-v)|\leq C_A r\,|u-v|.
\]
Thus the guidance correction is \(O((1-\omega^{-1}) r)\) tangentially
and \(O((1-\omega^{-1}) r^2)\) normally. Every candidate still has
\(\dot r\leq-r/2\) in a sufficiently small common tube.
Substitution in \eqref{eq:app-tube-ode}, followed by division
by \(\dot r\), gives radial-clock coefficients \(F_{\omega,T}\)
satisfying
\begin{equation}
\begin{aligned}
 \|F_{\omega,T}^{p}\|_\infty&\leq C_0+(1-\omega^{-1}) C_A,\\
 \operatorname{Lip}F_{\omega,T}^{p}
       &\leq C_0+(1-\omega^{-1}) C_A(1+L),\\
 \|F_{\omega,T}\|_\infty+\operatorname{Lip}F_{\omega,T}
       &\leq C_0\ell_q(r)+(1-\omega^{-1}) C_A(1+L),\\
 \|F_{\omega,T}-F_{\omega,\widetilde T}\|_\infty
       &\leq C_A(1-\omega^{-1})\|T-\widetilde T\|_*.
\end{aligned}
 \label{eq:app-endpoint-operator-bounds}
\end{equation}
For the last bound, if \(D=\|T-\widetilde T\|_*\), the
changes in \(\dot p,\dot r,\dot\xi\) have orders
\((1-\omega^{-1}) rD,(1-\omega^{-1}) r^2D,(1-\omega^{-1}) rD\), respectively.
Dividing by \(\dot r\), of order \(r\), gives the assertion;
the change of denominator in the angular equation also contains
the bounded factor \(r\ell_q(r)\).
The other estimates follow from the same calculation and the
coordinate Lipschitz bounds on \(a_T\).

Let \(E_\omega[T]\) be the endpoint of this radial flow.
Writing
\[
 G=\exp\!\left(C_0\int_0^{r_0}\ell_q(s)\,ds
                    +(1-\omega^{-1}) C_A(1+L)r_0\right),
\]
integration of its tangent component and comparison of trajectories
give
\[
 \|a_{E_\omega[T]}\|_\infty\leq C_0+(1-\omega^{-1}) C_A,\qquad
 \operatorname{Lip}a_{E_\omega[T]}
       \leq\bigl(C_0+(1-\omega^{-1}) C_A(1+L)\bigr)G.
\]
Choose \(A>C_0\), then \(L>C_0G|_{\omega=1}\), also including
the ordinary-map bounds. Guidance strength \(\omega\) sufficiently
close to one makes the operator preserve this class. Gronwall's inequality applied
to the last line of \eqref{eq:app-endpoint-operator-bounds} yields
\begin{equation}
 \|E_\omega[T]-E_\omega[\widetilde T]\|_*
 \leq C(1-\omega^{-1})\|T-\widetilde T\|_*.
 \label{eq:app-endpoint-contraction}
\end{equation}
Choose \(\omega_*>1\) so that this factor is less than one for
\(1\leq\omega\leq\omega_*\).
The contraction theorem gives a unique fixed point in the class.
Since \(E_\omega[T_\emptyset]=T_c^1\),
\[
 \|T_c^\omega-T_c^1\|_*
 \leq C(1-\omega^{-1})\bigl(
       \|T_c^\omega-T_c^1\|_*+\|T_c^1-T_\emptyset\|_*\bigr).
\]
Rearrangement proves \eqref{eq:app-local-comparison}.

\subsubsection{Regularity, conservation, and the localized update}

The endpoint map is Lipschitz in \(p\), with angular Lipschitz
constant \(O(r)\). To compare \(r\) and \(r+\varepsilon\), first
run the latter trajectory down to \(r\). Its tangent change is
\(O(|\varepsilon|)\), and its angular change is
\(O(\int_r^{r+\varepsilon}\ell_q(s)\,ds)\).
Multiplying the latter by the endpoint map's \(O(r)\) angular
Lipschitz constant bounds its effect by \(O(|\varepsilon|)\)
for \(|\varepsilon|<r/2\). Partitioning larger radial intervals
gives the same Lipschitz bound. Together with the identity trace,
these estimates give Cartesian Lipschitz regularity.

The normalized field is now locally Lipschitz off \(M\).
Its trajectories stay in the tube and tend to \(M\);
restarting a trajectory leaves its endpoint unchanged.
Therefore \(V^\omega[T_c^\omega]\cdot\nabla T_c^\omega=0\) almost
everywhere off \(M\), and hence
\(J_c^\omega[T_c^\omega]\cdot\nabla T_c^\omega=0\).
The normalized clock approaches \(M\) as time tends to infinity.
For comparison, the kernel with \(a^{-d}\) in place of
\(a^{-d-1}\) gives
\(\nu_c(x)\asymp r^{1-q}\), whereas \(h_c(x)\asymp r^{-q}\).
Thus the field \(J_c^\omega[T_c^\omega]/\nu_c\) has inward
radial speed bounded above and below by positive constants near
\(M\), and arrives in finite time.

Both endpoint maps differ from \(p(x)\) by \(O(r)\).
For \(I_s\) in the tube,
\[
 |(T_c^\omega-T_\emptyset)(I_s)|
 \leq C\operatorname{dist}(I_s,M)
 \leq C(1-s)|x_0-x_1|.
\]
The source has finite moments and the target is compact, so this
bound and the bounded local Jacobian make all localized stopped
pairings finite. Conservation and the identity trace then make
\eqref{eq:app-total-variation} vanish for variations supported
inside the tube. This completes the proof of
Theorem~\ref{thm:existence}.

\paragraph{Scope of the result.}
Uniqueness here is only within the controlled local class.
The theorem does not construct a global map on almost every
source input, prove convergence of neural-network training, or
identify its selected stationary branch.
If the conditional support consists of components of a larger
compact null support, separated by a positive distance from the
other components, the same proof applies provided the null
density is positive and smooth on the conditional components:
the additional components contribute bounded kernel moments
and derivatives in the terminal tube.

\subsection{Two-mode reweighting and nonuniqueness}
\label{app:binary}

This example supplies a fully global result for a subclass already
included in BTM's assumptions. It also shows why conservation, endpoint
self-consistency, and even zero loss cannot select a unique guided law.

Let $d\ge2$, $a>0$, $\mu_0=\mathcal N(0,I_d)$, and
\begin{equation}
 \mu_1^c=p\delta_{a e_1}+(1-p)\delta_{-a e_1},\qquad
 \mu_1^\emptyset=q\delta_{a e_1}+(1-q)\delta_{-a e_1},
 \quad 0<q<p<1.
 \label{eq:binarylaws}
\end{equation}
The null law is compatible with an actual class mixture: choose a class
prior $0<\pi<q/p$ and give another class plus weight
$(q-\pi p)/(1-\pi)\in(0,1)$.
The case $p<q$ follows by reflecting the first coordinate.

\begin{theorem}[A continuum of global self-consistent maps]
\label{thm:binary}
For every $\omega>1$ in the setting~\eqref{eq:binarylaws}, there is a
continuum of measurable, piecewise constant, self-consistent guided maps.
They fix both data atoms, have zero numerical loss in~\eqref{eq:loss},
and have pairwise different generated laws.

One member is the unchanged ordinary conditional map $T_c^1$, with
generated plus mass $p$. Other members assign strictly more than $p$
source mass to the plus atom. They converge to $T_c^1$ in $L^1(\mu_0)$
as their basin boundary approaches the ordinary conditional boundary.
All members agree in neighborhoods of both atoms.
\end{theorem}

\subsubsection{Ordinary basins as scalar graphs}

Write $x=(u,v)$ with $u\in\R$, $v\in\R^{d-1}$, and $r=|v|$.
For a plus weight $w\in(0,1)$, define $K_\pm$ by~\eqref{eq:app-kernel}
at $z=\pm ae_1$. Rotational symmetry of the Gaussian gives
\begin{equation}
 h_w=wK_++(1-w)K_-,\qquad
 V_w=(f_w(u,r),-v),\qquad
 f_w=a\frac{wK_+-(1-w)K_-}{h_w}-u.
 \label{eq:binaryfield}
\end{equation}
These functions are smooth away from the atoms. The normalized field
extends locally Lipschitz by zero at each atom: the local kernel is of
order $|x-z|^{-d}$ and the other atom's contribution is bounded, so
$V_w(x)=z-x+O(|x-z|^d)$ with derivative error $O(|x-z|^{d-1})$.
Every forward solution has $r(t)=e^{-t}r(0)$ and bounded $u(t)$ because
$f_w=-u+m_w$ with $|m_w|\le a$.

\begin{lemma}[Ordinary separating graph]
\label{lem:graph}
There is a smooth radial graph $u=s_w(r)$, $r\ge0$, separating the
ordinary plus and minus basins. Its value $u_w=s_w(0)$ lies in $(-a,a)$.
It satisfies
\begin{equation}
 r s_w'(r)=-f_w(s_w(r),r).
 \label{eq:graphode}
\end{equation}
Points above the graph end at $ae_1$, points below it end at $-ae_1$,
and points on it converge to $(u_w,0)$. Moreover $s_p(r)<s_q(r)$ for
every $r\ge0$.
\end{lemma}

\begin{proof}
\emph{The axial saddle is unique.}
On the axis, $f_w$ is negative just to the right of $-a$ and positive
just to the left of $a$; the local-kernel estimate above supplies these
signs. At any zero $u_*\in(-a,a)$, $V_w(u_*,0)=0$ and
$\nabla\cdot(h_wV_w)=\rho_0$ gives
\begin{equation}
 \partial_u f_w(u_*,0)=d-1+\frac{\rho_0(u_*,0)}{h_w(u_*,0)}>0.
 \label{eq:saddle}
\end{equation}
Every zero therefore crosses upward. There can only be one: two upward
crossings would require a zero without an upward crossing between them.
Call the zero $u_w$ and its derivative $\lambda_w>0$.
Outside $[-a,a]$, $f_w$ points toward that interval.

\emph{Construct the local graph explicitly.}
Near $(u_w,0)$, radial smoothness gives
$f_w(u_w+z,r)=\lambda_w z+R(z,r)$ with
$R=O(z^2+r^2)$ and $\partial_zR=O(|z|+r^2)$.
The solution of~\eqref{eq:graphode} approaching $u_w$ obeys
\begin{equation}
 z(r)=-r^{-\lambda_w}
          \int_0^r t^{\lambda_w-1}R(z(t),t)\,dt.
 \label{eq:volterra}
\end{equation}
On a small interval, this is a contraction in a closed ball of the norm
$\sup_{r>0}|z(r)|/r^2$. It gives $z=O(r^2)$, and differentiation of
the equation gives a smooth radial graph. This is the stable graph of
the saddle, obtained without a global stable-manifold assumption.
For every $r>0$, the graph ODE is nonsingular. Since
$|r s_w'|\le |s_w|+a$, it extends to every finite $r$ and obeys
$s_w(r)=O(r)$, $s_w'(r)=O(1)$ as $r\to\infty$.

\emph{Classify the endpoints.}
Uniqueness of the scalar ODE prevents crossing of the graph. Since
$r(t)\to0$ and $u(t)$ is bounded, its limit must be an axial equilibrium.
Here is a direct justification of that assertion. On bounded intervals,
$f_w(u,r(t))$ converges uniformly to $f_w(u,0)$. Near any point where
this limit is nonzero, sufficiently late crossings of a short interval
can only occur in its sign direction. Such an interval cannot be crossed
back and forth indefinitely. Thus the omega-limit interval of $u(t)$
is contained in the finite zero set $\{-a,u_w,a\}$, and is a singleton.
If $u(t)\to u_w$, the difference $d(t)=u(t)-s_w(r(t))$ obeys
$d'(t)=A(t)d(t)$ with $A(t)\to\lambda_w>0$.
A nonzero difference cannot converge to zero, so only points on the
graph approach $u_w$. Points above must therefore end at $a$, and points
below at $-a$.

Finally, positivity of both kernels gives $f_p>f_q$ everywhere off the
atoms. Hence $u_p<u_q$. If the ordered graphs first met at $r>0$,
their derivative difference at the meeting would be
$-(f_p-f_q)/r<0$, preventing a crossing from below. They remain
strictly ordered for all $r$.
\end{proof}

The graph has zero source and interpolation probability. Consequently,
up to irrelevant values on that graph,
$T_\emptyset(u,v)=ae_1$ above $s_q(r)$ and $-ae_1$ below it.
Ordinary BTM transport also gives
$\mu_0\{u>s_p(|v|)\}=p$.

\subsubsection{Move the conditional boundary while preserving its own flow}

For sufficiently small $\beta\ge0$, the equation
$f_p(u_\beta,0)+\beta=0$ has a root near $u_p$ with positive derivative.
Choose $\beta_*>0$ such that for $0\le\beta\le\beta_*$ this root is
in $(-a,u_q)$ and
\begin{equation}
 f_p(u,0)+\beta>0\quad(u_\beta<u\le u_q).
 \label{eq:biasedsign}
\end{equation}
This is possible by the simple root at $u_p$ and the strict positivity
of $f_p$ on $(u_p,u_q]$. The implicit function theorem gives
$du_\beta/d\beta=-1/\partial_u f_p(u_\beta,0)<0$.
Repeating~\eqref{eq:volterra} constructs the global graph for the
auxiliary field $(f_p+\beta,-v)$:
\begin{equation}
 r s_\beta'(r)=-f_p(s_\beta(r),r)-\beta,\qquad
 s_\beta(0)=u_\beta.
 \label{eq:betagraph}
\end{equation}
The same scalar comparison proves that $s_\beta$ strictly decreases
with $\beta$, and that $s_\beta<s_p<s_q$ for $\beta>0$.
Write $s_0=s_p$.

Fix $\omega>1$ and choose
$0\le\beta\le\min\{\beta_*,2(1-\omega^{-1}) a\}$, decreasing $\beta_*$
slightly if necessary to keep all endpoint inequalities strict.
Define $T_\beta$ to be $ae_1$ above $s_\beta(r)$ and $-ae_1$ below it,
and fix the two atoms. Values on the separating graph are arbitrary.
The actual guided field generated by this map is
\begin{equation}
 V_\omega[T_\beta](u,v)=
 \begin{cases}
 (f_p(u,r),-v),&u<s_\beta(r),\\
 (f_p(u,r)+2(1-\omega^{-1}) a,-v),&s_\beta(r)<u<s_q(r),\\
 (f_p(u,r),-v),&u>s_q(r).
 \end{cases}
 \label{eq:piecewisefield}
\end{equation}

\emph{Neither label region loses its own trajectories.}
At the lower interface, the derivative of $u-s_\beta(r)$ has the
minus-side value $-\beta\le0$ and the plus-side value
$2(1-\omega^{-1}) a-\beta\ge0$. Scalar comparison therefore preserves each
open label region. In the minus region, the actual field is the ordinary
conditional one and $u<s_\beta\le s_p$, so the endpoint is $-ae_1$.

\emph{Every plus trajectory reaches the ordinary plus basin above the
null interface.}
Until it reaches $s_q$, a plus trajectory is driven by $f_p+2(1-\omega^{-1}) a$.
It lies above the solution driven by $f_p+\beta$ with the same initial
condition. The latter starts strictly above $s_\beta$.
It must cross $s_q$ in finite time: otherwise its eventual values lie
between $u_\beta$ and $u_q$. The sign condition~\eqref{eq:biasedsign}
and the omega-limit argument in Lemma~\ref{lem:graph} force convergence
to $u_\beta$, which is impossible off its stable graph by the positive
axial derivative. Thus the actual trajectory also crosses in finite time.

At $s_q$, the conditional ordinary normal numerator is
$f_p(s_q,r)-f_q(s_q,r)>0$.
The guided normal numerators on the two sides are respectively this
positive quantity plus $2(1-\omega^{-1}) a$ and this quantity itself.
Hence the crossing is transverse in the same direction on both sides;
there is a unique local Carath\'eodory continuation across the null
interface. The trajectory cannot return below it.
Above it the ordinary conditional field ends at $ae_1$.
These statements also hold on the axis except for the separating point,
which has zero source measure. Near both atoms guidance vanishes, so
arrival is finite in the original autonomous clock, as in ordinary BTM.

This proves that $T_\beta$ is the endpoint map of its own guided field
for almost every source input. Its derivative is zero almost everywhere
under $\nu_c$, and it fixes the atoms. The numerical transport residual is consequently zero.
The clean guided boundary loss also vanishes, since $T_\beta$ and
$T_\emptyset$ fix the atoms. Hence the complete loss~\eqref{eq:loss}
is zero.

The strictly ordered graphs enclose open sets of positive Gaussian mass.
Therefore the generated plus mass strictly increases with $\beta$.
Pointwise convergence $s_\beta(r)\to s_p(r)$, followed by dominated
convergence away from the null-probability graph, proves
$T_\beta\to T_c^1$ in $L^1(\mu_0)$.
Because $u_\beta$ stays uniformly inside $(-a,a)$, all sufficiently
small members are constant with the correct label in common neighborhoods
of the two atoms. This completes the proof of Theorem~\ref{thm:binary}.

The strictly ordered basins already establish the change in generated
mass. These are distinct actual endpoint maps, so the example proves
nonuniqueness of the generated law even under endpoint self-consistency.
All members agree near the data; local existence therefore does not
select a global law.

\subsubsection{Transport identities for regular maps}
\label{app:transport}

Suppose that, for $\mu_0\otimes\mu_1^c$-almost every pair,
$s\mapsto T(I_s)$ is absolutely continuous on $[0,1]$, with derivative
$\dot I_s\cdot\nabla T(I_s)$, and that the direction, chain-rule
terms, and endpoint errors below are integrable. For
$\varphi\in C_c^1(\R^d)$, the fundamental theorem of calculus gives
\begin{equation}
\begin{aligned}
 \int\varphi\,d(T_\#\mu_0-\mu_1^c)
 ={}&\E_c[\varphi(T(x_1))-\varphi(x_1)]\\
 &-\int\nabla\varphi(T)\cdot
       (J_c^\omega[T]\cdot\nabla T)\,dx\\
 &+\int\nabla(\varphi\circ T)\cdot\Delta J_c^\omega[T]\,dx.
 \label{eq:regularfull}
\end{aligned}
\end{equation}
Indeed, the source-to-target difference for $\varphi(T(I_s))$ is
$-\E_c[\nabla\varphi(T(I_S))\cdot(\dot I_S\cdot\nabla T(I_S))]$.
Insert the guided correction and condition on $I_S$ to obtain the
displayed identity. Under conservation and the clean boundary condition,
\begin{equation}
 \int\varphi\,d(T_\#\mu_0-\mu_1^c)
 =\int\nabla(\varphi\circ T)\cdot\Delta J_c^\omega[T]\,dx.
 \label{eq:regulartransport}
\end{equation}
The absolute-continuity assumption excludes the piecewise constant
two-atom maps above; their reweighting follows directly from the ordered
basins, without applying this identity across jumps.

\section{Implementation details}
\label{appendix:training}

\subsection{Training algorithm}

Algorithm~\ref{algo:training_simple} gives the basic training rule;
Algorithm~\ref{algo:training_complicated} gives the full recipe.
We distinguish the network's boundary-level input $\gamma$ from
the sampled transport time $S$ and boundary time $S'$.
The live predictions are $T_\theta(I_S,c,\gamma)$ and
$T_\theta(I_{S'},c,\gamma)$, respectively. The guidance predictions
instead set the third input equal to the corresponding interpolation
time, $S$ or $S'$. Here $\bar T$ denotes the frozen guidance copy,
periodically refreshed from the EMA network $T_{\theta_{\mathrm{EMA}}}$.
The notation $x_1^\omega$ denotes the guided boundary target,
and $\kappa$ is the read-time label scale.

\begin{algorithm}[H]
\small
\DontPrintSemicolon
\caption{SBM training and one-step sampling ($\omega=1$ gives unguided BTM)}
\label{algo:training_complicated}

\KwIn{Map $T_\theta(x,c,\gamma)$ with null condition $c=\emptyset$;
guidance strength $\omega\geq1$ and band $[s_{\mathrm{lo}},s_{\mathrm{hi}}]$;
dropout probability $p$; minimum boundary level $\gamma_{\min}$;
guidance start $K_0$; refresh period $K$; EMA decay $\beta$;
learning rate $\eta$}
\KwOut{Trained map and EMA checkpoints}

Initialize $\theta_{\mathrm{EMA}}\leftarrow\theta$ for a new run,
or restore saved state when resuming\;

\For{each training step $k$}{
    \tcp{Sample a minibatch; the following draws are per example}
    Draw $c\sim\pi$, $x_1\sim\mu_1^c$, and $x_0\sim\mu_0$\;
    Replace $c$ by $\emptyset$ with probability $p$\;
    Draw $\gamma\sim U(\gamma_{\min},1)$\;
    Draw $S$ from $\operatorname{LogitNormal}(-1.2,1)$ or
    $U(0.3,\gamma)$ with equal probability\;
    Set $S'=\gamma$ with probability $\tfrac12$;
    otherwise draw $S'\sim U(\gamma,1)$\;
    $I_S\leftarrow(1-S)x_0+Sx_1$,
    \quad $I_{S'}\leftarrow(1-S')x_0+S'x_1$\;

    \BlankLine
    \tcp{Construct guided targets}
    $\dot I_S^\omega\leftarrow x_1-x_0$,
    \quad $x_1^\omega\leftarrow x_1$\;
    \If{$k=K_0$}{
        $\bar T\leftarrow T_{\theta_{\mathrm{EMA}}}$\;
    }
    \If{$k\geq K_0$ and $c\neq\emptyset$}{
        \If{$S\in[s_{\mathrm{lo}},s_{\mathrm{hi}}]$}{
            $\dot I_S^\omega\leftarrow\dot I_S^\omega$\;
            \Indp
            $\displaystyle
            {}+\frac{1-\omega^{-1}}{1-S}
            \bigl[\bar T(I_S,c,S)-\bar T(I_S,\emptyset,S)\bigr]$\;
            \Indm
        }
        \If{$S'\in[s_{\mathrm{lo}},s_{\mathrm{hi}}]$}{
            $x_1^\omega\leftarrow x_1$\;
            \Indp
            ${}+(1-\omega^{-1})
            \bigl[\bar T(I_{S'},c,S')-\bar T(I_{S'},\emptyset,S')\bigr]$\;
            \Indm
        }
    }

    \BlankLine
    \tcp{Compute per-example losses}
    $d_{\mathrm{tr}}\leftarrow T_\theta(I_S,c,\gamma)$\;
    \Indp
    ${}-\sg\!\left[
        T_\theta(I_S,c,\gamma)
        +D_xT_\theta(I_S,c,\gamma)\,\dot I_S^\omega
    \right]$\;
    \Indm
    $\mathcal L_{\mathrm{tr}}\leftarrow
        \sg\!\left[(\|d_{\mathrm{tr}}\|^2+0.01)^{-1}\right]
        \|d_{\mathrm{tr}}\|^2$\;
    $\mathcal L_\partial\leftarrow
        \bigl\|T_\theta(I_{S'},c,\gamma)-\sg(x_1^\omega)\bigr\|^2$\;

    \BlankLine
    \tcp{Update the map and EMA}
    Average $L_{\mathrm{tr}}$ and $L_\partial$ over the minibatch\;
    Choose $(w_{\mathrm{tr}},w_\partial)$ to scale up the smaller
    final-layer gradient norm to match the larger one\;
    $\theta\leftarrow\theta-\eta\,\nabla_\theta
        (w_{\mathrm{tr}}L_{\mathrm{tr}}+w_\partial L_\partial)$\;
    $\theta_{\mathrm{EMA}}\leftarrow
        \beta\theta_{\mathrm{EMA}}+(1-\beta)\theta$\;
    \If{$k\geq K_0$ and $k\bmod K=0$}{
        $\bar T\leftarrow T_{\theta_{\mathrm{EMA}}}$\;
    }
}

\BlankLine
\textbf{One-step sampling}\;
Select the evaluation EMA checkpoint and
$(\gamma_{\mathrm{read}},\kappa)$ for the experiment\;
Scale condition embeddings about the null embedding by $\kappa$\;
Draw $x_0\sim\mu_0$\;
$x\leftarrow T_{\theta_{\mathrm{EMA}}}
    (x_0,c,\gamma_{\mathrm{read}})$\;

\end{algorithm}

\subsection{Relation to the population analysis}
\label{app:weighted-implementation}

The clean stopped-update and local existence results concern the ideal
unweighted objective. Algorithm~\ref{algo:training_complicated} uses
frozen guidance predictions, a guidance band, nonuniform time sampling,
weighted transport residuals, and noisy boundary inputs.
These choices alter the current and the balance between the updates.

\paragraph{Loss weights.}
Under the uniform-time law of the ideal objective, a stopped transport
weight $W\geq0$ gives the current
\[
 J_{c,W}^\omega[T](x)
 =\nu_c(x)\E_c[W\dot I_S^\omega\mid I_S=x].
\]
A positive input-only weight $W=w(I_S)$ rescales the current and
preserves its conservation equation. Time-, pair-, or residual-dependent
weights generally change it. A stopped boundary weight similarly replaces
the fixed-time posterior by
$\E_c[Wx_1\mid I_\gamma=x]/\E_c[W\mid I_\gamma=x]$;
input-only weights cancel. For the full recipe, the conditional expectation
must also use its actual time law and frozen guided direction.

\paragraph{Noisy boundary targets.}
With the algorithm's frozen target $x_1^\omega$ held fixed, the
unweighted boundary loss alone has regression solution
\[
 T_\theta(x,c,\gamma)=\E_c[x_1^\omega\mid I_{S'}=x,\gamma].
\]
For $S'=\gamma$, active guidance, and frozen predictions equal to
the current predictions, this reduces to
\eqref{eq:app-anchor-fixed-point}. The recipe also uses $S'>\gamma$
and joint transport training, so exact fixed-time denoising is not
guaranteed. Reading the guidance predictions at boundary level $s$
is motivated by their denoising supervision at that input's noise level;
generation instead reads a fixed $\gamma_{\mathrm{read}}$.

\paragraph{The endpoint denominator.}
For continuous maps,
\[
 (1-s)\dot I_s^\omega
 \longrightarrow(1-\omega^{-1})(T-T_\emptyset)(x_1).
\]
For $\omega>1$, a nonzero trace gap on a positive-measure set therefore
gives infinite direction moments of every order $p\geq1$ under
uniform time sampling. If $T-T_\emptyset$ instead vanishes on the
data and is $L$-Lipschitz along the interpolation points, then
\[
 |(T-T_\emptyset)(I_s)|\leq L(1-s)|x_0-x_1|,
 \qquad
 |\dot I_s^\omega|\leq[1+(1-\omega^{-1})L]|x_0-x_1|.
\]
This supplies finite direction moments when the endpoint pairs have them;
a bounded Jacobian additionally gives a finite squared transport residual.
The practical guidance band stays away from $s=1$ and defines its own
modified current.

\section{Experimental details and additional results}
\label{app:experiments}

We retain the configurations and evaluation choices needed to interpret
the reported results. Full configurations, training scripts, and evaluation
code will accompany the public release.

\subsection{ImageNet}
\label{app:imagenet}

\subsubsection{Data and architectures}
\label{app:in-data}
\label{app:in-map}

We use the 1,281,167 ImageNet-1k training images at $256\times256$,
with ADM center cropping~\citep{deng2009imagenet,dhariwal2021diffusionmodelsbeatgans}.
The validation set is reserved for model selection.
SD-VAE maps use \texttt{sd-vae-ft-mse} latents of size
$32\times32\times4$, scaled by $0.18215$; posterior moments for each
image and its horizontal flip are cached, and training samples from one
of the two posteriors. RAE maps use unscaled $16\times16\times384$
tokens from frozen DINOv2-S with registers (224\,px input), encoded
without flips and decoded by the RAE ViT decoder~\citep{rombach2021highresolution,zheng2025diffusiontransformersrepresentationautoencoders}.
RAE interpolation times use the dimension shift
$t=u/[u+4.899(1-u)]$; time laws below are in nominal time $u$.

The SD-VAE maps are DiT-B/2 and XL/2; the RAE maps use
DiT$^{\mathrm{DH}}$ with patch size 1~\citep{peebles2023scalablediffusionmodelstransformers,zheng2025diffusiontransformersrepresentationautoencoders}.
The boundary level $\gamma$ occupies the time-embedding slot, and
condition dropout trains a null class embedding. The network predicts
the clean latent directly; the transport JVP differentiates the whole map.

\subsubsection{Training configurations}
\label{app:in-recipe}

All maps use Algorithm~\ref{algo:training_complicated}, batch 256,
$\gamma_{\min}=0.35$, condition dropout $0.1$, guide refresh every
10k updates from EMA $0.999$, and an additional EMA $0.9999$ for
evaluation. Muon updates matrix weights and AdamW updates embeddings,
biases, and norms at one tenth of the rate, with betas $(0.9,0.95)$,
no weight decay, warm-up, or gradient clipping, and bf16 autocast
\citep{jordan2024muon}. Table~\ref{tab:recipes} lists the model-specific
choices. The SD-VAE B/2 transport logit-normal mean is $0.8$ for the
first 25k updates and $0$ thereafter; the other maps use $-1.2$.

\begin{table}[t]
\centering
\caption{\textbf{ImageNet configurations.} Read settings are
$(\gamma_{\mathrm{read}},\kappa,\mathrm{EMA})$. Compute covers generator
training up to the checkpoint that is read; the XL/2 budget excludes its released BTM base.}
\label{tab:recipes}
\small
\setlength{\tabcolsep}{4pt}
\resizebox{\linewidth}{!}{%
\begin{tabular}{@{}lllll@{}}
\toprule
 & SBM-B/2 (SD-VAE) & SBM-XL/2 (SD-VAE) & SBM-B (RAE) & SBM-XL (RAE) \\
\midrule
architecture & DiT-B/2, 130M & DiT-XL/2, 675M & DiT-B/1-DH, 128.8M & DiT-XL/1-DH, 669.4M \\
start & scratch & BTM base, 1240 epochs & scratch & scratch \\
updates & 780k (156 epochs) & 355k guided; read at 340k & 750k (150 epochs) & 1M + 42.5k polish \\
Muon rate & $10^{-3}$ & $5\times10^{-4}$ & $5\times10^{-4}$ & $5\times10^{-4}$ \\
guidance starts & 35k & 2k & 35k & 35k \\
$\omega$ & 5 & 3 initially, then 5; 3.3 from 205k & 5 & 5; 3.7 from 475k \\
gradient balance & adaptive, EMA 0.99 & fixed $\times2000$ transport & measured to 35k, then fixed & measured to 35k, then fixed \\
late boundary weight & none & $\min\{0.4/(1-S'),8\}$ for $S'>0.6$ & as XL/2 & as XL/2 \\
read & $(0.8,1.25,0.999)$ & $(0.95,1.2,0.9999)$ & $(0.8,1,0.9999)$ & $(0.8,1.35,0.9999)$ \\
compute (EFLOP) & 100 & 228 & 110 & 840 \\
FID-50k & 5.85 & 3.5 & 2.79 & 1.61 \\
\bottomrule
\end{tabular}}
\end{table}

The B/2 balancer rescales transport by the smoothed boundary-to-transport
gradient-norm ratio. RAE maps measure the ratio with EMA $0.999$ until
guidance starts and then freeze it. Guidance uses the band $[0.35,0.6]$.
The XL/2 continuation uses $\omega=3$ for its first 10k updates, then
5 until update 205k; its guidance predictions come from the live map
until 35k, the fast EMA until 45k, and a periodically frozen copy thereafter.
Its base and first 107.5k continuation updates use bilinearly resized
images; subsequent updates use ADM preprocessing.

\paragraph{Final checkpoints.}
\label{app:in-long}
The B/2 run uses a cosine decay to zero over its last 36 epochs.
The RAE XL run adds autoguidance of weight $0.09$ against its own
160k checkpoint from update 475k~\citep{karras2024autoguidance}.
Its final polish averages the EMA-$0.9999$ trunks and embedders of seven
late checkpoints of the autoguided run, retaining the final checkpoint's
head. It then trains the head at $5\times10^{-5}$ with
a cool-down, followed by the head and last four trunk blocks at $10^{-5}$.
The averaging changes FID from 2.18 to 1.74 and the subsequent updates
to 1.61. Continuations resume optimizer and EMA state; the reported
compute includes the final training stages.

\subsubsection{Sampling and evaluation}
\label{app:in-protocol}

Sampling evaluates $T_{\theta_{\mathrm{EMA}}}(x_0,c,\gamma_{\mathrm{read}})$
once in bf16 and decodes the result. The read-time label scale acts as
$e_c\leftarrow e_\emptyset+\kappa(e_c-e_\emptyset)$ and adds no
network call. Checkpoints, EMA copies, read levels, and label scales
are selected using 10k-sample FID against the ImageNet validation set;
final FID~\citep{heusel2017fid} uses 50k samples (50 per class) against ADM
training-set statistics. The evaluator reproduces released iMF-B/2 and XL/2 at
their reported 3.32 and 1.72. Results use one sampling seed; the final
RAE XL checkpoint gives 1.67 on a second seed, compared with 1.61.

IS uses the same samples and ten splits. FDr6~\citep{fdloss} averages
training-reference Fr\'echet distances in Inception-V3, ConvNeXt-v2
(base, 224 px), DINOv2-L, MAE-L, SigLIP-2, and CLIP-L, each normalized
by its validation-set value. We score our 50k pool with the six encoders
and reference statistics of~\citet{fdloss}, as packaged in the released
evaluator of~\citet{priming} (commit \texttt{b29aaf8}); the record reads
6.22, with components 0.99, 1.77, 5.26, 6.59, 10.87 and 11.82 in that
order. Table~\ref{tab:fdr6} lists the baselines as measured
by~\citet{fdloss} on their own samples of the released checkpoints; those
are the values in Figure~\ref{fig:imagenet_baselines}, paired with each
paper's own FID.

\begin{table}[t]
\centering
\small
\begin{tabular}{lc}
\toprule
model & FDr6 \\
\midrule
Drifting-L, pixel & 10.51 \\
Drifting-L, SD-VAE latent & 10.92 \\
iMF-XL/2 & 8.39 \\
pMF-L & 9.09 \\
pMF-H & 6.87 \\
SBM-XL (ours) & 6.22 \\
\bottomrule
\end{tabular}
\caption{\textbf{FDr6 of one-step maps.}
FDr6~\citep{fdloss}: the Fr\'echet distance in six feature spaces
(Inception-V3, ConvNeXt-v2, DINOv2-L, MAE-L, SigLIP2, CLIP-L) against the
ImageNet training set, each divided by its value for the validation set,
averaged over 50k samples; the validation set reads 1.00 by construction.
Baseline rows are the measurements of~\citet{fdloss} on their samples of
the released checkpoints; our row is our 50k pool scored with the same
encoders and reference statistics. Drifting trains on its own MAE
features and pMF on a ConvNeXt-V2 perceptual loss; maps post-trained on the evaluation encoders themselves are left out. One network
evaluation per sample throughout.}
\label{tab:fdr6}
\end{table}

Precision and recall use $k=3$ in Inception space with 50k generated
and 50k real images~\citep{kynkäänniemi2019improvedprecisionrecallmetric}.
The class diagnostics compare generated and validation class-mean
separation and within-class standard deviation, using 50 samples per class.

Every point in the pretraining curves is an ADM FID-50k read. RAE curves
use $\gamma_{\mathrm{read}}=0.8$, $\kappa=1$, and the better EMA copy;
SD-VAE B/2 uses the better EMA and $\kappa\in\{1,1.25,1.5\}$.
Baselines use their own guidance scales selected by 10k reads.

\subsubsection{Compute and baselines}
\label{app:in-compute}

Training FLOPs count two operations per multiply-accumulate and are
measured over a real training step, including the transport JVP and
backward pass. SBM costs 7.05 generator forwards per sample
before guidance and 11.05 after it, measured at base size and used for all
four maps; the four frozen guidance reads account
for the difference, and the two weak-model reads of autoguidance raise the
RAE XL run to 13.05 from update 475k. Forward costs for the four maps in
Table~\ref{tab:recipes} are 46.0, 237, 52.6, and 263 GFLOP.
Training budgets exclude data encoding and the sampling decoder. Baseline
budgets are our own counts under the same rule: each model's forward cost
from its paper, times the forward-equivalents per sample of its released
training step, counted on the lowered computation graph for the JAX codes of
iMF and MeanFlow, times its stated epochs or updates and batch. iMF's step runs a shared trunk and two eight-layer heads, so its
forward-equivalents fall with depth: 9.1 at B/2, 7.8 at L/2 and 7.5 at XL/2,
measured at each size; pMF, whose implementation follows iMF's, takes the
value of its depth, without its LPIPS and ConvNeXt-V2 perceptual losses,
which we do not count. iMF-XL/2, at 349 GFLOP per forward and 800 epochs,
comes to 2{,}700 EFLOP. MeanFlow costs 6.05. Drifting and W-Flow are counted per update from the
released Drifting program and the sample counts of each paper: 8{,}192
generated samples, each costing 3.2 generator forwards (161 GFLOP for
DiT-L/2, 237 for DiT-XL/2) and 463 GFLOP forward and backward through the
MAE-640 feature network, and the real samples of the update, 12{,}288 for
Drifting-L/2 and 20{,}480 for W-Flow-XL/2, at 232 GFLOP each; W-Flow adds
its second generated batch of 8{,}192 without gradient. Both train for 200k
updates. The XL/2 fine-tuning budget excludes its
base pretraining.
The RAE XL record costs 842 EFLOP, rounded to 840 in the table: the main run
to update 475k, the autoguided run from 475k to 1M, and the 42.5k polish
updates, counted at the full rate.

Inference costs one generator forward plus one decoder call: 274 and
484 GFLOP for RAE B and XL, and 666 and 857 GFLOP for SD-VAE B/2 and XL/2.
On four H200s, B/2 achieves 5.1 unguided or 3.2 guided updates/s and its
156-epoch run takes 65 hours. XL/2 takes 19 minutes per 1k updates on
four H200s; RAE XL trains on eight H200s.

\paragraph{Baseline runs.}
\label{app:in-baselines}
We run released baseline code on the same ImageNet data with the authors'
recipes. iMF-B/2 uses its epoch-matched setting (batch 256, Adam $10^{-4}$,
10-epoch warm-up, $\omega\in[1,8]$), to 176 epochs/101 EFLOP;
MeanFlow-B/2 uses batch 256, Adam $10^{-4}$, $\omega'=2$, to
282 epochs/101 EFLOP; Shortcut-B uses its ImageNet-256 recipe (batch 256,
CFG 1.5, bootstrap fraction $1/4$), to 180 epochs/35 EFLOP.
Latent Drifting-B/2 uses MAE-256 features, 4{,}096 generated and 4{,}096 real
images per update (its ablation setting) across eight GPUs,
and 90k updates/100 EFLOP. Their per-sample training costs are 450,
278, 151, and approximately 271 GFLOP, respectively
\citep{geng2025improvedmeanflow,geng2025meanflow,frans2025shortcut,deng2026drifting}.
Final-quality comparisons use published results; pretraining curves use
these reruns and the common evaluation protocol above.

\paragraph{The iMF recipe on the same network and latent.}
\label{app:in-imf-rae}
To separate the latent from the recipe, we trained iMF's released recipe on the SBM-B network (DiT-B/1-DH, 128.8M parameters; 133.4M with iMF's inputs and velocity output) in
the same RAE latent with the same batch of 256: AdamW at $10^{-4}$ without weight decay, 50k warm-up updates and then a constant
rate, logit-normal times with the RAE paper's shift of $4.899$, iMF's guidance-aware objective, and one-step reads at the best of
$w \in \{1.5, 2, 3\}$ and two EMA rates per checkpoint on the guidance window $[0.6, 1]$; 760k updates, 152 epochs, 82~EFLOP. Its training step, counted as ours (Appendix~\ref{app:in-compute}), costs 421~GFLOP per sample,
8.0 forwards of the network: three reads without gradient and one JVP with its backward pass. Table~\ref{tab:imf_rae} sets it against SBM-B in the same latent and against iMF's
own run in the SD-VAE latent. The latent helps iMF early ($12.7$ at 20 epochs against $27.9$ in the SD-VAE latent) but not late:
from 50 epochs the RAE run trails iMF's SD-VAE run and reads $6.6$ at 150 epochs, where SBM-B reads $3.3$ at a fixed read
setting ($2.8$ at its final read, Table~\ref{tab:recipes}). The same network, latent and batch under the two recipes differ
by a factor of two in FID after 150 epochs, at 81 against 110~EFLOP; after 100 epochs and 73~EFLOP SBM-B already reads $3.6$. The gap to iMF in Figure~\ref{fig:imagenet_baselines} is therefore the training recipe, not the
latent. The curve is the dashed navy line in the left panel of that figure, drawn through the reads at 5, 10 and 20 epochs and
from 50 epochs on. Between 15 and 45 epochs the run stalls (12.9, 12.7, 13.3, 14.6, 13.4, 11.7 and 10.7): the reads at guidance
scales 2 and 3 get worse by one to two points there while the scale-1.5 read only stalls, as the best read scale moves from 3 to
1.5 over training; the training losses show no event, and from 40 epochs the run improves at every scale.

\begin{table}[t]
  \centering
  \caption{\textbf{The iMF recipe on the SBM-B network in the RAE latent.} One-step FID-50k (ADM reference) during training, base
  size, batch 256, one seed. SBM-B is read at one fixed setting (anchor $0.8$, label scale $1.25$, EMA $0.9999$); the iMF runs at the
  best of their guidance-scale and EMA grid per checkpoint. The last column is the last read of each line: 150 epochs for the two RAE runs,
  176 for iMF-B/2.}
  \label{tab:imf_rae}
  \small
  \begin{tabular}{llcccc}
    \toprule
    recipe & network, latent & 20 ep & 50 ep & 100 ep & end \\
    \midrule
    SBM & DiT-B/1-DH, RAE & 6.7 & 4.5 & 3.6 & 3.3 \\
    iMF & DiT-B/1-DH, RAE & 12.7 & 10.3 & 7.5 & 6.6 \\
    iMF & DiT-B/2, SD-VAE & 27.9 & 9.3 & 5.4 & 4.3 \\
    \bottomrule
  \end{tabular}
\end{table}

\subsubsection{ImageNet ablations}
\label{app:ablations}
\label{app:in-levers}

Figure~\ref{fig:abl_two} uses DiT-B/2 in SD-VAE latents, trained from
scratch for 20 epochs with one seed. The cumulative sequence starts at
FID 75.6. Reducing batch size to 256, using the native late transport-time
schedule, switching to ADM preprocessing, and applying the native
learning-rate drop gives 46.7. Guided transport and boundary targets,
together with the smoothed gradient balancer, give 20.7 at equal epochs
and 25.9 at equal compute. The remaining unguided recipe changes net to
zero in this comparison.

\begin{table}[t]
\centering
\small
\begin{tabular}{@{}lr@{}}
\toprule
Configuration / change & FID / change in FID \\
\midrule
Starting BTM configuration & 75.6 \\
Native BTM after the four setup changes & 46.7 \\
SBM at 20 epochs & 20.7 \\
SBM at native BTM's compute budget & 25.9 \\
\midrule
Remove transport-residual normalization & $+40.3$ \\
Use live guidance predictions instead of the frozen copy & $+7.8$ \\
Remove guided boundary targets and diagonal boundary rows & $+5.7$ \\
Use the tested fixed gradient ratio on the B/2 recipe & $+4.7$ \\
Add the late boundary weight to the B/2 recipe & $+5.4$ \\
\bottomrule
\end{tabular}
\caption{\textbf{ImageNet ablation summary.} Absolute FID values and
reported changes at base size; positive changes worsen FID. The
equal-compute row uses the native BTM budget. The fixed-ratio ablation
uses a ratio measured with late boundary weighting and applied without
it, so its effect is specific to that mismatch.}
\label{tab:ablations}
\end{table}

The transport normalization divides each squared residual by its stopped
value plus $0.01$. Diagonal boundary rows set $S'=\gamma$ for half
the boundary examples. The frozen guidance copy is refreshed every 10k
updates from EMA $0.999$. These definitions specify the components
removed in Figure~\ref{fig:abl_two}. The late boundary weight is absent
from the final B/2 recipe; the earlier XL/2 and RAE runs retain it.
Applying the same residual normalization to boundary rows gives FID 95.6.

\paragraph{Guidance and read-time tradeoffs.}
\label{app:abl-tradeoff}
At 20 epochs, raising training guidance from $\omega=5$ to 10 changes
FID from 20.7 to 19.3 without changing per-update compute; widening the
guidance band beyond $[0.35,0.6]$ did not improve the train-reference
read. Read-time $\kappa$ trades class-typicality against fidelity
(Figure~\ref{fig:fid-is}); the final RAE XL model reaches its lowest
FID at $\kappa=1.35$. For B/2, the best read scale falls from 2 at
20 epochs to 1.25 from 100 epochs. A final 36-epoch cosine cool-down
improves B/2 FID from 6.44 to 5.85 at the same 156-epoch budget.

\paragraph{Implicit against explicit guided targets.}
\label{app:abl-guide-variance}
Five B/2 arms fork from the recipe at the step where guidance starts (35k
updates, 7 epochs) and differ in the guided target alone: the implicit
correction of \eqref{eq:direction} or the explicit guided targets, at the same
$\omega$ (Table~\ref{tab:guide-variance}). Both arms follow
Algorithm~\ref{algo:training_complicated}, with its noisy boundary inputs and
its frozen guidance copy $\bar T$; on the rows where guidance is active, the
explicit arm replaces the transport direction by
$\dot I_S^{\omega,\mathrm{exp}}=\omega \dot I_S-(\omega-1)\frac{\bar T(I_S,\emptyset,S)-I_S}{1-S}$
and the boundary target $x_1^\omega$ by
$\omega x_1-(\omega-1)\,\bar T(I_{S'},\emptyset,S')$, so that
\begin{equation}
 \mathcal L_{\partial}^{\mathrm{exp}}
 =\E_c\left\|T_\theta(I_{S'},c,\gamma)-\sg\!\left(\omega x_1-(\omega-1)\,\bar T(I_{S'},\emptyset,S')\right)\right\|^2 .
 \label{eq:boundary-loss-explicit}
\end{equation}
Given $I_{S'}$, the conditional mean of this target is
$\omega\,\E[x_1\mid I_{S'}]-(\omega-1)\bar T(I_{S'},\emptyset,S')$, the CFG
denoiser when the null branch of $\bar T$ is the null denoiser, and its
fluctuation is $\omega$ times that of $x_1$. The implicit target's gradient variance does
not grow with $\omega$ and stays at the level of unguided BTM; the explicit
target's grows faster than $\omega^2$, and the FID follows: a tie at
$\omega=1.5$, 2.5 points at $\omega=3$ and 45 at the recipe's $\omega=5$. The
explicit update skips the conditional read of the frozen copy and costs 0.85
times as much; at $\omega=3$ this would roughly close its gap at matched
compute, at $\omega=5$ it would not.

\begin{table}[t]
\centering
\small
\begin{tabular}{c cc cc}
\toprule
& \multicolumn{2}{c}{FID at 20 epochs} & \multicolumn{2}{c}{gradient variance} \\
\cmidrule(lr){2-3} \cmidrule(lr){4-5}
$\omega$ & implicit & explicit & implicit & explicit \\
\midrule
1.5 & 32.4 & 31.4 & 3.1 & 5.6 \\
3   & 23.5 & 25.9 & 3.8 & 24 \\
5   & 20.9 & 66.1 & 3.8 & 323 \\
\bottomrule
\end{tabular}
\caption{\textbf{Implicit against explicit guided targets.} DiT-B/2 maps in
SD-VAE latents forked from the recipe at 7 epochs and trained to 20 with either
target; one seed. FID-50k (ADM reference, best of two EMA copies and three
label scales) at 20 epochs, and the gradient variance of the boundary term: the
trace of the covariance of its parameter gradient over 16 batches of 256,
averaged over the 12, 16 and 20-epoch checkpoints ($\times 10^{3}$). The
control is a re-run of the recipe (20.9 against 20.7 in
Table~\ref{tab:ablations}).}
\label{tab:guide-variance}
\end{table}

\subsection{Text-to-image}
\label{app:t2i}

\subsubsection{Data and architecture}
\label{app:t2i-data}
\label{app:t2i-map}

We use MiniT2I's recaptioned CC12M for pretraining and its curated
120k mix for alignment~\citep{minit2i2026,changpinyo2021cc12m,priming}.
The CC12M mirror contains 10,968,539 images, paired by URL with
LLaVA-NeXT recaptions. After filtering prompts longer than 256 tokens,
the alignment mix contains 119,228 pairs from BLIP3o, DALL-E~3, and
ShareGPT-4o. Its shards are reshuffled to mix sources.
Prompts are encoded by frozen FLAN-T5-Large in fp32, with a maximum
length of 256 tokens~\citep{chung2022flan}.

Images use the frozen DINOv2-S encoder with registers at 224\,px,
producing $16\times16\times384$ tokens normalized with ImageNet
statistics. Encoding is performed on the fly. The ImageNet-trained
RAE ViT-XL decoder produces 256\,px images; neither encoder nor decoder
is fine-tuned. Nominal time $u$ is mapped to interpolation time $s$
through the dimension shift
$s=u/[u+4.899(1-u)]$.
All reported boundary levels, time-sampling intervals, and guidance
bands are in nominal time.

The maps extend the ImageNet DiT$^{\mathrm{DH}}$ architecture
(Appendix~\ref{app:in-map}) with pooled prompt conditioning in adaLN
and cross-attention in every trunk block, between self-attention
and the MLP~\citep{chen2023pixart}. Prompt states pass through a linear
projection, sinusoidal positions, and two text-refiner blocks.
Cross-attention uses qk-normalization and zero-initialized output
projections. The null prompt replaces all tokens, including padding,
with a learned mask token; the wide head receives conditioning through
the trunk. Only image inputs carry a tangent in the transport JVP.
SBM-XL has 814.7M trainable parameters. SBM-B has a 15-block trunk
of width 768 and a two-block head of width 1024, totaling 253.8M.

\subsubsection{Training and checkpoint construction}
\label{app:t2i-recipe}

We use Algorithm~\ref{algo:training_complicated} and the ImageNet RAE
recipe of Table~\ref{tab:recipes}, including its time sampling,
transport normalization, late boundary weighting, and gradient
balancing. Guidance uses $\omega=5$ on the nominal band $[0.35,0.6]$,
starts at update 35k, and reads a frozen copy refreshed every 10k
updates from EMA $0.999$. Prompt dropout is $0.1$.

Muon uses learning rate $5\times10^{-4}$ on matrix weights and
AdamW $5\times10^{-5}$ on the remaining parameters, with constant
rates, no warm-up, batch 256, and bf16 autocast. Training uses
16 H200s and maintains EMA copies $0.999$ and $0.9999$.
Student attention uses explicit matrix products, which support the
transport JVP and avoid the instability observed with fused bf16
attention.

Both maps train for 1M updates on CC12M and 160k on the alignment
mix, matching MiniT2I's sample budgets. Checkpoint averaging is
performed before and after alignment, with the head treated
separately. XL additionally receives 5k head-only updates at
$5\times10^{-5}$ after the pretraining average. The final head-only
continuation uses rate $2\times10^{-4}$ and equal sampling from
CC12M and the alignment mix, with the remaining network frozen.
Table~\ref{tab:t2i_stages} specifies the checkpoint construction
and final read settings.

\begin{table}[t]
\centering
\small
\setlength{\tabcolsep}{4pt}
\caption{\textbf{Text-to-image checkpoints and final results.}
Checkpoint indices are global updates at batch 256. Reported reads
use EMA $0.9999$. Trunk averages include the embedders; averages
are uniform except where stated. Read boundary levels are nominal.
GenEval means and ranges use four seeds; the main paper reports
the best seed.}
\label{tab:t2i_stages}
\resizebox{\linewidth}{!}{%
\begin{tabular}{@{}lll@{}}
\toprule
 & SBM-B & SBM-XL \\
\midrule
Pre-alignment trunk average
 & 850k--1,000k
 & 500k--1,000k, exponential, half-life 100k \\
Pre-alignment head
 & 995k
 & 1,000k; then 5k head-only updates \\
Post-alignment trunk average
 & 1,040k--1,160k
 & 1,020k--1,165k \\
Post-alignment head
 & 1,160k
 & average of 1,050k--1,070k \\
Final head-only updates & 5k & 35k \\
Total updates & 1,165k & 1,200k \\
Read $(\gamma_{\mathrm{read}},\kappa)$
 & $(0.3,1.25)$ & $(0.35,1.125)$ \\
GenEval mean & 0.826 & 0.848 \\
GenEval range & 0.819--0.833 & 0.844--0.856 \\
GenEval best & \textbf{0.833} & \textbf{0.856} \\
Training compute (ZFLOP) & 0.35 & 1.15 \\
Inference compute (TFLOP/image) & 0.33 & 0.57 \\
\bottomrule
\end{tabular}}
\end{table}

\paragraph{Effect of the training stages.}
\label{app:t2i-levers}
Alignment raises GenEval from 0.37 to 0.81 for B and from 0.46
to 0.82 for XL. Post-alignment averaging adds approximately
0.02--0.03. Subsequent head-only training changes the four-seed
mean by 0.004 for B and less than 0.001 for XL, within the
observed seed spread.

\subsubsection{Sampling and evaluation}
\label{app:t2i-protocol}

After prompt encoding, sampling uses one EMA-$0.9999$ map evaluation
on fresh Gaussian noise and one decoder call. Read-time label
scaling applies
$e\leftarrow e_\emptyset+\kappa(e-e_\emptyset)$
to the prompt embeddings without evaluating the null branch.
Read levels and label scales are selected on GenEval itself using
seed-0 sweeps, refined after alignment and checked with additional
seeds. There is no separate held-out GenEval selection split;
uncertainty is assessed with paired bootstrap resampling of prompts
within each task, retaining each prompt's four images.

GenEval uses all 553 prompts, four images per prompt, and the
released scorer; the overall score averages the six task
accuracies~\citep{ghosh2023geneval}. All our reads use H200 GPUs.
Table~\ref{tab:t2i_stages} reports four-seed means, ranges, and
best-seed scores. Our images are generated at 256\,px, whereas
published baselines use 512\,px. As a resolution check, downsampling
MiniT2I-B/16's images to 256\,px changes its score from 0.868 to
0.869 under the same scorer. Our reproduction of its released
100-step, CFG-6 sampler scores 0.868, compared with the published
0.873.

\subsubsection{Compute accounting}
\label{app:t2i-compute}

We count two FLOPs per multiply-accumulate with torch's FLOP counter,
attention as a plain matrix product and 256 prompt tokens; text-encoder
compute is excluded throughout. A forward pass costs
54.3 GMACs for SBM-B, 171.9 for SBM-XL, and 110.7 for the decoder,
giving the inference costs in Table~\ref{tab:t2i_stages}. Counted the
same way on our implementation of the MiniT2I architectures at 512\,px, a
B/16 forward costs 201 GMACs and an L/16 forward 660, so the 100-step
samplers with CFG cost 80 and 264 TFLOP per image and the four-step
distilled model 1.6 TFLOP. The MiniT2I project page lists 570 and 1,493
GFLOP per forward for the two models, which would give 114 and 299 TFLOP
per image; we keep our own count, which measures every model by one rule.

Training compute uses 7.05 forward-equivalents per sample before
guidance and 11.05 afterwards. These factors are transferred from
measurements of the ImageNet B/2 recipe, rather than measured
directly on the text-to-image maps. Head-only updates are counted
at the full training rate, and DINOv2 image encoding is excluded.
This gives 0.35 ZFLOP for B and 1.15 ZFLOP for XL.

Pixel-space baseline budgets follow the MiniT2I authors' estimates.
For MeanFlow distillation, whose training compute is unreported,
we add an estimated 0.13 ZFLOP to the teacher's 0.36 ZFLOP:
50k updates at batch 1,024 and 2.55 TFLOP per example, accounting
for teacher guidance, student forward/backward passes, and the
stopped JVP. The compiled guided training step processes
approximately 64 images/s per H200, giving about 69 hours for
1M updates on 16 GPUs.

\subsection{World modeling}
\label{app:worldmodel}

\subsubsection{Data, architecture, and training}
\label{app:wm-data}
\label{app:wm-map}
\label{app:wm-recipe}

We use DriftWorld's conversion scripts and splits for BridgeData V2
and RT-1 (\texttt{fractal20220817\_data})~\citep{bridge2,rt1,driftworld}.
Bridge-V2 uses the \texttt{image\_0} camera, seven-dimensional actions,
53,192 training episodes, and 6,872 validation episodes (6,871 evaluated).
RT-1 uses the main camera at 3 fps, ten-dimensional actions, and a seeded
$2.5\%$ validation hold-out from 87,212 episodes (2,165 evaluated).
Frames are $256\times256$ and cached as $32\times32\times16$ SD3
autoencoder latents, scaled by $1.5305$~\citep{esser2024sd3}.
Training samples five-frame windows: four history frames and one target,
with the corresponding actions.

The map uses DriftWorld's Bridge-V2 U-Net (175.6M parameters by our
count), concatenating the four history latents with the interpolant and
conditioning on actions through FiLM and cross-attention. A Fourier
embedding of $\gamma$ is added to the FiLM vector. Condition dropout
replaces actions with a learned null vector but retains the history;
RT-1 changes only the action tokenizer. Attention is evaluated with
explicit matrix products to support the transport JVP.

We use Algorithm~\ref{algo:training_complicated} at $\omega=1$
and without an auxiliary feature loss, with the following overrides. Boundary times are $S'\sim U(\gamma,1)$;
from step 30k, rows with $S'>0.6$ receive weight
$\min\{0.4/(1-S'),8\}$. The normalized transport loss has fixed
weight 30 relative to the boundary loss. AdamW uses weight decay $0.01$,
1,000 warm-up steps, gradient clipping at 2, and bf16 autocast.
Bridge-V2 uses batch 128 and learning rate $2\times10^{-4}$ on two
H200s; RT-1 uses batch 256 and $4\times10^{-4}$ on four H200s.
Reported checkpoints use EMA $0.999$. Training is teacher-forced, with
a constant rate followed by a cosine decay to $5\%$ of that rate.
The final maps use 600k + 60k cool-down updates on Bridge-V2 and
800k + 120k on RT-1; the low-compute RT-1 result is the 120k constant-rate
checkpoint. Other defaults, including the transport-time mixture,
$\gamma_{\min}=0.35$, and dropout $0.1$, follow the algorithm.

\subsubsection{Sampling, evaluation, and compute}
\label{app:wm-protocol}

Each frame uses fresh Gaussian noise and one map evaluation, with
$\gamma_{\mathrm{read}}=0.8$ on Bridge-V2 and $0.9$ on RT-1.
Quantitative evaluation follows DriftWorld: each validation episode is
generated in eight-frame chunks, resetting the four-frame history to
ground truth at each chunk. The reference is the autoencoder reconstruction
of the ground truth. SSIM, PSNR, and LPIPS are averaged per video;
FID uses all frames; FVD uses the first sixteen generated frames of
videos with at least sixteen frames (1,951 Bridge-V2 and 1,936 RT-1
videos), with the StyleGAN-V I3D implementation
\citep{skorokhodov2022styleganv}. On DriftWorld's released Bridge-V2
checkpoint, our evaluator gives SSIM/PSNR/LPIPS of
$0.833/22.62/0.091$ on the first 256 validation episodes, compared
with the published $0.835/22.69/0.091$.
Development uses those 256 episodes; all retained quantitative results
use the full validation split. Free rollouts without history resets
are shown separately in Appendix~\ref{app:wm-qual}.

\paragraph{Compute.}
\label{app:wm-compute}
A measured forward-and-backward training step costs
$1.84\times10^{11}$ FLOP per example (8.0 U-Net forwards).
Multiplying by updates and batch gives 15.6 EFLOP for Bridge-V2 and
43.4 EFLOP for the final RT-1 map. DriftWorld's budgets are our count of
its released training step, 550 and 549 TFLOP per update, over the
post-training stage of the released checkpoints, 55{,}200 and 29{,}400
updates, which gives 30.4 EFLOP on
Bridge-V2 and 16.1 on RT-1; the length of its first stage is not reported,
so these are lower bounds. DriftWorld draws 64 generated negatives per
target frame and decodes each through the SD3-VAE for the DINOv3 loss;
that decoding is nine tenths of its training compute. Inference uses 23.1 GFLOP per
frame in the map; on one H200 at batch 1 it takes 14.2 ms in bf16, with another
6.6 ms for the SD3 decoder. Table~\ref{tab:rt1} reports the sum, 0.021 s per
frame; DriftWorld's released evaluation code times generation and decoding
together.

\subsubsection{Additional comparisons}
\label{app:wm-levers}

The training choices were selected on Bridge-V2 with 256-episode reads
and transferred to RT-1. Increasing normalized transport weight from
1 to 10 improves PSNR by 0.8 dB at matched steps; weight 30 adds
0.3 dB. A final cool-down improves quality in the reported runs.
Table~\ref{tab:wm_ladder} retains additional full-validation configurations
complementing the main results.

\begin{table}[t]
\centering
\small
\setlength{\tabcolsep}{4pt}
\resizebox{\linewidth}{!}{%
\begin{tabular}{llrccccc}
\toprule
Dataset & Configuration & EFLOP & SSIM $\uparrow$ & PSNR $\uparrow$ & LPIPS $\downarrow$ & FID $\downarrow$ & FVD $\downarrow$ \\
\midrule
Bridge-V2 & 400k + 60k cool-down & 10.9 & 0.866 & 24.62 & 0.062 & 4.71 & 18.3 \\
RT-1 & 600k + 200k cool-down & 37.7 & 0.879 & 26.85 & 0.058 & 4.08 & 16.8 \\
\bottomrule
\end{tabular}}
\caption{Additional full-validation world-model results. Runs are cool-down
forks of the same constant-rate run per dataset, without auxiliary feature losses.}
\label{tab:wm_ladder}
\end{table}

\paragraph{DriftWorld without its feature encoder.}
\label{app:wm-featurefree}
We run DriftWorld's released code with its DINOv3 term disabled through
both training stages, retaining its batch of one clip per GPU on two GPUs.
Table~\ref{tab:wm_featurefree} gives full-split results. These runs have
higher SSIM and PSNR than the published DriftWorld model but worse
LPIPS, FID, and FVD. SBM at 10.9 EFLOP outperforms all three DriftWorld
rows on all five metrics; the table reports their different training budgets.

\begin{table}[!htbp]
\centering
\small
\setlength{\tabcolsep}{4pt}
\resizebox{\linewidth}{!}{%
\begin{tabular}{lrccccc}
\toprule
Model & EFLOP & SSIM $\uparrow$ & PSNR $\uparrow$ & LPIPS $\downarrow$ & FID $\downarrow$ & FVD $\downarrow$ \\
\midrule
DriftWorld, no feature loss, stage 1 & 0.98 & 0.847 & 23.20 & 0.140 & 46.5 & 252 \\
DriftWorld, no feature loss, stages 1 + 2 & 2.47 & 0.855 & 23.83 & 0.139 & 45.9 & 231 \\
DriftWorld, DINOv3 (published) & $\geq$30.4 & 0.835 & 22.69 & 0.091 & 4.86 & 46.1 \\
SBM, 400k + 60k cool-down & 10.9 & 0.866 & 24.62 & 0.062 & 4.71 & 18.3 \\
\bottomrule
\end{tabular}}
\caption{Bridge-V2 comparison with DriftWorld trained without its auxiliary
feature loss. The published DriftWorld row uses DINOv3; all other rows
use no auxiliary feature loss. All rows use the full validation split;
training budgets differ as shown.}
\label{tab:wm_featurefree}
\end{table}

\FloatBarrier
\section{Additional samples}
\label{app:samples}

\subsection{Text-to-image}
\begin{figure}[!htbp]
 \centering
 \includegraphics[width=\linewidth,height=0.92\textheight,keepaspectratio]{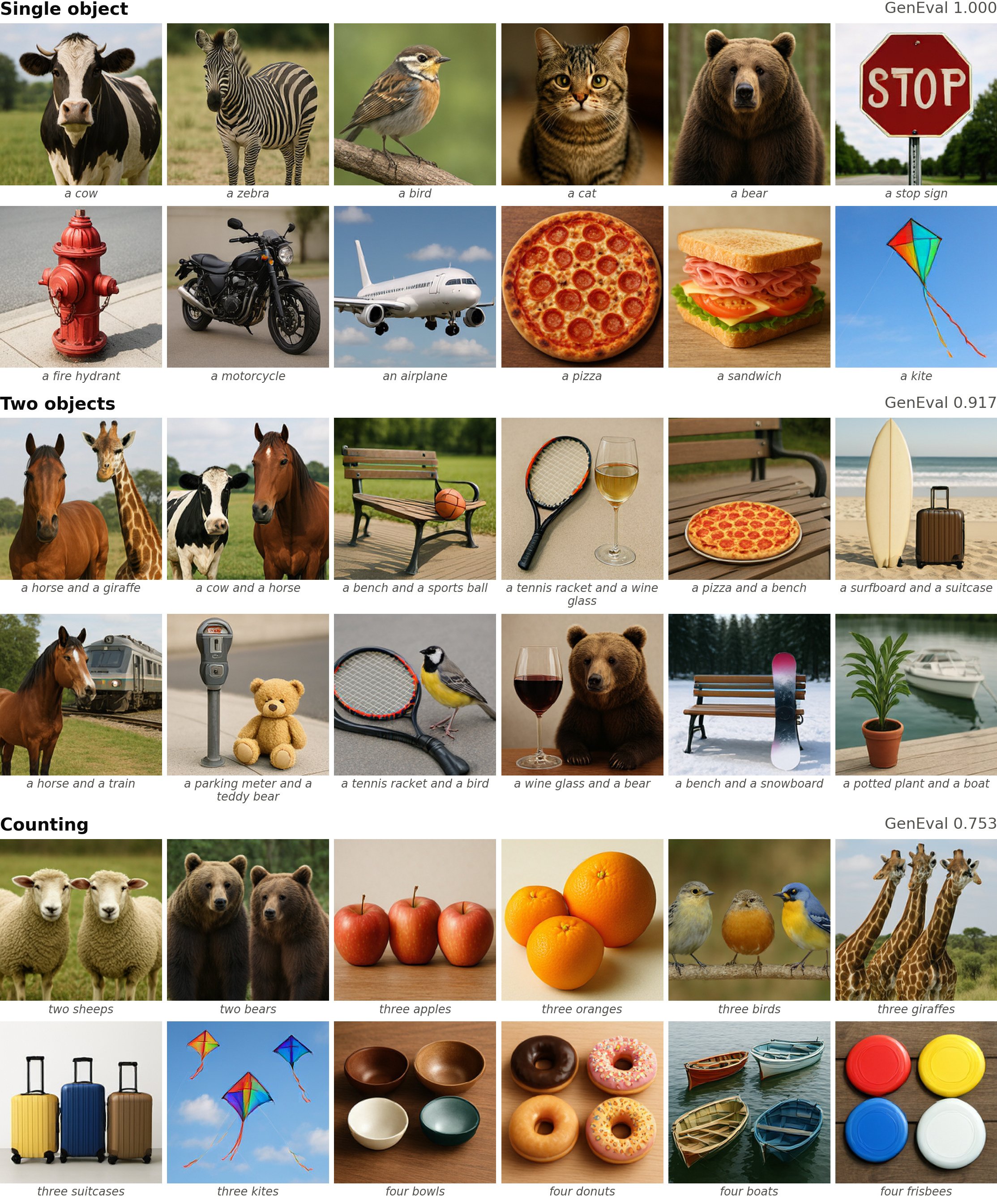}
 \caption{\textbf{Uncurated examples of Scalable Beckmann Maps (SBM)} On three submetrics of GenEval: single object, two object, and counting}
 \label{fig:geneval class1}
\end{figure}

\begin{figure}[!htbp]
 \centering
 \includegraphics[width=\linewidth,height=0.92\textheight,keepaspectratio]{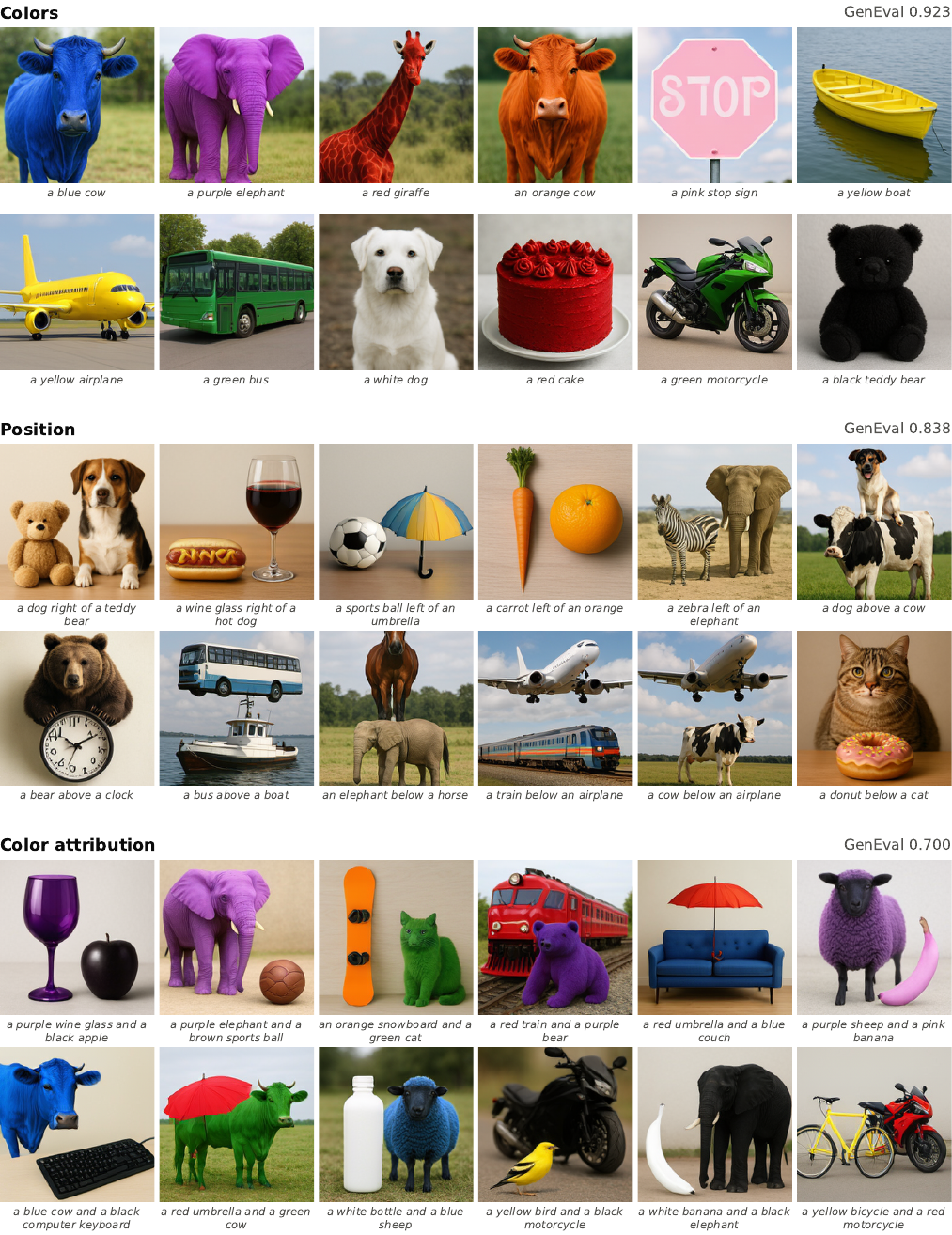}
 \caption{\textbf{Uncurated examples of Scalable Beckmann Maps (SBM)} On three submetrics of GenEval: colors, position, and color attribution}
 \label{fig:geneval class2}
\end{figure}

\subsection{World-model rollouts}
\label{app:wm-qual}

\begin{figure}[!htbp]
\centering
% Repack the four episodes into 2x2 grids instead of narrow vertical strips.
\begin{minipage}[t]{0.49\linewidth}
  \vspace{0pt}
  \includegraphics[width=\linewidth,trim=0 3152bp 0 0,clip]{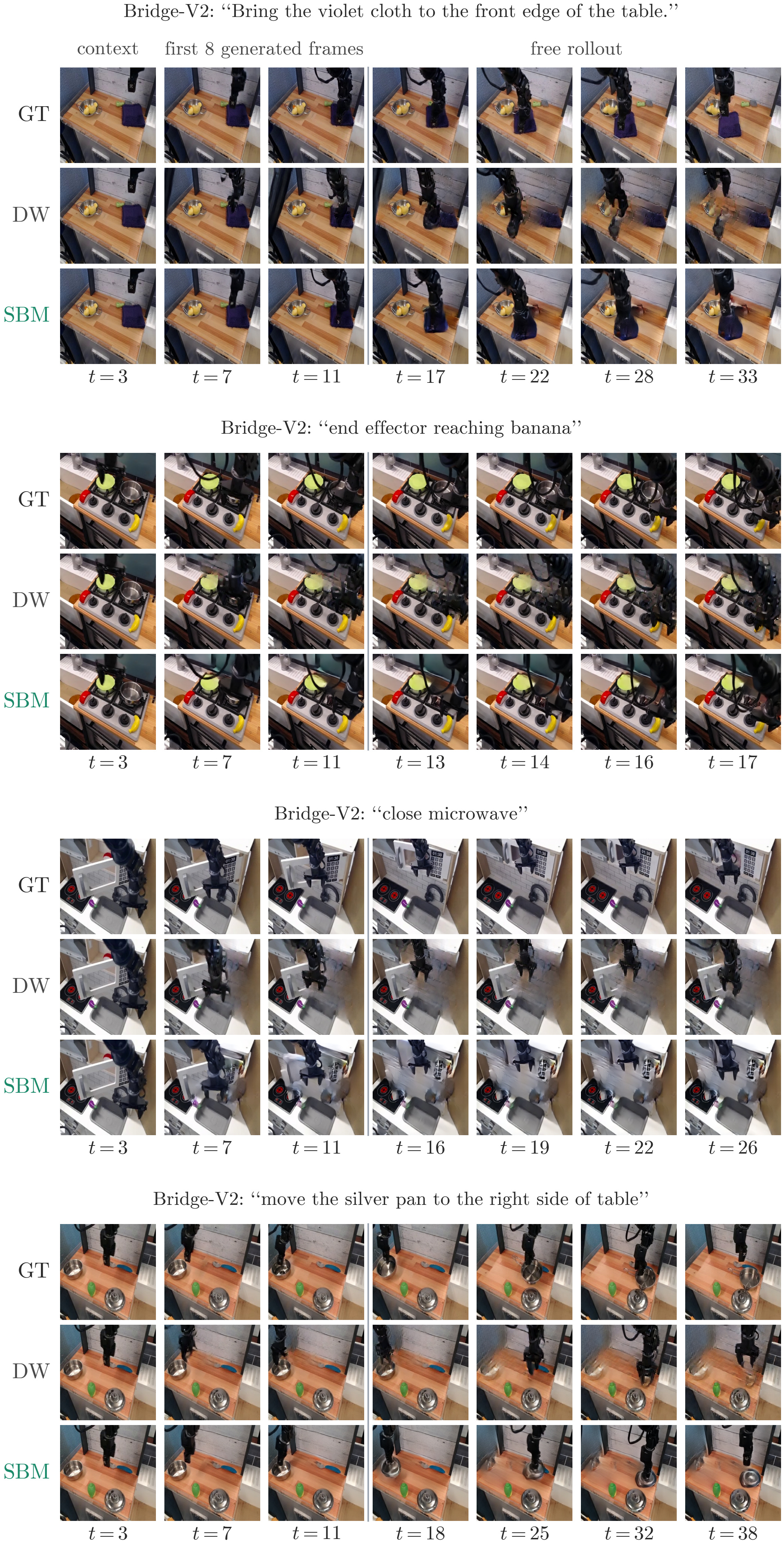}\par
  \vspace{3pt}
  \includegraphics[width=\linewidth,trim=0 1062bp 0 2186bp,clip]{fig_bridge_app.jpg}
\end{minipage}\hfill
\begin{minipage}[t]{0.49\linewidth}
  \vspace{0pt}
  \includegraphics[width=\linewidth,trim=0 2107bp 0 1139bp,clip]{fig_bridge_app.jpg}\par
  \vspace{3pt}
  \includegraphics[width=\linewidth,trim=0 16bp 0 3230bp,clip]{fig_bridge_app.jpg}
\end{minipage}
\caption{Bridge-V2 rollouts, four validation episodes in a $2\times2$ grid (rows within each panel: ground truth, DriftWorld's released model, SBM). As in
Figure~\ref{fig:wm}, the columns left of the divider lie inside the first chunk of eight generated frames and the columns right of
it continue the same rollout from the model's own frames without resetting the conditioning history. The first eight generated frames of a free rollout
coincide with the protocol's first chunk.}
\label{fig:wm_bridge_appendix}

\vspace{8pt}
\begin{minipage}[t]{0.49\linewidth}
  \vspace{0pt}
  \includegraphics[width=\linewidth,trim=0 2269bp 0 0,clip]{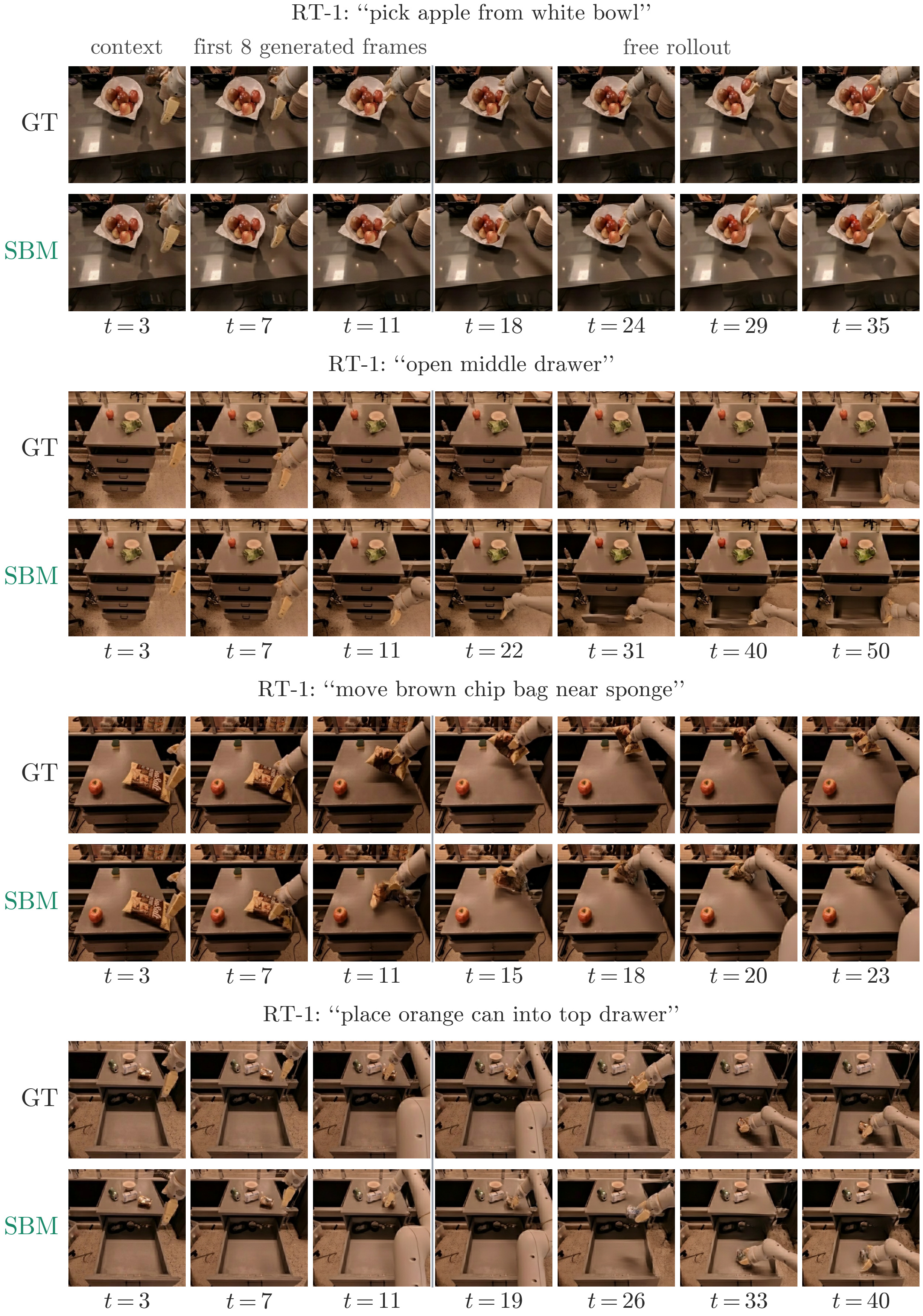}\par
  \vspace{3pt}
  \includegraphics[width=\linewidth,trim=0 767bp 0 1572bp,clip]{image.jpg}
\end{minipage}\hfill
\begin{minipage}[t]{0.49\linewidth}
  \vspace{0pt}
  \includegraphics[width=\linewidth,trim=0 1518bp 0 821bp,clip]{image.jpg}\par
  \vspace{3pt}
  \includegraphics[width=\linewidth,trim=0 16bp 0 2323bp,clip]{image.jpg}
\end{minipage}
\caption{RT-1 rollouts, four validation episodes (rows within each panel: ground truth, SBM). Layout as in
Figure~\ref{fig:wm_bridge_appendix}.}
\label{fig:wm_rt1_appendix}
\end{figure}

Figures~\ref{fig:wm_bridge_appendix} and~\ref{fig:wm_rt1_appendix}
show both the scored eight-frame chunk and free rollouts without history
resets. The latter use generated frames and ground-truth actions for the
rest of the episode (30--60 frames on Bridge-V2). SBM retains scene and
arm structure longer in these examples; large camera-filling motion,
such as opening a fridge, remains difficult for both models.

\end{document}